\documentclass[final,5p,times,twocolumn,authoryear]{elsarticle}

\usepackage{amssymb}

\usepackage{color}
\usepackage{xcolor}
\usepackage{amsmath}
\usepackage{wrapfig}
\usepackage{soul}
\usepackage{hyperref}
\usepackage{etoolbox}
\usepackage{verbatim}
\usepackage{multirow} 
\usepackage{algorithm}
\usepackage{algpseudocode}
\usepackage{amsmath}
\usepackage{appendix}
\usepackage{hyperref}
\usepackage{rotating}
\usepackage{adjustbox}
\usepackage{array}
\usepackage{rotating}
\usepackage{graphicx}
\usepackage{float}
\usepackage{subcaption} 
\usepackage{rotating}  
\newcommand{\ours}{TLSA}
\newcommand{\eg}{\textit{e}.\textit{g}.}
\usepackage{booktabs}
\usepackage{colortbl}
\usepackage{graphicx}
\usepackage{amssymb}
\usepackage{pifont}
\usepackage[para,online,flushleft]{threeparttable}%
\newcommand{\cmark}{\ding{51}}%
\newcommand{\xmark}{\ding{55}}%

\journal{Engineering Applications of Artificial Intelligence}

\begin{document}



\newpage

\begin{frontmatter}



\title{Training-Free Label Space Alignment for Universal Domain Adaptation}


\affiliation[inst1]{organization={Department of Artificial Intelligence, Korea University},
            addressline={145 Anam-ro, Seongbuk-gu}, 
            city={Seoul},
            postcode={02841},
            country={South Korea}}

\affiliation[inst2]{organization={OMRON SINIC X},
            addressline={5-24-5 Hongo, Bunkyo-ku}, 
            city={Tokyo},
            postcode={113-0033}, 
            country={Japan}}


\author[inst1]{Dujin Lee}
\ead{ldj0305@korea.ac.kr}
\author[inst1]{Sojung An}
\ead{sojungan@korea.ac.kr}
\author[inst1]{Jungmyung Wi}
\ead{wjm333@korea.ac.kr}
\author[inst2]{Kuniaki Saito}
\ead{kuniaki.saito@sinicx.com}

\cortext[mycorrespondingauthor]{Corresponding author}
\author[inst1]{Donghyun Kim\corref{mycorrespondingauthor}}
\ead{d_kim@korea.ac.kr}

\begin{abstract}
Universal domain adaptation (UniDA) transfers knowledge from a labeled source domain to an unlabeled target domain, where label spaces often differ and the target domain often contains private classes. 
Previous UniDA methods primarily focused on visual space alignment, however, severe domain shifts often lead to visual ambiguities, undermining their ability to predict the label space shift.
Our preliminary experiment shows that focusing exclusively on label space alignment can be a more effective solution to these limitations. While generative vision-language models (VLMs) offer strong zero-shot capabilities that could help uncover target-private classes, naively employing them introduces challenges such as noisy and semantically ambiguous labels. To address these challenges, we propose a novel approach that carefully utilizes generative VLMs to improve label space alignment. We introduce an adaptive thresholding strategy that leverages meaningful relationships between source and discovered target labels. This allows a training-free mechanism to filter out noisy and ambiguous labels, effectively refining the label space by identifying shared and discovering target-private label subsets. We then build a \textit{universal classifier} for robust self-training, which effectively utilizes reliable pseudo-labels derived from the identified label spaces.
The results reveal that the proposed method considerably outperforms existing UniDA techniques across key DomainBed benchmarks, delivering an average improvement of \textcolor{blue}{+7.9\%} in H-score and \textcolor{blue}{+6.1\%} in H$^3$-score.
Furthermore, incorporating self-training further enhances performance and achieves an additional (\textcolor{blue}{+1.6\%}) increment in both H- and H$^3$-scores.
\end{abstract}


\end{frontmatter}

\section{Introduction}
Domain shift, the discrepancy between training and real-world data, remains a pivotal challenge for learning generalizable vision-language representations. Unsupervised domain adaptation (UDA) \citep{long2015learning, ganin2016domain, saito2018maximum} aims to transfer knowledge from a labeled source domain to an unlabeled target domain, based on the stringent assumption that the label space remains unchanged across domains. However, in practical settings, the target label space often comprises only a subset of source classes that overlap with both domains or include target-exclusive classes unique to the target domain. 
Various studies have explored these complexities. 
Open-set domain adaptation (ODA) and partial-set domain adaptation (PDA) assume that the label space of each domain is a subset of another’s \citep{panareda2017open,cao2018partial}. 
Nonetheless, these approaches rely on prior knowledge of label space shifts, which limits their real-world applicability. 
To address this constraint, Universal Domain Adaptation (UniDA) was introduced \citep{You_2019_CVPR,saito2020universal} for broader applicability.

\begin{figure*}[t]
    \centering
    \begin{minipage}{0.49\textwidth}
        \centering
        \includegraphics[width=\linewidth]{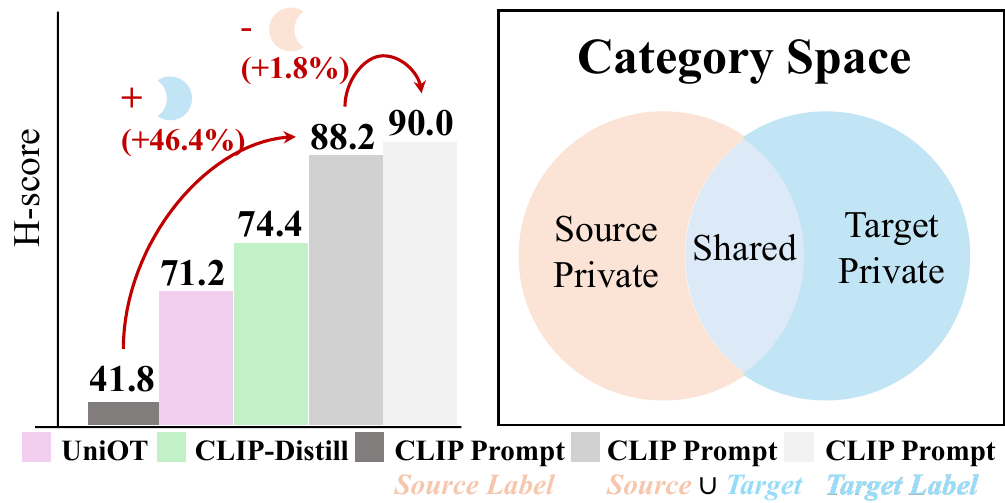}
        \caption{Comparison between CLIP Prompt, a zero-shot classifier enhanced with corresponding label text prompts, and previous state-of-the-art (SOTA) methods. Simply adding target-private classes or removing source-private classes leads to substantial gains in H-score, achieved entirely without additional training.}
        \label{fig:teaser}
    \end{minipage}
    \hfill
    \begin{minipage}{0.49\textwidth}
        \centering
        \includegraphics[width=\linewidth]{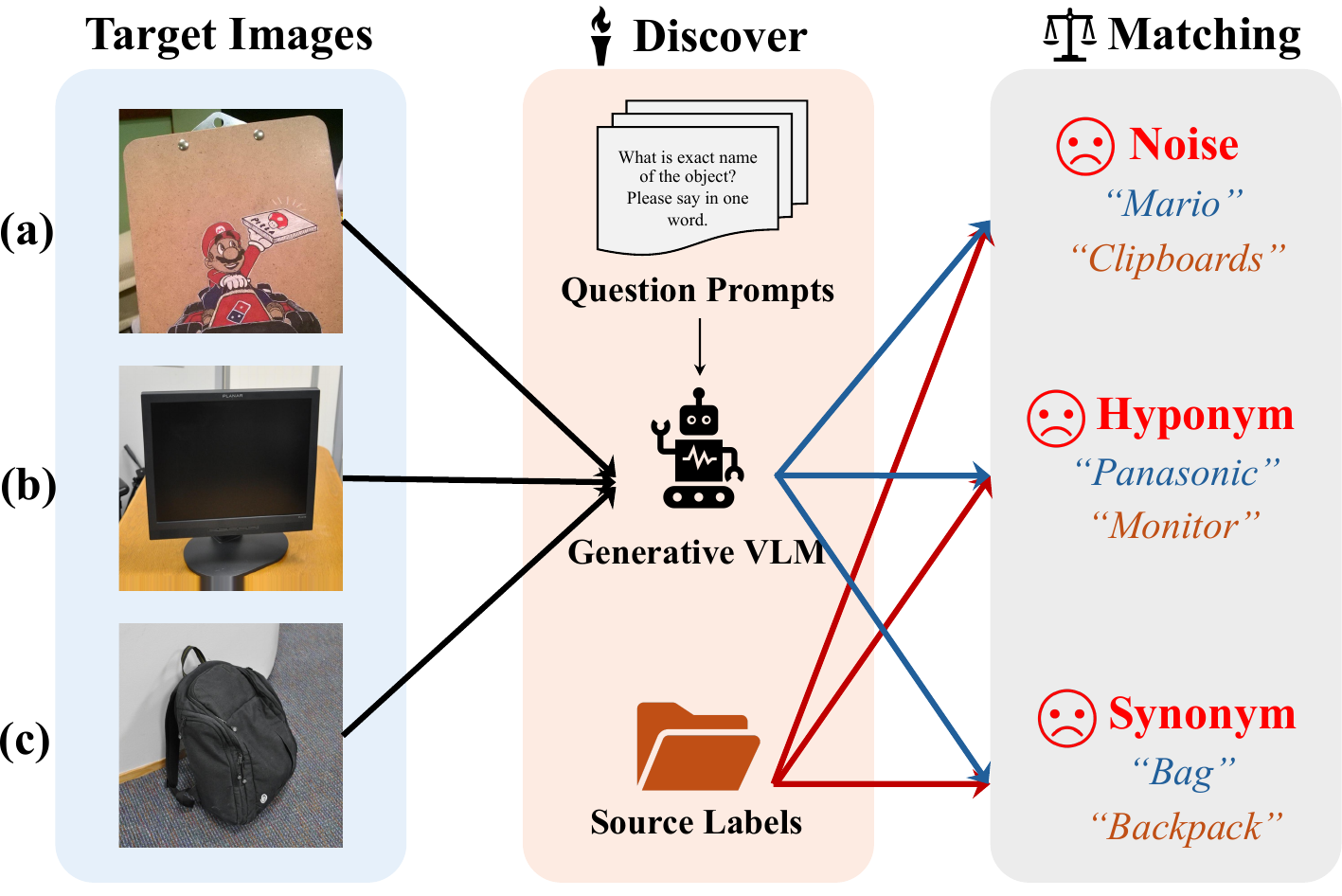}
        \caption{Examples of semantic ambiguity in generative VLM outputs: (a) noisy response, (b) hyponym prediction, and (c) synonym prediction.}
        \label{fig:misclassified_example}
    \end{minipage}
\end{figure*}

In UniDA, a primary challenge is distinguishing target-private samples from source-private ones. Many studies have proposed innovative detection criteria utilizing entropy \citep{saito2020universal, qu2023upcycling}, binary classifiers \citep{saito2021ovanet, lu2024mlnet}, optimal transport \citep{chang2022unified}, energy-based methods \citep{chen2022geometric}, class prototypes \citep{li2021domain, lai2023memory}, and calibration techniques \citep{deng2023universal}. 
These methods still suffer from severe domain shifts, which involve not only style variations but also content changes among shared classes. The reason is that the methods rely on the geometric structure of visual embeddings; however, such features are often drastically shifted across domains, leading to inconsistent decisions.

This study addresses these limitations by focusing exclusively on label space alignment. To achieve this, we leverage the robust zero-shot capabilities of Vision-Language Models (VLMs) such as CLIP~\citep{radford2021learning} and ALIGN~\citep{li2021align}. For instance, CLIP effectively recognizes diverse classes across domains by prompting label names through its fixed text encoder. To evaluate this ability, we conducted a pilot study assessing the performance of a zero-shot classifier across various category spaces within the UniDA framework. As shown in Fig.~\ref{fig:teaser}, zero-shot classifiers using customized label spaces exhibit substantial improvements in UniDA performance. Interestingly, merely augmenting the zero-shot classifier with oracle category spaces already surpasses existing UniDA approaches, most of which attempt to achieve alignment solely in the visual space. These findings suggest that addressing the UniDA problem is more efficiently achieved by aligning label spaces rather than visual spaces across domains.

Inspired by Fig.~\ref{fig:teaser}, we aim to identify target-private classes and obtain an empirical version of these classes, termed \textit{predicted target-private labels}, by employing generative VLMs~\citep{li2022blip,li2023blip}. In particular, predicted target-private labels are derived from labels discovered in target images using generative VLMs. Nevertheless, distinguishing whether discovered labels belong to part of the source labels or not becomes challenging due to noise (Fig.~\ref{fig:misclassified_example}a) and semantic ambiguities, such as hyponyms (Fig.~\ref{fig:misclassified_example}b) and synonyms (Fig.~\ref{fig:misclassified_example}c). For example, in Fig.~\ref{fig:misclassified_example}b, the generative VLM outputs ``Panasonic,’’ a semantic hyponym of ``Monitor’’ in the source labels. Without proper alignment, ``Panasonic’’ is incorrectly predicted as a target-private label, leading to degraded performance.

To address this, we introduce \textbf{T}raining-free \textbf{L}abel \textbf{S}pace \textbf{A}lignment for UniDA (\textbf{\ours}). Specifically, we first identify the raw labels of the target samples using a generative VLM. Subsequently, we implement three steps to eliminate incorrect or semantically ambiguous discovered labels from the source labels and accurately predict the true target private labels: 
(i)	Synonym label alignment (Sec.~\ref{sec:synonym}): We identify synonyms in raw discovered labels using WordNet and align them with the source labels to remove synonyms; 
(ii) Semantic label alignment (Sec.~\ref{sec:instance}): To resolve the ambiguous similarity between source and discovered labels, we identify a set of predictions that are considered correct for an image within the CLIP joint embedding space through an adaptive thresholding process. If the resulting set excludes source labels and includes only discovered labels, the filtered discovered labels are retained as candidate labels;
(iii) Frequency-based noise-candidate filtering (Sec.~\ref{sec:refine}): To eliminate residual noise, we monitor the frequency of target private candidate labels that pass previous filters across the entire target domain, removing noisy candidate labels based on their occurrence patterns. 

Lastly, leveraging TSLA, we derive the predicted target-private labels, which are used together with source labels to construct a universal classifier for robust self-training. Compared to methods that rely on the geometric structure of visual representations~\citep{chang2022unified,lai2023memory,wang2024exploiting}, our universal classifier more effectively distinguishes between known and unknown labels and produces more reliable pseudo-labels, leading to further performance enhancements (Sec.~\ref{sec:self-training}).
 
To validate our approach, we conducted experiments on four standard UniDA benchmarks, each representing a unique visual recognition domain-adaptation task. In summary, our contributions are:
\begin{enumerate} 
    \item We propose a novel training-free label space alignment method for Universal Domain Adaptation (UniDA), shifting the focus from visual feature alignment to label space alignment using generative vision-language models (VLMs).
    \item To address semantic challenges in naïve label space alignment with generative VLMs, we propose new filtering techniques to resolve semantic ambiguities and noise, yielding a refined shared label space between source and target domains, as well as a well-defined target-private label space.
	\item By applying the filtered target private classes, we enhance the CLIP zero-shot classifier for UniDA, referred to as the universal classifier, which can be further enhanced through self-training to achieve full adaptation to the target domain.
	\item We perform comprehensive analyses to validate the effectiveness of our approach, attaining state-of-the-art performance on H-score and H$^3$-score metrics.
\end{enumerate}

    

Furthermore, as demonstrated in prior studies, fine-tuning the backbone CLIP on source data diminishes its ability to detect target private classes~\citep{deng2023universal}. This issue, known as \textit{feature distortion}~\citep{kumar2022fine, shu2022test, trivedi2023closer, saito2023mind}, arises when the source data is used for tuning. Our method circumvents tuning on the source data by leveraging the source label space to filter and identify target private labels from the generative VLM. 
Furthermore, our approach prevents feature distortion, enabling faster experimentation and improved performance.

\begin{figure*}[t]
    \begin{center}
        \includegraphics[width=\textwidth, trim=0 0 0 0, clip]{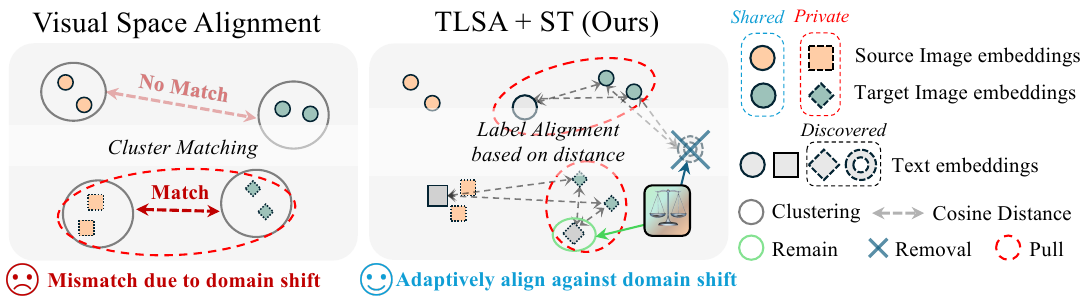}
    \end{center}
    \vspace{-1em}
    \caption{
        Comparison between prior UniDA methods and our \ours+ST.
        \textbf{Left:} Visual space alignment approaches~\citep{li2021domain, chang2022unified, lai2023memory} rely on cluster matching, which becomes ambiguous under domain shift.
        \textbf{Right:} Our TLSA+ST discovers labels from target images and aligns them with source labels with adaptive thresholding using the distance in the embedding space between source labels. Since TLSA+ST uses pretrained CLIP's joint embedding space, it is robust against domain shift.
        }
    \label{fig:comparison}
    \vspace{-1mm}
\end{figure*}

\section{Related Work}
\label{sec:Related Work}
 
\subsection{Vision-Language Models}
Vision-language models have recently demonstrated great potential in learning generic visual representations. \citep{radford2021learning,li2021align,zhou2022learning}. Their transferability is attributed to the ability to inject rich information into the model by aligning image-text pairs, rather than just a single word label. 
For example, CLIP~\citep{radford2021learning} and ALIGN~\citep{jia2021scaling} demonstrate strong zero-shot performance in visual recognition. 
Not surprisingly, many studies have achieved high performance within only zero-shot pipelines by using the Vision-Language Models (VLMs)~\citep{adila2023zero, ge2023improving, qu2024lmc, zhao2024ltgc,zanella2024test, wang2024hard}.
Generative VLMs, including BLIP~\citep{li2022blip, li2023blip}, GPT-4~\citep{achiam2023gpt}, and LLaVA~\citep{liu2024visual}, have demonstrated strong performance on a wide range of vision–language generation tasks.
In light of the latest advancements, we utilize BLIP to discover a label agnostic to the source domain for unlabeled target domain samples.

\subsection{Universal Domain Adaptation} \label{sec:related_work;subsec:Unida}
Unlike UDA, UniDA assumes a distinct scenario where the label set of the target domain often partially intersects with that of the source domain. Determining whether a given data belongs to a known or unknown class is a highly challenging task because the model can only predict the classes it is trained on. 
Therefore, baseline methods for UDA, such as adversarial learning~\citep{ganin2016domain, tzeng2017adversarial, long2018conditional} and metric learning~\citep{long2015learning, long2017deep}, lead to degradation in performance.
The problem occurs because domain alignment is attempted without considering the possibility that the label space might differ, leading to negative transfer. 

With the recent progress of VLMs, there has been growing recognition of the relatively limited exploration compared to pretrained backbones on ImageNet~\citep{russakovsky2015imagenet}, such as ResNet~\citep{he2016deep} and ViT~\citep{dosovitskiy2020image}. 
In response, CLIP-Distillation~\citep{deng2023universal} was proposed, utilizing foundation models~\citep{radford2021learning, oquab2023dinov2} trained on large datasets through self-supervised learning. 
This approach achieved state-of-the-art performance among CLIP-pretrained backbones by performing auto-calibration via learning the classifier temperature. 
Similarly, MemSPM~\citep{lai2023memory} adopts a memory-based approach that models sub-classes by storing and updating sub-prototypes, while keeping the CLIP backbone frozen. 
Here, CLIP is employed not for its zero-shot capability, but merely as a generic feature extractor.
Fig.~\ref{fig:comparison} illustrates the key differences between prior works and our approach. 
Unlike previous methods, and as motivated by Fig.~\ref{fig:teaser}, we reformulate UniDA as the simpler task of discovering target-private class names. 
Specifically, we generate candidate labels directly from target-domain samples using a generative VLM, and then apply a filtering strategy that exploits image–text similarity to refine them.

\subsection{Out-Of-Distribution Detection} \label{sec:related_work;subsec:ood_detection}

Out-of-Distribution (OOD) detection aims to identify samples during testing that come from distributions not encountered in training.
Many research~\citep{ming2022delving, wang2023clipn} leveraged vision-language foundation models, which integrate both visual and textual information to enhance the performance and flexibility of OOD detection.
On the other hand, there is a sub-task of OOD detection known as Open-Set Recognition (OSR). 
OSR addresses the issue of OOD detection at a class level and is even more closely related to UniDA.
\citet{scheirer2012toward} first articulated the concept of OSR, which has since inspired extensive research across various fields~\citep{bendale2016towards, ge2017generative, oza2019c2ae, vaze2022openset, esmaeilpour2022zero, cen2023devil}. 

Among these, the recent study aims to leverage CLIP's zero-shot capability for open-set domain adaptation~\citep{zara2023autolabel}. 
However, there are significant limitations when applying this method directly to UniDA. The number of clusters, $k$, is determined based on prior knowledge from the open-set, where the number of clusters exceeds the number of source labels. In a universal set, the decision of $k$ is more challenging as its extent is not pre-defined. Specifically, when using K-means to predict the number of classes, the elbow method is typically applied \citep{hartigan1979algorithm}. However, the elbow point is often unreliable due to a smooth score plot, leading to inaccuracies. Furthermore, we argue that texts discovered from the source domain are ineffective for detecting target private classes due to domain shifts, as the shapes of objects can differ entirely between the source and target domains, especially for abstract labels.

Unlike previous methods, our approach leverages instance-level discovery by comparing each discovered label with source labels based on image-text similarity. This eliminates the need to predefine the number of clusters, which would otherwise depend on the specific configurations of datasets. Also, we propose discovering target private classes by examining the relationship between an unbiased text representation and only the target image representation.

\subsection{Self-Training} \label{sec:related_work;subsec:self_training}


Self-training is a widely studied approach in semi-supervised learning that has recently been explored for enhancing the performance of minority classes~\citep{amini2024self}. In situations where domain shift exists, research has primarily focused on Source-free domain adaptation (SFDA) and test-time adaptation (TTA). One of the most well-known approaches is a pseudo labeling technique, which is both simple to implement and intuitive. Pseudo labeling assumes that predictions for confident samples are trustworthy. Initially, training leverages these trustworthy predictions, and as the model advances, the scope of pseudo labeling is progressively broadened. The most important aspect is to maintain class balance. Class imbalance can lead to overconfidence in certain classes, which ultimately degrades the model's performance. To mitigate this issue, various source-free domain adaptation methods, such as SHOT~\citep{liang2020we} and DIFO~\citep{tang2024source}, incorporate a diversity term to perform entropy maximization. ST-SGG~\citep{kim2024adaptive} devises a method to adapt class-adaptive thresholds based on frequency, adjusting them according to the situation in previous iterations.

Inspired by this research direction, we choose to select the top-k most confident samples for each predicted class as pseudo labels to ensure class balance. Furthermore, to maintain this balance among the pseudo labels, we define $k$ as the batch size divided by the number of predicted classes. This approach limits the selection to samples that can be easily classified. Additionally, we empirically confirm that our universal classifier, created by predicting target private labels, discriminates better than the 
$|C_s|+1$ way classifier, where $|C_s|$ is the number of source classes (see Fig.~\ref{fig:our_tsne}).

\section{Methodology} \label{sec:method}
\subsection{Problem Definition}\label{subsec:preliminaries}

In UniDA, we are given a labeled source domain $\mathcal{D}^s=\{(x^s_i,y^s_i)\}^{N_s}_{i=1}$ where $x^s_i \in \mathcal{X}^s, y^s_i \in \mathcal{Y}^s \subset \mathbb{R}^{|C_s|}$, and an unlabeled target domain $\mathcal{D}^t = \{x^t_j\}^{N_t}_{j=1}$ where $x^t_j \in \mathcal{X}^t$. We denote the ground-truth labels of the target domain as $\mathcal{Y}^t \subset \mathbb{R}^{|C_t|}$. $|C_s|$ represents the number of source labels, and $|C_t|$ represents the number of target labels. $C = C_s \cap C_t$ denotes the common label sets shared by source and target domains. $\bar{C_s} = C_s-C$ and $\bar{C_t} = C_t-C$ stand for source-private label and target private label, respectively.
Our goal is to filter the raw discovered label $R$ from the generative VLM to align the ground-truth target private classes $\Bar{C_t}$. 

\begin{figure*}[t]
    \vspace{0.75em}
    \begin{center}
        \includegraphics[width=\textwidth]{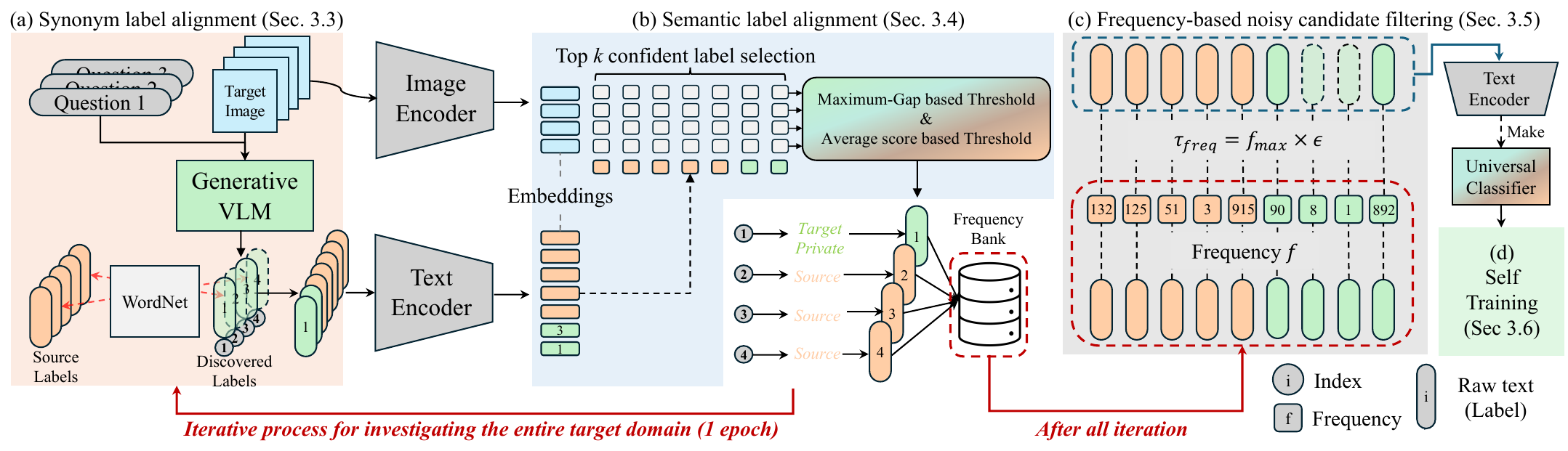}
    \end{center}
    \vspace{-1em}
    \caption{Overview of \ours. We first gather candidate target-private labels from a generative VLM, then refine them to train a universal classifier. (a) Synonym label alignment (Sec.~\ref{sec:step1}) removes source label synonyms via WordNet: (b) Semantic label alignment (Sec.~\ref{sec:instance}) re-scores image-label pairs in the embedding space and decides shared vs. target-private; (c) Frequency-based noisy candidate filtering (Sec.~\ref{sec:refine}) prunes noisy low support candidates using frequency banks; and (d) Self-training (Sec.~\ref{sec:self-training}) applies a teacher-student scheme on class-balanced top-k confident samples.}
    \label{fig:filtering}
    \vspace{0em}
\end{figure*}

\subsection{Overview} \label{sec:overview}
We verified that selecting target private labels derived from discovered labels substantially boosted performance under domain and label space shifts, as displayed in Fig.~\ref{fig:teaser}. 
We introduce TLSA, which iteratively aligns ambiguously similar labels with source labels from the discovered labels and ultimately filters them based on their occurrence patterns after all iterations.
Following the TLSA pipeline, we employ a self-training strategy to further enhance discrimination and performance. The overall framework is depicted in Fig.~\ref{fig:filtering}. \\
\noindent $\bullet$ \textit{Step 1.} 
The goal of this step is to filter the discovered labels by explicitly identifying and removing lexical synonyms of source labels using the WordNet database (Sec.~\ref{sec:synonym} and Fig.~\ref{fig:filtering}a).  \\ 
\noindent $\bullet$ \textit{Step 2.}
The objective of this step is to remove discovered labels that are semantically ambiguous or resemble the source labels by identifying a similar label set in CLIP's joint embedding space (Sec.~\ref{sec:instance} and Fig.~\ref{fig:filtering}b). \\
We iterate \textit{Step 1} and \textit{Step 2} in each cycle until the entire domain is comprehensively explored. We define the refined label set as target private candidate labels. \\
\noindent $\bullet$ \textit{Step 3.} 
To address rare instances missed in the previous phase, we removed labels based on their occurrence frequency, operating under the assumption that infrequent target private candidate labels were likely incorrect target private labels (Sec.~\ref{sec:refine} and Fig.~\ref{fig:filtering}c).\\
%
\noindent $\bullet$ \textit{Step 4.}
This step aims to fully adapt the zero-shot model to the target domain, thereby enhancing its performance beyond zero-shot inference. This enhancement is achieved by utilizing a universal classifier weighted by CLIP text embeddings (Sec.~\ref{sec:self-training} and Fig.~\ref{fig:filtering}d).

\subsection{Synonym Label Alignment} \label{sec:step1}
In this subsection, we introduce a method for generating raw labels and eliminating synonyms. 
To acquire the discovered labels, we employ a generative VLM to generate labels for unknown target images.
Furthermore, we use WordNet~\citep{miller1990introduction} to remove semantically ambiguous labels (i.e., synonyms) from the discovered labels, ensuring their alignment with the source label space, as illustrated in Fig.~\ref{fig:filtering}a.

\noindent{\textbf{Discovering Labels on Unlabeled Target Images.}} 
To ensure reliable labels for the input images, we query each target sample $x^t_i$ using the BLIP-VQA model~\citep{li2022blip,li2023blip}.
Specifically, we utilize $k=5$ prompts that are semantically equivalent yet lexically diverse, as presented in Tab.~\ref{tab:prompts}. 
The design of these prompts is inspired by the principles of prompt engineering in CLIP~\citep{radford2021learning}, employing multiple textual variants to reduce prompt sensitivity.
Consider prompts like ``What is the exact name of the object in the painting?’’ and ``Identify the object in this photo with its exact name,’’ which share identical semantics but vary in phrasing. 
The final label was determined via majority voting over $k$ responses. Responses that are overly specific (e.g., specifying a color) or do not correspond to a single valid label are filtered out using a rule-based algorithm.
Although a comprehensive sensitivity analysis across different $k$ values was not performed, preliminary observations suggested that $k=5$ struck a balance between computational efficiency and stability; hence, we consistently employed this setting throughout our experiments.

\noindent{\textbf{Synonym Label Alignment using WordNet.}}
\label{sec:synonym}
WordNet~\citep{miller1990introduction} is a widely utilized lexical database that provides taxonomic relationships among words.
Path-based similarity measures are computed by assessing the distances between two synsets within the hierarchy. 
Consequently, we employed WordNet as an essential filtering step to eliminate noisy labels in $R$.

\subsection{Semantic Label Alignment}\label{sec:instance}
This subsection elaborates on eliminating semantically ambiguous labels, such as hypernyms, hyponyms, and non-WordNet synonyms.
We employed semantic similarity to evaluate the relationship between the text representations of the source and candidate labels, based on their embedding distances from the input images. If candidate labels have a similar proximity to the image as the source labels, they are deemed identical to the corresponding source label, as displayed in Fig.~\ref{fig:filtering}b.

\noindent{\textbf{Leveraging Semantic Text-Image Similarity.}} Initially, text is converted into semantic embeddings to enable a more precise evaluation of image-text similarity. The filtered discovered labels $\Tilde{R}$ are encoded as $\mathbf{z}_{\Tilde{R}} = f_t(\tilde{R})$ using the text encoder $f_t$ of CLIP, and subsequently augmented with the source class embeddings $\mathbf{z}_{C_s} = f_t(C_s)$. We assess the similarity between the augmented embedding $\mathbf{z}_{\mathbf{t}} = [\mathbf{z}_{C_s}, \mathbf{z}_{\Tilde{R}}]$, which comprises the source labels $\mathbf{z}_{C_s}$ and the filtered discovered labels $\Tilde{R}$, and the image embedding $\mathbf{z}_v = f_v(X^t)$ produced by CLIP's visual encoder $f_v$. This evaluation determines whether the sample belongs to the target private classes or to the shared classes. The similarity score $\mathbf{S} = sim(\mathbf{z}_v, \mathbf{z}_t)$ indicates if a text is source-related, where $sim(\cdot,\cdot)$ represents a cosine similarity function. $\mathbf{S}$ is defined by Eq.~\ref{eq:similarity}.
\begin{equation}
\label{eq:similarity}
\mathbf{S} =
\begin{pmatrix}
s_{1,1} & \cdots & s_{1,|C_s|} & s_{1,|C_s|+1} & \cdots & s_{1,|\tilde{R}|} \\
\vdots  & \ddots & \vdots      & \vdots        & \ddots & \vdots            \\
s_{N,1} & \cdots & s_{N,|C_s|} & s_{N,|C_s|+1} & \cdots & s_{N,|\tilde{R}|}
\end{pmatrix}
\end{equation}
where $N$ denotes the batch size. The left portion of $\mathbf{S}$ reflects the cosine similarity between the embedding of the images and source labels, while the right portion represents the cosine similarity between images and the filtered discovered labels.

\begin{figure*}[t]
    \centering
    \includegraphics[width=0.7\textwidth]{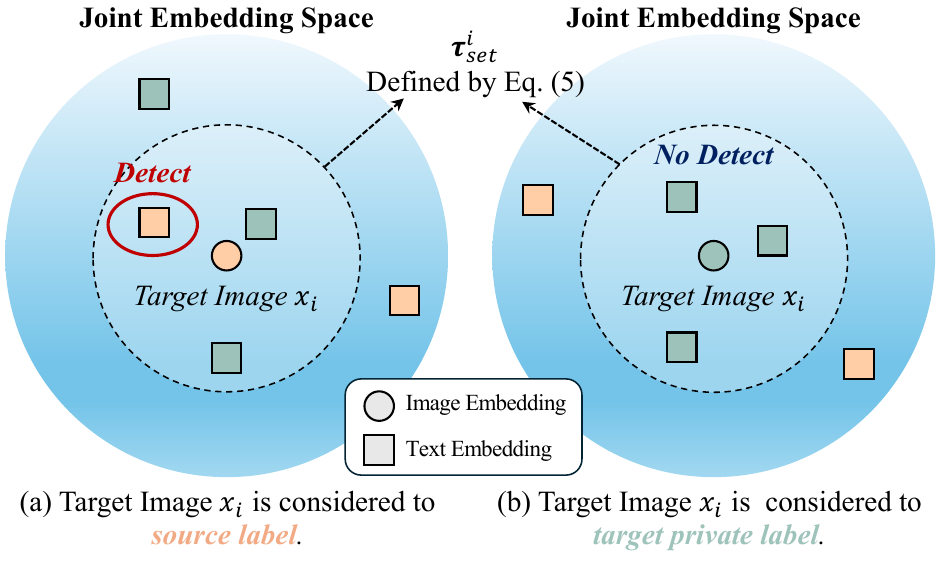}
    \caption{Illustration of semantic label alignment. In this step, we establish the relationship between discovered labels and source label predictions within the prediction set $\mathcal{C}$, as defined by Eq.~\ref{eq:thres4}, for a given image $x_i$. (a) When source labels are present in $\mathcal{C}$, the input image is classified as belonging to one of the source classes. (b) When no source labels exist in $\mathcal{C}$, the input image is categorized as a target private class. Subsequently, we maintain a frequency count of classes in the frequency bank to enable noisy candidate filtering.}
    \label{fig:align}
\end{figure*}

\noindent{\textbf{Identification of Target Private Labels from Filtered Discovered Labels.}} Using semantic similarity, we aim to determine whether an input image belongs to the source or target private classes. An overview of this procedure is presented in Fig.~\ref{fig:align}. When determining whether a label originates from a source-private or target, we found that naive alignment methods—such as top-1 prediction-based alignment and margin-based alignment—performed poorly in our experiments (see Tab.~\ref{tab:IAT}). For instance, when `backpack' is the source label and `bag' is the discovered label (Fig.~\ref{fig:misclassified_example}c), `bag' and `backpack' are not synonyms in WordNet; however, within the Office31 context, they effectively represent the same class.
Treating them as distinct categories leads to numerous `backpack' samples being misclassified as `bag', resulting in artificial label splits.
In such cases, top-1 predictions alternate between `backpack' and `bag' due to uncontrollable image conditions, without signifying any meaningful semantic difference.
These semantically ambiguous labels might be included among certain target private labels. To address this, the margin value between the top-1 and top-2 predicted probabilities can evaluate the distance between source labels and the filtered discovered labels. However, this method relies on a predefined fixed threshold to assess reliability, ignoring class similarities that often vary adaptively across datasets and domains.

To overcome this limitation, we adaptively align the filtered discovered labels with source labels, removing the need for a fixed threshold.
By analyzing the similarity matrix $S$, we first eliminate noisy labels and identify only reliable predictions within the top $k$ predictions. We define a prediction set $\mathcal{C}$ as the collection of refined and filtered predictions from the top $k$ predictions using the two proposed criteria.

\noindent $\bullet$ \textit{Criterion 1: Maximum-gap-based thresholding.} 
A prominent gap in similarity scores was identified among the top $k$ rankings. The labels preceding this gap accurately correspond to multiple image predictions. Utilizing this criterion, we establish the meaningful prediction set $\mathcal{C}$. The threshold is defined as follows:
\begin{equation} \label{eq:thres2}
J = \underset{j \in \{1, \dots, k-1\}}{\arg\max} \left( s_{i, j-1} - s_{i, j} \right),
\end{equation}
\begin{equation} \label{eq:thres1}
\tau^i_{\text{gap}} = s_{i, J},
\end{equation}
where \textit{i} represents the image index, \textit{j} the ranking position, and $J$ denotes the point at which a significant gap occurs.

\noindent $\bullet$ \textit{Criterion 2: Average score-based thresholding.}  We define the prediction set as comprising scores that exceed the average score of the top $k$ predictions. The threshold is defined as the average of the top $k$ scores:
\begin{equation} \label{eq:thres3}
    \begin{aligned}
    \tau^i_{avg} = \frac{\sum_{j=1}^k{s_{i,j}}}{k}. \\
    \end{aligned}
\end{equation}

Empirically, we maximize the prediction set size by selecting the lower of the two criteria (i.e., $\tau^i_{gap}$ and $\tau^i_{avg}$). This approach enhances the inclusion of source labels within the label set, thereby reducing the likelihood of ambiguous labels being identified as private target candidates. In other words, by selecting the smaller threshold between $\tau_{\text{gap}}$ and $\tau_{\text{avg}}$, the criterion for classifying a sample as target private becomes more stringent. The threshold $\tau_{\text{set}}$ and prediction set $\mathcal{C}$ for each instance are formulated as follows:

\begin{equation} \label{eq:thres4}
    \tau^i_{set} = \min(\tau^i_{gap}, \tau^i_{avg}), 
\end{equation}

\begin{equation} \label{eq:thres5}
    \mathcal{C}^i = \{\mathbf{t}_{i,j} |j \in I^i_k \land s_{i,j} > \tau_{set}^i\},
\end{equation}
where $\mathbf{t}$ represents the text comprising extensions of source labels and filtered discovered labels, and $I^i_k$ denotes the top-$k$ indices for the $i^{\text{th}}$ sample. Utilizing the selected prediction set $\mathcal{C}$, we perform a source label presence check to achieve semantic label alignment.
Specifically, if a source label is present in $\mathcal{C}^i$ (Fig.~\ref{fig:align}a), sample $i$ is classified as a source sample. When it is absent from the prediction set $\mathcal{C}$, sample $i$ is identified as a target private sample (Fig.~\ref{fig:align}b). 

\subsection{Frequency-based Noisy Candidate Filtering} \label{sec:refine} 
Noisy and incorrectly predicted private target candidate labels may still persist. To address this, we developed a method that filters these noisy labels by analyzing their frequencies, as shown in Fig.~\ref{fig:filtering}c.
We hypothesized that low-frequency classes are likely noisy labels in the target domain. To precisely measure label frequency, we employed a frequency bank. 
For samples classified as source classes, the top-1 prediction $p_i$ is stored in the frequency bank as a reference standard (Fig.~\ref{fig:align}a). Similarly, for samples classified as target private classes, the discovered label $r_i$ is added to the frequency bank even if $r_i \notin \mathcal{C}^i $(Fig.~\ref{fig:align}b). The rationale for storing $r_i$, the label identified by BLIP, instead of $p_i$, the closest label in the CLIP space, is detailed in Appendix~\ref{sec:blip}.
The frequency bank is defined as $F = \{c : f_c\}$, where $f_c$ represents the frequency of class $c$ after semantic label alignment, including both source and filtered discovered labels.
Considering the presence of noisy labels in the frequency bank, we used the count of the most frequent class rather than the median or mean of the label frequency distribution. We set a threshold $\tau_{\text{freq}}$ as the fixed number of samples in the most frequent class. This scaling factor, denoted by $\epsilon$, is a dataset-agnostic parameter.
Finally, the predicted target private labels $\mathcal{C}_p$ are derived from the raw discovered labels $R$ as follows:

\begin{equation} \label{eq:thres6}
    \tau_{freq} = \epsilon \times \max(F),
\end{equation}
\begin{equation} \label{eq:thres7}
    C_p = \{c|f_c > \tau_{freq} \land c \notin C_s\}.
\end{equation}
We employ an empirically determined value of $\epsilon=0.01$ across all datasets, as presented in Tab.~\ref{tab:k}.

By integrating complementary components of the pipeline—lexical filtering validated through WordNet, semantic filtering within a joint image–text embedding space, and a frequency-based filtering step that removes residual noisy discovered labels depending on dataset scale—our method achieves strong complementarity among the three, as evidenced in Tab.~\ref{tab:filtering}.

\subsection{Self-Training with Universal Classifier}
\label{sec:self-training}
In UniDA, the prior typically uses randomly initialized prototype-based classifiers~\citep{li2021domain, chang2022unified, lai2023memory}. However, this method complicates determining the number of target clusters, which corresponds to the number of private target classes. In contrast, we utilize a universal classifier that automatically determines the number of target private classes, as depicted in Fig.~\ref{fig:filtering}d. Furthermore, we applied self-training, a prevalent technique in source-free domain adaptation \citep{liang2020we, zhang2023rethinking, karim2023c, tang2024source}. This method reduces bias toward the source domain, offering an advantage over approaches that depend on visual-space domain alignment and source-supervised learning. Additionally, we selected reliable pseudo labels based on the top-$k$ confidence samples per class and employed a student-teacher framework with EMA updates.

\noindent\textbf{Creating a Universal Classifier.}  
Our inference leverages the weights of the \textit{universal classifier}, denoted by $\mathbf{z}_{C_s}$ and $\mathbf{z}_{C_p}$, which are produced by inputting $C_p$ into the text encoder of CLIP. For post-processing, instances classified as $C_s$ were retained unchanged, while those labeled as $C_p$ were adjusted to $|C_s| + 1$, ensuring a threshold-free method.

\noindent\textbf{Reliable Pseudo Label Selection.}  To leverage pseudo-labels, we trained a student model to enhance the teacher model's performance using delayed updates. However, naive pseudo-labeling tends to select only easily classifiable classes, leading to overfitting of these simpler categories.
To address this, we introduce a balanced pseudo-label selection strategy that improves the generalization of the predictions of the pretrained model.
Rather than requiring a large dataset, our approach prioritizes maximum diversity by selecting the top-k (\%) most confident pseudo-labels per class, as proposed by \citet{zara2023autolabel}. Nonetheless, even with a top-k (\%) threshold, more prevalent classes are disproportionately selected, which can degrade performance. To mitigate this, we define a balanced $k$ value to select only the top-$k$ most confident samples for each class. Specifically, $k$ is calculated as $k = \text{min}\{n_{c_i}, N/n_{c}\}$, where $n_c$ is the number of unique predicted classes and $n_{c_i}$ is the number of predictions for class $c_i$.

\noindent\textbf{Robust Self-Training.}  To ensure robust self-training, we apply Exponential Moving Average (EMA) updates to the teacher model, defined by $\theta_t = (1-\alpha)\theta_{s} + \alpha\theta_{t}$. Here, $\theta_t$ and $\theta_s$ represent the parameters of the teacher and student models, respectively, effectively mitigating confirmation bias. The student model updates with each iteration, whereas the teacher model updates once per epoch, as defined by the preceding equation. This configuration is crucial for the performance of the model (Sec.~\ref{sec:results}). Additionally, we implemented adapter tuning to limit further tuning capabilities. We then perform self-training using the following loss function: Let $M$ represent the set of samples employed as pseudo-labels, $\hat{y}^T_{t, i}$denotes the output of the teacher model, and $\hat{y}^S_{t, i}$represents the output of the student model.

\begin{equation}
\mathcal{L}_{self} = -\sum_{i \in M}{\hat{y}^T_{t,i}\log{\hat{y}^S_{t,i}}},
\label{eq:loss}
\end{equation}

By utilizing a frozen classifier, a common approach in SFDA~\citep{liang2020we}, adapter tuning mitigates error accumulation by addressing the delayed conversion rate.

\section{Experiments} \label{sec:exp}
This section outlines the dataset, implementation, experimental settings, and results.

\noindent{\textbf{Datasets.}}
We utilize four benchmark datasets for visual recognition domain adaptation: Office31 (O), Office-Home (OH), VisDA (VD), and DomainNet (DN).
Office31~\citep{saenko2010adapting} is a small-scale benchmark dataset comprising images from 31 categories distributed across three domains: \textbf{A}mazon (2,817 images), \textbf{D}slr (498 images), and \textbf{W}ebcam (795 images). Office-Home~\citep{venkateswara2017deep} is a benchmark dataset designed for domain adaptation, featuring images from 65 categories across four domains: \textbf{Ar}t (9,655 images), \textbf{Cl}ipart (9,516 images), \textbf{Pr}oduct (15,688 images), and \textbf{R}eal\textbf{w}orld (14,840 images). VisDA \citep{peng2017visda} is a benchmark dataset for domain adaptation, containing images from 12 labels distributed across two domains: \textbf{Syn}thetic (152,397 images) and \textbf{Real} (55,088 images). DomainNet \citep{peng2019moment} is a large-scale benchmark dataset for domain adaptation, notable for its extensive size. It includes images from 345 labels distributed across six domains: \textbf{Clipart}, \textbf{Infograph}, \textbf{Painting}, \textbf{QuickDraw}, \textbf{Real}, and \textbf{Sketch}. The sheer scale of DomainNet makes it a particularly challenging and comprehensive dataset for evaluating domain adaptation methods. We conducted experiments using only three domains from DomainNet that are commonly employed in existing methods for UniDA - Painting, Sketch, and Real. 

\begin{table}[t]
    \centering
    \caption{Split-setting of our experiments. Details of experimental split settings indicating the counts of source-private and shared classes (Source-Private / Shared) in each dataset under different adaptation scenarios (open-partial, open, closed, partial).}
    \begin{tabular}{c cccc} \toprule
         \multirow{2}{*}{Datasets} & \multicolumn{4}{c}{Split settings} \\ \cline{2-5}
                                & open-partial & open & closed & partial \\
         \midrule
         Office31   & 10/10 & 0/10 & 0/31 & 21/10 \\
         Office-Home & 5/10 & 0/15 & 0/65 & 40/25 \\
         VisDA    & 3/6 & 0/6 & 0/12 & 6/6 \\
         DomainNet & 50/150 & 0/150 & 0/345 & 195/150 \\
    \bottomrule
    \end{tabular}
    \label{tab:split-settings}
\end{table}

\noindent{\textbf{Baselines.}}
This section lists the baseline methods we used for comparison. Most of the performance results are referenced from GLC++\citep{qu2024glc++}, MLNet\citep{lu2024mlnet}, MemSPM~\citep{lai2023memory}, and LEAD~\citep{qu2024lead}. The lists below are the ones that link to their respective references: Source-only(SO),
DANCE$^{\hypertarget{cite:a}{\text{a}}}$ \citep{saito2020universal},
OVANet$^{\hypertarget{cite:b}{\text{b}}}$ \citep{saito2021ovanet},
UniOT$^{\hypertarget{cite:c}{\text{d}}}$ \citep{chang2022unified},
WiSE-FT$^{\hypertarget{cite:d}{\text{e}}}$ \citep{wortsman2022robust},
CLIP-Distill$^{\hypertarget{cite:e}{\text{e}}}$ \citep{deng2023universal},
GLC$^{\hypertarget{cite:f}{\text{f}}}$ \citep{sanqing2023GLC},
MemSPM$^{\hypertarget{cite:g}{\text{g}}}$ \citep{lai2023memory},
UAN$^{\hypertarget{cite:h}{\text{h}}}$ \citep{you2019universal},
CMU$^{\hypertarget{cite:i}{\text{i}}}$ \citep{fu2020learning},
DCC$^{\hypertarget{cite:j}{\text{j}}}$ \citep{li2021domain},
GATE$^{\hypertarget{cite:k}{\text{k}}}$ \citep{chen2022geometric},
MLNet$^{\hypertarget{cite:l}{\text{l}}}$ \citep{lu2024mlnet},
LEAD$^{\hypertarget{cite:m}{\text{m}}}$ \citep{qu2024lead}.

\noindent{\textbf{Implementation Details.}}
We use CLIP~\citep{radford2021learning} backbone with ViT-B/16~\citep{dosovitskiy2020image} as our image encoder. For the generative VLM, we utilize BLIP~\citep{li2022blip}, a model pretrained on VQAv2~\citep{balanced_vqa_v2} within the LAVIS library\footnote{\url{https://github.com/salesforce/LAVIS}}. This model incorporates a visual encoder based on ViT-Base~\citep{dosovitskiy2020image} and employs a text encoder and decoder that adhere to the BERT~\citep{devlin2018bert} architecture. Most of the baseline code and pipelines are based on the uniood repository\footnote{ \url{https://github.com/szubing/uniood}}. 

Our experimental class split settings follow the standard protocol used by many existing UniDA approaches~\citep{saito2020universal, saito2021ovanet, chang2022unified, lu2024mlnet, qu2024lead, qu2024glc++, deng2023universal}, as detailed in Tab.~\ref{tab:split-settings}.

During the alignment phase, to expedite filtering, we conducted experiments with a batch size of 128. These experiments were executed on a single H100 GPU, consuming approximately 15GB of memory during the filtering process. As the source domain was not trained separately, the overall experimental time was significantly reduced.

In the self-training phase, we conducted experiments with a batch size of 64, a learning rate of 0.01, and an adapter with bottleneck dimensions of \textit{r}=64, adhering to the default settings of AdaptFormer~\citep{chen2022adaptformer}. All baseline methods were implemented with identical iterations, batch sizes, and learning rates using a frozen backbone, as described in~\citet{deng2023universal}. The iterations were set to 2,500; 5,000; and 10,000, depending on the task setting, as presented in Tab.~\ref{tab:training-iterations}, following \citep{deng2023universal}. Our methods required approximately 30 minutes per domain, allowing us to complete all experiments on the entire dataset within a day.

\begin{table}[t]
    \centering
    \caption{Training iterations on different split settings. We follow the same setting in~\citep{deng2023universal}. Since we use a batch size 64, which is twice the original value, we reduce the number of iterations accordingly by half. }
    \begin{tabular}{c cccc} \toprule
         \multirow{2}{*}{Datasets} & \multicolumn{4}{c}{Split settings} \\ \cline{2-5}
                                & open-partial & open & closed & partial \\
         \midrule
         Office31   & 2500 & 2500 & 5000 & 5000 \\
         Office-Home & 2500 & 2500 & 5000 & 5000 \\
         VisDA    & 5000 & 5000 & 10000 & 10000 \\
         DomainNet & 5000 & 5000 & 10000 & 10000 \\
    \bottomrule
    \end{tabular}
    \label{tab:training-iterations}
\end{table}

\noindent{\textbf{Evaluation Metric.}}
We use the existing two evaluation metrics for UniDA in addition to classification accuracy. H-score \citep{fu2020learning} is the harmonic mean of the instance accuracy on common class $a_C$ and accuracy on the single private class $a_{\bar{C_t}}$. This metric effectively captures the trade-off between known-class accuracy and out-of-class detection accuracy.  Moreover, we also use the H$^3$-score. The H$^3$-score is first proposed by UniOT \citep{chang2022unified} and extends the H-Score by incorporating the Normalized Mutual Information (NMI), a well-known measure that indicates how well the target-private clusters are formed. In the PDA and CDA settings where target-private instances do not exist, the average class accuracy is used in place of the H-score. Similarly, the $H^3$ score is also substituted with this measure.

\section{Results} \label{sec:results}

\begin{table*}[t]
    \renewcommand{\arraystretch}{1.0}
    \centering
    \caption{H-score (\%) and H$^3$-score (\%) comparisons between existing methods and the proposed method using the CLIP ViT-B/16 backbone across four UniDA settings (open-partial, open, closed, partial). An asterisk (*) represents results for some methods referenced from the previous paper~\citep{lai2023memory}. A dash (-) represents that any metrics not reported in the official papers.}
    \begin{adjustbox}{width=1.0\textwidth,center}
    \begin{threeparttable}
    \begin{tabular}{c  c c c c   c c c c   c c c c   c c c c   c}
    \toprule
    \multirow{2}{*}{Methods} & \multicolumn{4}{c}{Office31} & \multicolumn{4}{c}{Office-Home} & \multicolumn{4}{c}{VisDA} & \multicolumn{4}{c}{DomainNet} & \multirow{2}{*}{Avg}\\ \cmidrule(lr){2-5} \cmidrule(lr){6-9} \cmidrule(lr){10-13} \cmidrule(lr){14-17}
                             & (10/10) & (10/0) & (10/21) & (31/0) & (10/5) & (15/0) & (25/40) & (65/0) & (6/3) & (6/0) & (6/6) & (12/0) & (150/50) & (150/0) & (150/195) & (345/0) & \\
    \midrule
    \multicolumn{18}{c}{H-score} \\
    \midrule
    SO	&	85.8	&	86.8	&	83.8	&	70.8
    &	79.1	&	76.5	&	51.8	&	49.6	
    &	67.8	&	71.4	&	49.1	&	51.1	
    &	55.8	&	59.7	&	32.9	&	34.8	&	62.9	\\
    DANCE   &	91.7	&	93.2	&	73.8	&		67.7
    &	83.6	&	78.3	&	46.9	&	49.4	
    &	66.6	&	76.6	&	33.7	&   46.3	
    &	55.1	&	59.4	&	30.0	&	34.1	&	61.6	\\
    OVANet	&	89.0	&	89.3	&	84.0	&	65.9	
    &	80.5	&	76.3	&	58.8	&	56.1	
    &	56.6	&	44.4	&	36.6	&   40.9	
    &	62.0	&	64.4	&	45.0	&	46.2	&	62.2	\\
    UniOT	&	82.3	&	93.9	&	31.5	&	54.2	
    &	86.3	&	81.8	&	31.8	&	46.4	
    &	58.2	&	77.3	&	33.6	&	54.2	
    &	65.6    &	69.4	&	40.9	&  	51.5 &	59.8	\\
    WiSE-FT	&	70.4	&	88.3	&	38.5	&	30.9	
    &	72.8	&	65.8	&	11.7	&	8.8
    &	54.7	&	67.8	&	19.1	&	22.2	
    &2.4	&6.0& 0.19 &	0.2	 & 35.0	\\
    CLIP-Distill	&	83.5	&	84.3	&	85.6	&	73.7	
    &	84.7	&	81.3	&	74.4	&	69.0	
    &	82.8	&	82.2	&	78.7	&		71.1
    &	64.8    &	66.7	&	56.4	&	51.3	&	74.4	\\
    GLC*	&	84.8	&	90.1	&	91.5	&	-	
    &	82.6	&	86.2	&	75.4	&	-	
    &	80.3	&  81.6	&	86.2	&	-	
    &	46.8	&	-	&	- &	-	&	-	\\ 
    MemSPM*	&	88.5	&	\textbf{95.6}	&	94.4	&	-	&	84.2	&	87.9	&	74.5	&	-	
    &	79.3	&  79.7	&	\textbf{87.9}	&	-	
    &	56.7	&	-	&	-	&	-	&	-	\\
    \hline
    \rowcolor{gray!10} \ours &	89.9	&	92.6	&	96.1	&	80.5	
    &	91.3	&	89.2	&	82.0	&	79.0
    &	86.3	&	83.9	&	81.8	&	72.6	&	
\textbf{75.4}	&	\textbf{76.9}	&	70.1	&	\textbf{68.9}	&	82.3	\\
    \rowcolor{gray!10} \ours+ST &	 \textbf{91.9} & 95.2 & \textbf{96.9} & \textbf{82.9} & \textbf{93.5} & \textbf{91.4} & \textbf{85.3} & \textbf{81.8} & \textbf{87.6} & \textbf{86.0} & 83.0 & \textbf{75.6} & 74.3 & 74.9 & \textbf{74.4} & 67.9&	\textbf{83.9}	\\
    \midrule
    \multicolumn{18}{c}{H$^3$-score} \\
    \midrule
    SO	&	83.4	&	85.8	&	83.8	&	70.8	
    &	77.2	&	75.2	&	51.8	&	49.6	
    &	72.6	&	73.4	&	49.1	&	51.1	
    &	59.6	&	62.5	&	32.9	&	34.8	 &	63.3	\\
    DANCE	&	87.0	&	89.8	&	73.8	&	67.7	
    &	80.5	&	77.7	&	46.9	&	49.4
    &	71.7	&	76.9	&	33.7	&  46.3
    &	59.1	&	62.2    &	30.0	&	34.1	&	61.6		\\
    OVANet	&	85.2	&	87.3	&	84.0	& 	65.9
    &	78.0	&	75.2	&	58.8	&	56.1	
    &	63.6	&	51.8	& 36.6      &	40.9	
    &	64.1	&	65.8    &	45.0	&	46.2	&	62.8	\\
    UniOT	&	80.7	&	89.8	&	31.5	&	54.2	
    &	82.3	&	79.7	&	31.8	&	46.4	
    &   61.3   &	75.8	&	34.0   &	53.6
    &	66.6	&	69.0	&	40.9	&	51.5	&	59.3 \\
    WiSE-FT	&	72.5	&	86.6	&	38.5	&	30.9	
    &	72.0	&	68.1	& 	11.7    &	8.8	
    &	71.5	&	78.3	&	19.1	&	22.2	
    &	3.6	&	8.6	&	0.2	& 0.2		&	37.1	\\
    CLIP-Distill	&	81.7	&	83.8	&	85.6	&	73.7
    &	80.2	&	78.3	&	74.4	&	69.0
    &	81.8	&	86.0	&	78.7	&	71.1	
    &	66.1	&	67.4	&	56.4	&	51.3	&	74.1	\\
    \hline
    \rowcolor{gray!10} \ours &	\textbf{85.8}	&	89.1	&	96.1	&	80.5	
    &	84.8	&	82.9	&	82.0	&	79.0	
    &	85.7 &	81.7&	81.8	&	72.6	
    &	73.0	&	73.9	&	69.6	&	\textbf{68.2}	&	80.5	\\
     \rowcolor{gray!10} \ours+ST &	85.3 & \textbf{89.9} & \textbf{96.9} & \textbf{82.9} & \textbf{90.2} & \textbf{89.0} & \textbf{85.3} &\textbf{ 81.8} &\textbf{ 88.4 }& \textbf{85.0 }& \textbf{83.0} & \textbf{75.6} & \textbf{75.4} & \textbf{75.5} & \textbf{74.4} & 67.9 & \textbf{83.2}	\\
    \bottomrule
    \end{tabular}
    \end{threeparttable}
\end{adjustbox}
\label{table:results-all-clip}
\end{table*}

\begin{table*}[htbp]
  \centering
  \caption{H-score (\%) comparison in OPDA scenarios on the Office-Home dataset. Results for some methods are referenced from GLC++~\citep{qu2024glc++} and LEAD~\citep{qu2024lead}. Here, SF indicates source data-free methods, and SM denotes approaches that use a source model. We compare TLSA with both source-free methods and those using source models. An asterisk (*) indicates reproduced results.}
  \vspace{-0.05in}
  \addtolength{\tabcolsep}{-3.0pt}
  \resizebox{0.99\textwidth}{!}{
   \begin{threeparttable}
    \begin{tabular}{lcccccccccccccccc}
    \toprule
    \multirow{2}[4]{*}{Backbone} & \multirow{2}[4]{*}{Methods} & \multirow{2}[4]{*}{SF} & \multirow{2}[4]{*}{SM} & \multicolumn{12}{c}{Office-Home} & \multirow{2}[4]{*}{Avg} \\
    \cmidrule(lr){5-16} 
    & & & & \textbf{Ar2Cl} & \textbf{Ar2Pr} & \textbf{Ar2Re} & \textbf{Cl2Ar} & \textbf{Cl2Pr} & \textbf{Cl2Re} & \textbf{Pr2Ar} & \textbf{Pr2Cl} & \textbf{Pr2Re} & \textbf{Re2Ar} & \textbf{Re2Cl} & \textbf{Re2Pr} \\
    \midrule
    \multirow{11}{*}{ImageNet} 
    & SO& \cmark & \cmark & 47.3  & 71.6  & 81.9  & 51.5   & 57.2  & 69.4  & 56.0  & 40.3  & 76.6 & 61.4  & 44.2 & 73.5 & 60.9 \\
    & UAN\hyperlink{cite:h}{\tnote{h}}  & \xmark & - &  51.6  & 51.7  & 54.3  & 61.7  & 57.6  & 61.9  & 50.4  & 47.6  & 61.5  & 62.9  & 52.6  & 65.2  & 56.6  \\
    & CMU\hyperlink{cite:i}{\tnote{i}} & \xmark &  -  & 56.0 & 56.9  & 59.2  & 67.0  & 64.3  & 67.8  & 54.7  & 51.1  & 66.4  & 68.2  & 57.9  & 69.7  & 61.6  \\
    & DANCE\hyperlink{cite:a}{\tnote{a}} & \xmark &  - & 61.0  & 60.4  & 64.9 & 65.7  & 58.8  & 61.8  & 73.1 & 61.2  & 66.6  & 67.7  & 62.4  & 63.7 & 63.9 \\
    & DCC\hyperlink{cite:j}{\tnote{j}} & \xmark &  - &58.0  & 54.1  & 58.0  & 74.6  & 70.6  & 77.5  & 64.3 & 73.6  & 74.9  & 81.0  & 75.1  & 80.4 & 70.2 \\
    & GATE\hyperlink{cite:k}{\tnote{k}} & \xmark &  - & 63.8  & 75.9  & 81.4  & 74.0  & 72.1  & 79.8  & 74.7  & 70.3  & 82.7  & 79.1  & 71.5  & 81.7 & 75.6 \\
    & OVANet\hyperlink{cite:b}{\tnote{b}} & \xmark &  - & 61.7  & 77.3  & 80.6  & 68.9  & 68.7  & 61.9  & 50.4  & 47.6  & 61.5  & 62.9  & 52.6  & 65.2 & 56.6 \\
    & UniOT\hyperlink{cite:c}{\tnote{c}} & \xmark & - & 67.3  &  80.5  & 86.0  & 73.5  & 77.3  & 84.3  & 75.5   & 63.3  & 86.0  & 77.8  & 65.4  & 81.9 & 76.6 \\
    & GLC\hyperlink{cite:f}{\tnote{f}} & \cmark &   \cmark &64.3  & 78.2  & 89.8  & 63.1  & 81.7  & 89.1  & 77.6   & 54.2  & 88.9  & 80.7  & 54.2  & 85.9 & 75.6 \\
    & MLNet\hyperlink{cite:l}{\tnote{l}} & \xmark &  - & 68.2  & 83.8  & 85.0   & 73.6  & 78.2  & 82.2  & 75.2  & 64.7  & 85.1  & 78.8  & 69.9  & 83.9 & 77.4 \\
    & LEAD\hyperlink{cite:m}{\tnote{m}} & \cmark & - & 62.7  & 78.1  & 86.4  & 70.6  & 76.3  & 83.4  & 75.3  & 60.6   & 86.2  & 75.4  & 60.7  & 83.7 & 75.0 \\
    \midrule
    \multirow{12}{*}{CLIP} 
    & SO* & \cmark & \cmark & 69.8  & 80.3  & 83.8  & 80.5  & 81.5  & 85.3  & 77.5  & 71.2  & 84.7  & 80.5  & 76.0  & 77.6 & 79.1 \\
    & DANCE*\hyperlink{cite:a}{\tnote{a}} & \xmark & - & 78.0  & 86.4  & 87.1  & 81.1  & 85.2  & 85.6  & 82.4  & 79.9  & 90.5  & 83.6  & 79.6  & 84.1 & 83.6 \\
    & OVANet*\hyperlink{cite:b}{\tnote{b}} & \xmark & - & 72.7  & 88.0  & 86.4  & 80.6  & 84.0  & 85.3  & 75.2  & 63.6  & 89.1  & 82.0  & 74.4  & 85.0 & 80.5 \\
    & UniOT*\hyperlink{cite:c}{\tnote{c}} & \xmark & - & 85.5  & 89.8  & 91.1  & 79.9  & 87.3  & 89.3  & 80.8  & 83.5  & 88.9 & 83.5  & 84.3  & 91.6 & 86.3 \\
    & WiSE-FT*\hyperlink{cite:d}{\tnote{d}} & \cmark & \cmark & 59.4  & 86.9  & 87.0  & 57.6  & 80.3  & 80.9 & 62.7  & 50.2  & 85.1  & 71.0  & 64.7  & 88.4 & 72.8 \\
    & GLC\hyperlink{cite:f}{\tnote{f}} & \cmark & \cmark & 79.4  & 88.9  & 90.8  & 76.3  & 84.7  & 89.0  & 71.5  & 72.9  & 85.7  & 78.2  & 79.4  & 90.0 & 82.6 \\
    & DCC\hyperlink{cite:j}{\tnote{j}} & \xmark & - & 62.6  & 88.7  & 87.4  & 63.3  & 68.5  & 79.3  & 67.9  & 63.8  & 82.4  & 70.7  & 69.8  & 87.5 & 74.4 \\
    & MemSPM\hyperlink{cite:g}{\tnote{g}} & \xmark & - & 78.1  & 90.3  & 90.7  & 81.9  & 90.5  & 88.3  & 79.2  & 77.4  & 87.8  & 78.8  & 76.2  & 91.6 & 84.2 \\
    & CLIP-Distill*\hyperlink{cite:e}{\tnote{e}} & \cmark & \cmark & 81.4  & 79.6  & 86.1  & 87.9  & 80.3  & 86.7  & 87.0  & 81.3  & 80.6  & 85.2  & 82.0  & 88.9 & 84.3 \\
    
    \rowcolor{gray!10} &\textbf{TLSA} & \cmark & \textcolor{blue}{\xmark} &  87.4  & 95.2  & 95.8  & 86.8  & 95.2  & 95.8  & 86.8  & 87.4  & 95.8  & 86.8  & 87.4  & 95.2 & 91.3\\
    \rowcolor{gray!10} & \textbf{TLSA+ST} & \cmark  &\textcolor{blue}{\xmark} & \textbf{90.2} & \textbf{96.3} & \textbf{96.7} &\textbf{ 90.6} & \textbf{96.3} & \textbf{96.7} & \textbf{90.6} &\textbf{ 90.2 }& \textbf{96.7} & \textbf{90.6} & \textbf{90.2} & \textbf{96.3} & \textbf{93.4} \\
    \bottomrule
    \end{tabular}%
    \end{threeparttable}
  }
  \label{tab:officehome}%
\end{table*}

\begin{table*}[htbp]
  \centering
  \caption{H-score (\%) comparison in OPDA scenarios on the Office31, DomainNet, and VisDA datasets, respectively. MLNet~\citep{lu2024mlnet} does not provide experimental results for DomainNet, any metrics not reported in their paper are marked as “-”. An asterisk (*) indicates reproduced results.}
  \vspace{-0.05in}
  \addtolength{\tabcolsep}{-2.0pt}
  \resizebox{0.99\textwidth}{!}{
  \begin{threeparttable}
    \begin{tabular}{lccc ccccccc ccccccc c}
    \toprule
    \multirow{2}[4]{*}{Backbone} & \multirow{2}[4]{*}{Methods} & \multirow{2}[4]{*}{SF} & \multirow{2}[4]{*}{SM} & \multicolumn{7}{c}{Office31}              & \multicolumn{7}{c}{DomainNet} & \multicolumn{1}{c}{VisDA} \\
    \cmidrule(lr){5-11} \cmidrule(lr){12-18} \cmidrule(lr){19-19}
    &     &     &    & A2D   & A2W   & D2A   & D2W   & W2A   & W2D   & \textbf{Avg}   & P2R & P2S & R2P & R2S & S2P & S2R & \textbf{Avg} & S2R  \\
    \midrule
    \multirow{11}{*}{ImageNet}
    & SO   & \cmark & \cmark & 70.9  & 63.2  & 39.6  & 77.3  & 52.2  & 86.4  & 64.9   & 57.3      & 38.2      & 47.8      & 38.4      & 32.2      & 48.2      & 43.7 & 25.7 \\
     & UAN\hyperlink{cite:h}{\tnote{h}}   & \xmark  & - & 59.7  & 58.6  & 60.1  & 70.6  & 60.3  & 71.4  & 63.5  & 41.9 & 39.1 & 43.6 & 38.7 & 38.9 & 43.7 & 41.0 & 34.8 \\
    & CMU\hyperlink{cite:i}{\tnote{i}}   & \xmark   & -  & 68.1  & 67.3  & 71.4  & 79.3  & 72.2  & 80.4  & 73.1  & 50.8 & 45.1 & 52.2 & 45.6 & 44.8 & 51.0 & 48.3 & 32.9 \\
    & DANCE\hyperlink{cite:a}{\tnote{a}} & \xmark  & - &  78.6  & 71.5  & 79.9  & 91.4  & 72.2  & 87.9  & 80.3  & 21.0 & 37.0 & 47.3 & 46.7 & 27.7 & 21.0 & 33.5 & 42.8 \\
    & DCC\hyperlink{cite:j}{\tnote{j}}   & \xmark   & - & 88.5  & 78.5  & 70.2  & 79.3  & 75.9  & 88.6  & 80.2  & 56.9 & 43.7 & 50.3 & 43.3 & 44.9 & 56.2 & 49.2 & 43.0 \\
    & GATE\hyperlink{cite:k}{\tnote{k}}  & \xmark   & - & 87.7  & 81.6  & 84.2  & 94.8  & 83.4  & 94.1  & 87.6  & 57.4 & 48.7 & 52.8 & 47.6 & 49.5 & 56.3 & 52.1 & 56.4 \\
    & OVANet\hyperlink{cite:b}{\tnote{b}} & \xmark  & - & 85.8 &  79.4 & 80.1 & 95.4 & 84.0 &94.3 & 86.5 & 56.0 & 47.1 & 51.7 &44.9 &47.4 &57.2 &50.7 & 53.1 \\
    & UniOT\hyperlink{cite:c}{\tnote{c}}  & \xmark  & -  &83.7 &85.3& 71.4 & 91.2 & 70.9  & 90.8 & 82.2 & 59.3 & 47.8 & 51.8 & 46.8 & 48.3 & 58.3 & 52.0 & 57.3 \\
    & GLC\hyperlink{cite:f}{\tnote{f}}   & \cmark   & \cmark & 81.5 & 84.5 & 89.8 & 90.4 & 88.4 & 92.3 & 87.8 & 63.3 & 50.5 & 54.9 & 50.9 & 49.6 & 61.3 & 55.1 & 73.1 \\
    & MLNet\hyperlink{cite:l}{\tnote{l}}   & \xmark  & - &83.9 & 78.5 & 81.3 & 95.4 & 83.2 & 96.4 & 86.5 & - & - & - & - & - & - & - & 69.9 \\
    & LEAD\hyperlink{cite:m}{\tnote{m}}   & \cmark  & \cmark & 85.4 & 85.0 & 86.3 & 90.9 & 86.2 & 93.1 & 87.8  & 59.9 & 46.1 & 51.3 & 45.0 & 45.9 & 56.3 & 50.8 & 76.6 \\
    \midrule
    \multirow{12}{*}{CLIP}
    & SO*   & \cmark   & \cmark & 84.4  & 79.0 & 88.1  & 93.3  & 80.2 & 89.8  &  85.8  & 64.2 & 50.2 & 53.1 & 52.1 & 48.3 & 67.1 & 55.8 &   67.8 \\
    & DANCE*\hyperlink{cite:a}{\tnote{a}}   & \xmark  & - & 91.6  & 84.7 & \textbf{93.7}  & \textbf{ 95.8}  & 87.8  & \textbf{96.4}  & 91.7  & 64.2 & 50.1 & 50.8 & 51.2 & 47.4 & 67.2 & 55.1 &  66.6\\
    & OVANet*\hyperlink{cite:b}{\tnote{b}}   & \xmark  & - & 86.8 & 76.1  & 90.1  & 94.9  & 91.0 & 95.2 & 89.0  & 68.2 & 58.6 & 60.5 & 59.1 & 57.0 & 68.6 & 62.0 &  56.6\\
    & UniOT*\hyperlink{cite:c}{\tnote{c}}   & \xmark  & - & 78.5 & 73.4 & 82.5 & 88.4  & 83.0  & 88.1  &  82.3 & 71.4 & 62.8 & 62.5 & 64.6 & 59.5 & 72.7 & 65.6 &  56.3 \\
    & WiSE-FT*\hyperlink{cite:d}{\tnote{d}}& \cmark  & \cmark & 60.4  & 50.7  & 72.6  & 77.9  & 70.6  & 90.2  & 70.4  & 5.07 & 1.15 & 2.84 & 1.64 & 0.84 & 3.06 & 2.43 & 46.5 \\
    & DCC\hyperlink{cite:j}{\tnote{j}}   & \xmark & - & 82.2 & 76.9 & 83.6 & 75.2 & 85.8 & 88.7 & 82.1 & 61.1 & 38.8 & 51.8 & 49.3 & 49.1 & 60.3 & 52.2 & 61.2  \\
    & GLC\hyperlink{cite:f}{\tnote{f}}   & \cmark  & \cmark & 80.5 & 80.4 & 77.5 & 95.6 & 77.7 & 96.9 & 84.8 & 51.2 & 44.5 & 55.6 & 43.1 & 47.0 & 39.1 & 46.8 & 80.3\\
    & MemSPM\hyperlink{cite:g}{\tnote{g}}   & \xmark & - & 88.0 & 84.6 & 88.7 & 87.6 & 87.9 & 94.3 & 88.5 & 62.4 & 52.8 & 58.5 & 53.3 & 50.4 & 62.6 & 56.7 & 79.3 \\
    & CLIP-Distill*\hyperlink{cite:e}{\tnote{e}}   & \cmark  & \cmark & 81.6  & 81.1  & 87.7  & 81.0 & 87.6  & 81.8  & 83.5  & 74.2 & 61.6 & 58.5 & 60.2 & 59.7 & 74.8 &  64.8  & 81.9 \\
    \rowcolor{gray!10} & TLSA   & \cmark  &\textcolor{blue}{\xmark} & 92.7  & 86.8  & 90.1  & 86.8  & 90.1  & 92.6  & 89.9  & \textbf{83.6} & \textbf{71.2} & \textbf{71.3} & \textbf{71.2} & \textbf{71.3} & \textbf{83.6 }& \textbf{75.4} & 86.3 \\
    \rowcolor{gray!10} & TLSA+ST   & \cmark  &\textcolor{blue}{\xmark} & \textbf{95.4} & \textbf{88.7} & 91.5 & 88.7 & \textbf{91.5} & 95.4 & \textbf{91.9} & 82.3 & 69.4 & 71.2 & 69.4 & 71.2 & 82.3 & 74.3 & \textbf{87.6}\\
    \bottomrule
    \end{tabular}%
    \end{threeparttable}
  }
  \label{tab:opda_rest}%
  \vspace{-0.1in}
\end{table*}

\subsection{Quantitative Comparison} \label{subsec:analysis} 

\noindent{\textbf{Comparison with Baseline Methods.}} 
We compared TLSA against existing UniDA approaches. Among these, CLIP-Distillation achieves state-of-the-art (SOTA) performance. Except for the Office31 open-set and VisDA partial settings, we attain SOTA across all scenarios. Although the impact of self-training is more significant on certain datasets, it generally enhances the H$^3$-score in most cases. The results are summarized in Tab.~\ref{table:results-all-clip}. Since calculating the H-score and H$^3$-score for partial and closed-set scenarios is infeasible, we employed the average accuracy score as a substitute for the H-score and H$^3$-score, following \citet{deng2023universal}. These findings highlight the versatility and effectiveness of TLSA and TLSA+ST (Self-Training) across diverse UniDA settings, particularly in open-partial and open scenarios where managing unseen classes is critical.

In the open-partial setting, TLSA+ST achieved the highest scores across datasets, with H-scores of 93.5 for Office-Home and 87.3 for VisDA. The H$^3$-scores were similarly high, reaching 89.2 on Office-Home and 87.6 on VisDA, demonstrating the ability of TLSA+ST to adapt to partially overlapping classes. In the open setting, TLSA+ST leads with H-scores of 95.2 on Office31 and 87.6 on VisDA, while H$^3$-scores are also competitive at 89.9 on Office31 and 84.9 on VisDA, showcasing the robustness of the method in managing unseen target classes. In the closed setting, where all classes overlap, TLSA+ST maintained competitive performance with an H-score of 75.6 on VisDA, significantly surpassing other methods. In the partial setting, TLSA+ST also delivered strong results, achieving an H-score of 74.4 on DomainNet. Collectively, these results highlight the effectiveness of TLSA and TLSA+ST across various UniDA scenarios, particularly excelling in open-partial and open settings where handling unseen classes is crucial.

 \begin{figure}[t] \centering \includegraphics[width=\columnwidth]{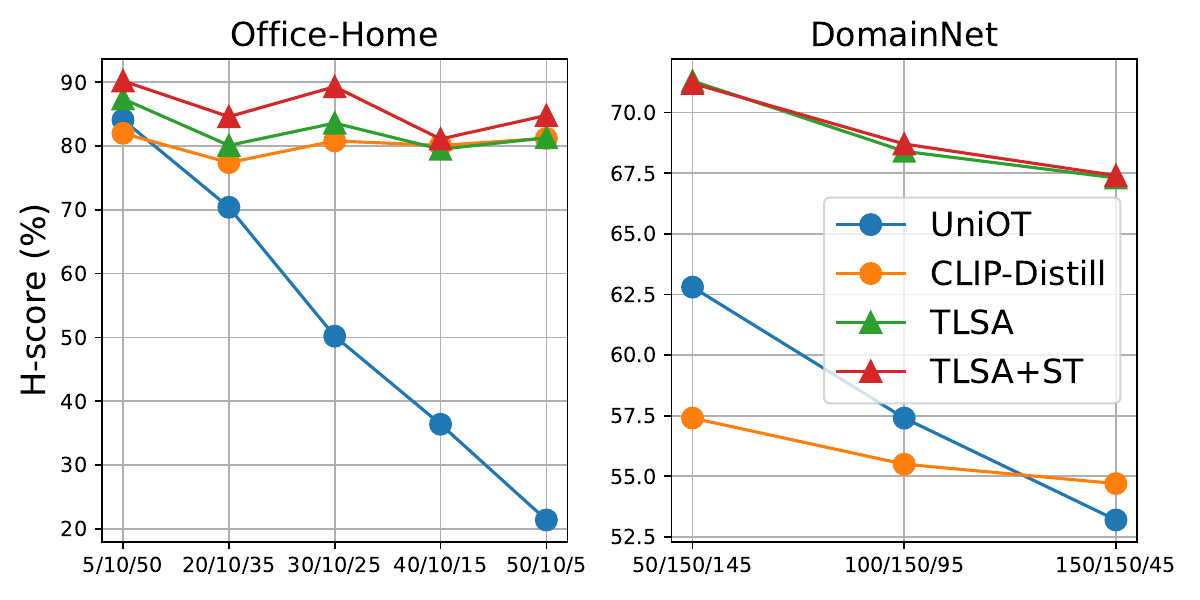} \caption{Robustness across different class split settings in Office-Home (left) RealWorld to Clipart, and DomainNet (right) Real to Painting.}\label{fig:robust} \end{figure}

 \begin{figure}[t]
    \centering
    \includegraphics[width=\columnwidth]{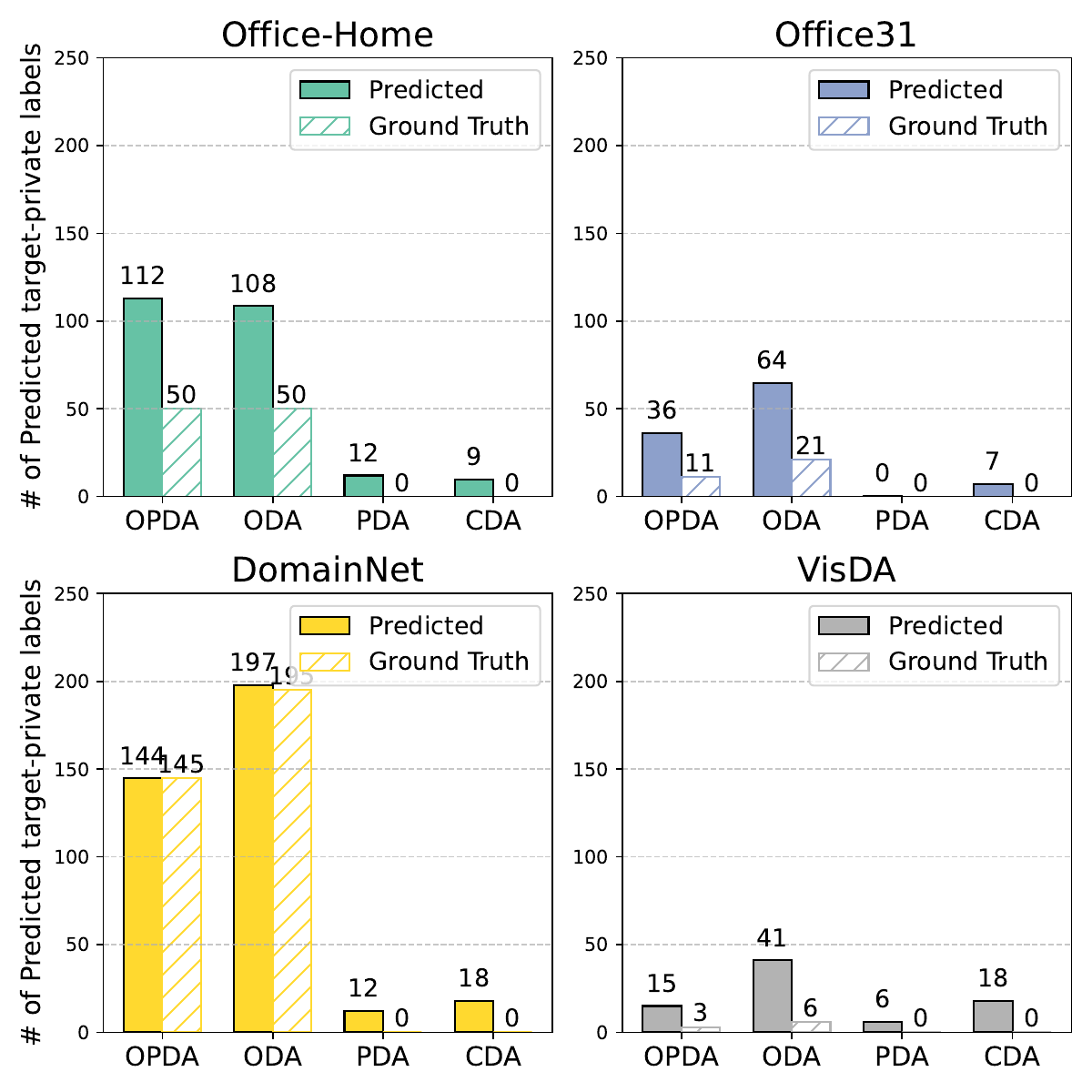}
    \caption{The number of predicted target-private labels across different split settings and datasets.}
    \label{fig:number}
\end{figure}

Tab.~\ref{tab:officehome} and~\ref{tab:opda_rest} compare our results with existing methods in the open partial domain adaptation scenario using both ResNet50 (ImageNet-pretrained) and ViT-B/16 (CLIP-pretrained) backbones. Our approach achieved state-of-the-art performance across nearly all domain-to-domain adaptations. However, a slight reduction in H-score was observed due to self-training in DomainNet. While closed-set accuracy improved, OOD detection performance tended to decrease, and vice versa. This indicates a trade-off: self-training shifts the focus of the model, often leading to a decline in H-score performance.

\noindent{\textbf{Robustness in Various UniDA Split Setting.}} 
We compared the performance of our TLSA method with existing approaches, including CLIP-distillation and UniOT, across different category splits. With a fixed number of shared labels, we designed five settings for $\bar{C_s}$/$C$/$\bar{C_t}$: 5/10/50, 20/10/35, 30/10/25, 40/10/15, and 50/10/5 for Office-Home, and three settings: 50/150/145, 100/150/95, and 150/150/45 for DomainNet. Fig.~\ref{fig:robust} demonstrates the robustness of our TLSA method to variations in these split settings.

 \begin{figure*}[t]
    \centering
    \begin{subfigure}[t]{0.32\textwidth}
        \centering
        \textbf{H-score: 41.7\%}\\[2mm]
        \includegraphics[width=\linewidth]{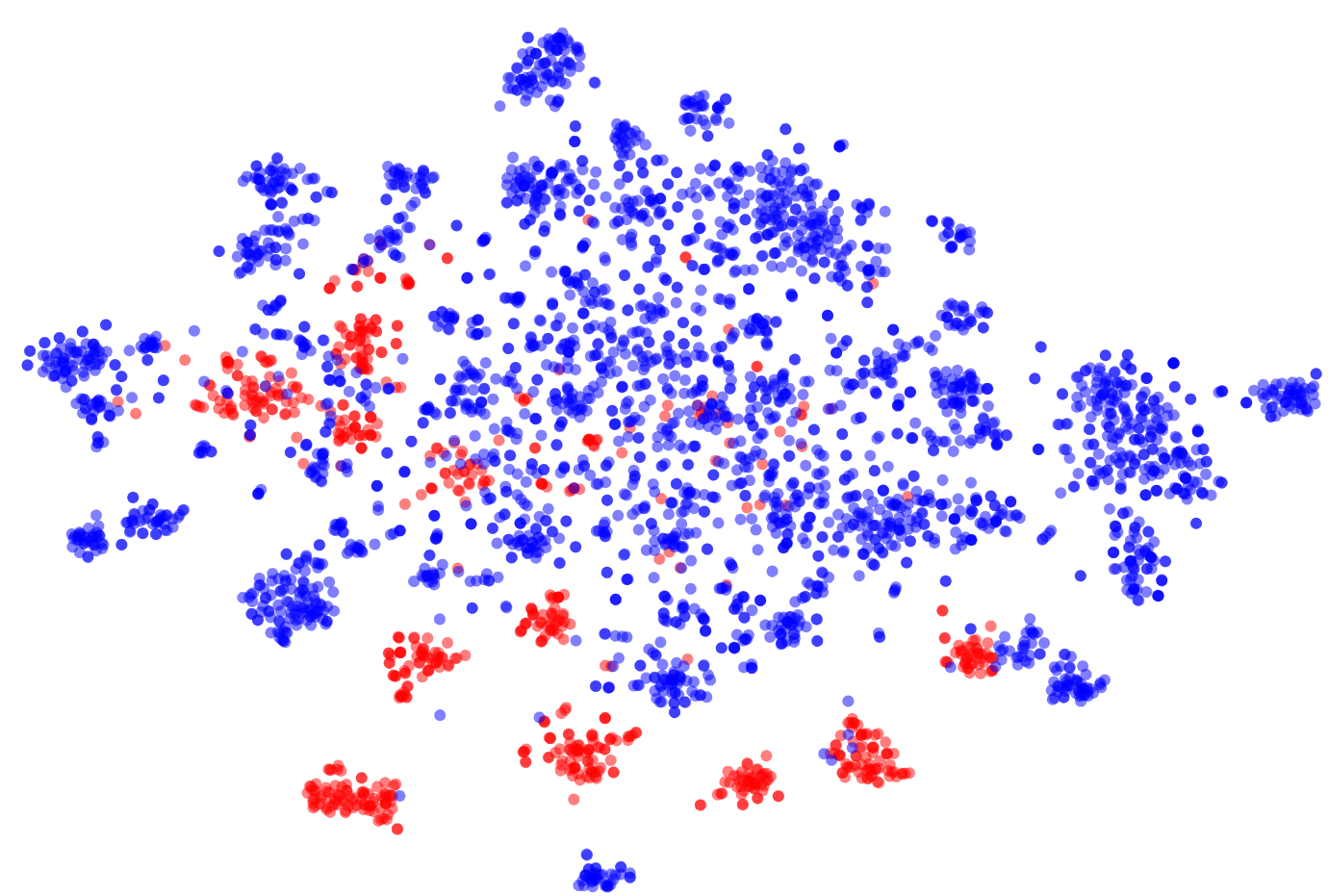}
        \caption{CLIP zero-shot}
    \end{subfigure}\hfill
    \begin{subfigure}[t]{0.32\textwidth}
        \centering
        \textbf{H-score: 66.02\%}\\[2mm]
        \includegraphics[width=\linewidth]{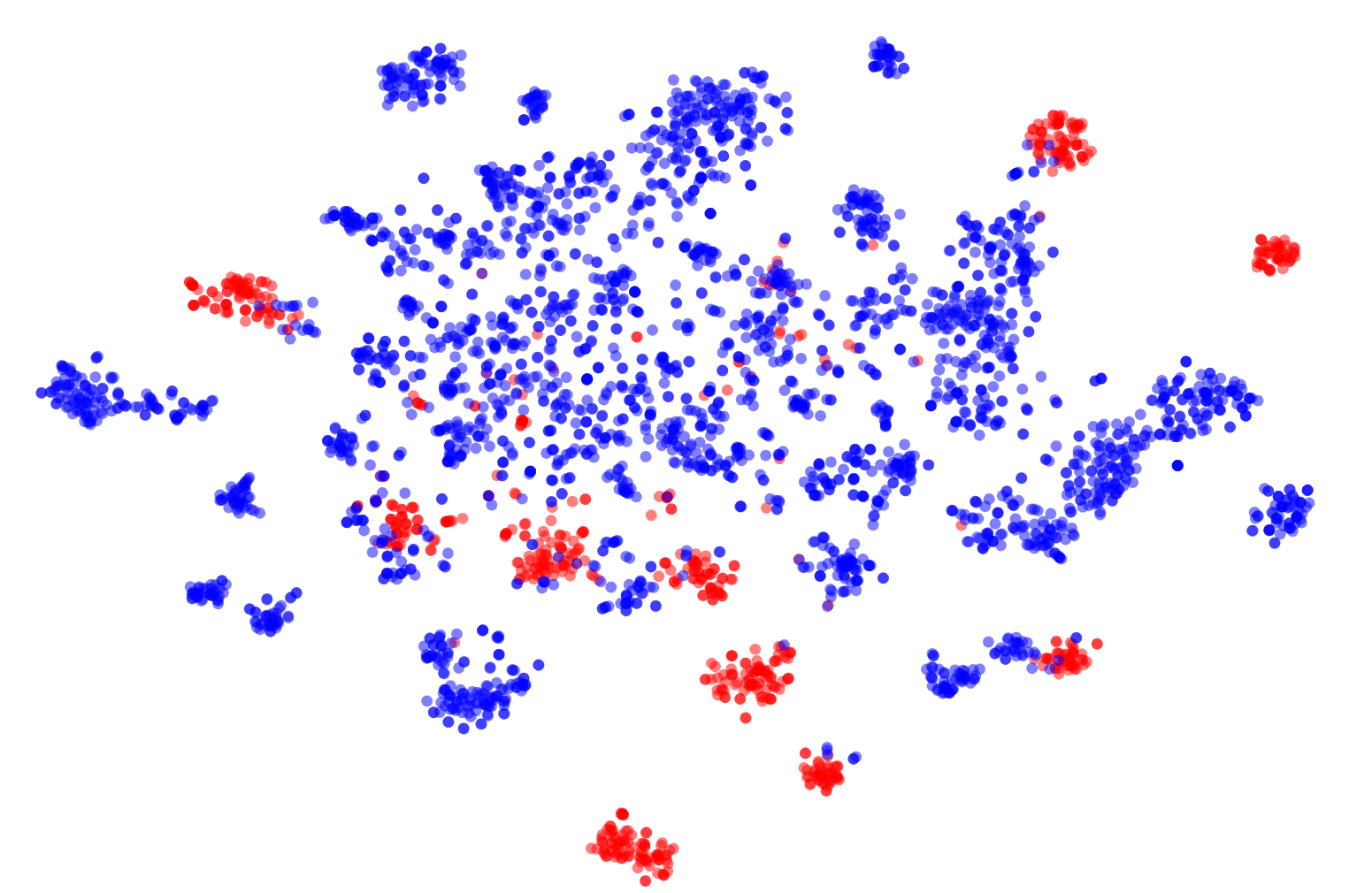}
        \caption{TLSA+ST ($|C_s|{+}1$ classifier)}
    \end{subfigure}\hfill
    \begin{subfigure}[t]{0.32\textwidth}
        \centering
        \textbf{H-score: 90.2\%}\\[2mm]
        \includegraphics[width=\linewidth]{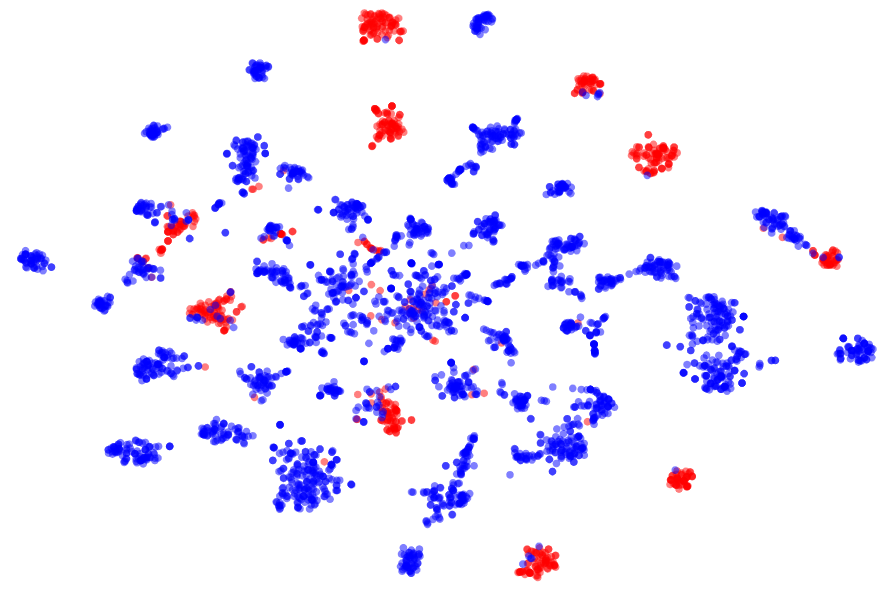}
        \caption{TLSA+ST (Universal classifier)}
    \end{subfigure}

    \caption{The t-SNE plots of target domain samples for the zero-shot and \ours+ST methods in the Office-Home RealWorld $\to$ Clipart OPDA setting.
    \textcolor{blue}{Blue points} indicate target-private samples, while \textcolor{red}{red points} represent source-class samples.}
    \label{fig:our_tsne}
\end{figure*}

\begin{table}[h!]
    \centering
    \caption{List of ground truth target-private labels and final candidates extracted from TLSA for the Office31, DSLR, open-partial setting. The same color indicates the same semantic group.}
    \begin{tabular}{c c c}
        \toprule
        \rowcolor{white!10} \textbf{True target-private} & \multicolumn{2}{c}{\textbf{Predicted target-private}}  \\ 
        \midrule
         \multirow{2}{*}{\textcolor{red}{\raisebox{-1ex}{\includegraphics[height=3ex]{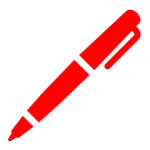}} pen}} & \textcolor{red}{pen} & \textcolor{red}{bat} \\
        & \textcolor{red}{pencil} & - \\
        \rowcolor{gray!10} \textcolor{blue}{\raisebox{-1ex}{\includegraphics[height=3ex]{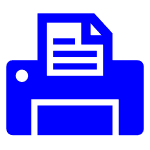}} printer} & \textcolor{blue}{printer} &  \textcolor{blue}{fax machine} \\
        \rowcolor{white!10} \textcolor{green}{\raisebox{-1ex}{\includegraphics[height=3ex]{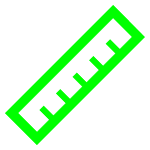}} ruler} & \textcolor{green}{ruler} & - \\ 
         \rowcolor{gray!10} \textcolor{purple}{\raisebox{-1ex}{\includegraphics[height=3ex]{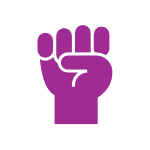}} punchers} & \textcolor{purple}{stapler} & - \\ 
        \rowcolor{white!10} \textcolor{pink}{\raisebox{-1ex}{\includegraphics[height=3ex]{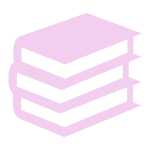}} ring binder} & \textcolor{pink}{fax machine} & - \\ 
         \rowcolor{gray!10} \textcolor{orange}{\raisebox{-1ex}{\includegraphics[height=3ex]{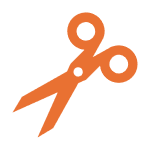}} scissors} & \textcolor{orange}{scissors} & - \\ 
        \rowcolor{white!10} \textcolor{black}{\raisebox{-1ex}{\includegraphics[height=3ex]{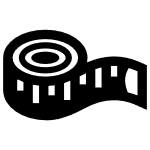}} tape dispenser} & \textcolor{black}{paper dispenser} & - \\ 
         \rowcolor{gray!10} \textcolor{cyan}{\raisebox{-1ex}{\includegraphics[height=3ex]{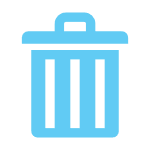}} trash can} & \textcolor{cyan}{trash can} & \textcolor{cyan}{recycle bin} \\ 
       \rowcolor{white!10} \textcolor{lime}{\raisebox{-1ex}{\includegraphics[height=3ex]{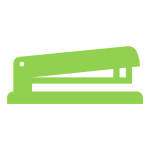}} stapler} & \textcolor{lime}{stapler} & - \\ 
         \rowcolor{gray!10} \textcolor{brown}{\raisebox{-1ex}{\includegraphics[height=3ex]{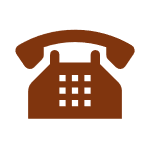}} phone} &  \textcolor{brown}{modem} & - \\ 
        \rowcolor{white!10} \textcolor{violet}{\raisebox{-1ex}{\includegraphics[height=3ex]{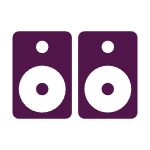}} speaker} & \textcolor{violet}{speaker} & \textcolor{violet}{usb drive} \\
         \rowcolor{gray!10} \multirow{2}{*}{-} & tv & scale \\
        \rowcolor{gray!10} & remote &  \\
        \bottomrule
    \end{tabular}
    \label{tab:exampleoftargetprivateset}
\end{table}

\subsection{Qualitative Comparison}

\noindent{\textbf{Analysis of the Number of Predicted Target-Private Labels.}}
Fig.~\ref{fig:number} displays the number of predicted target-private labels in our study. Across various datasets and settings, our method accurately predicts target-private labels compared to their ground-truth counterparts. This illustrates the flexibility of our design in adapting to different split configurations and datasets.

\noindent{\textbf{Feature Visualization.}}
We present t-SNE~\citep{long2016unsupervised} plots in Fig.~\ref{fig:our_tsne} to compare feature discrimination between the CLIP zero-shot and $|C_s|+1$classifiers using our approach. As other methods employ a fixed backbone identical to that in~\citep{deng2023universal}, we omitted their representations from the visualization. Furthermore, to emphasize our universal classifier's discriminative ability, we utilized a $|C_s|+1$classifier with LP-ST (linear probing via self-distillation and self-training using top-$k$ confident sample selection), similar to LP-FT~\citep{kumar2022fine}. The $|C_s|+1$-way classifier underperforms compared to TLSA due to its constrained capacity to represent unknown classes.

\noindent{\textbf{Predicted Target-Private Labels Analysis.}} 
Tab.~\ref{tab:exampleoftargetprivateset} presents examples of the target-private set extracted by the proposed method.
We observed that these predicted labels align with the ground-truth target-private categories in the target domain under the DSLR setting of Office31, indicating that TLSA effectively identifies genuine target-private classes.
Importantly, the discovered labels do not need to precisely match the ground-truth annotations; they often also capture prominent attributes of target-domain images.
This flexibility is beneficial because generative VLMs can function effectively even on objects not encountered during pretraining, making them particularly suited for open vocabulary or unknown classes.
This process generates a set of discovered labels, $R = \{r_i\}_{i=1}^N$, where $N$ represents the batch size of the target image.

\subsection{Ablation Study}
\noindent{\textbf{Ablation Study and Comparison of the Alignment Pipeline.}} 
We implement an ablation study of label space alignment.  As can be seen in Tab.~\ref{tab:filtering}, Synonym label alignment serves as a fundamental filtering method, contributing to performance to a certain extent.

Top-1 prediction-based alignment relies solely on the top-1 prediction. If the top-1 prediction corresponds to a discovered label, the sample is classified as a target-private sample; otherwise, if it corresponds to a source label, it is identified as a shared sample. In contrast, margin-based alignment requires that, even when a discovered label is the top-1 prediction, it is considered a target-private candidate only if its score exceeds that of the top-1 source label by a specified margin. This margin is empirically fixed at $\tfrac{1}{|C_s|}$ across all instances, which yields the best performance for this method.

Frequency-based noisy candidate filtering is a vital component of our pipeline; without it, highly noisy candidate labels could be incorrectly designated as target-private labels. A comparison of these methods is shown in Tab.~\ref{tab:filtering}.

\begin{table*}[!t]
   \caption{Ablation study with different components of our alignment pipeline approach. The semantic alignment strategy refers to the comparison of alignment methods used in step 2. Top-1 refers to top-1 prediction-based alignment, Margin refers to margin-based alignment, and \ours~refers to our method.}
    \vspace{-2mm}
    \resizebox{\textwidth}{!}{
    \setlength{\tabcolsep}{4pt}
    \begin{tabular}{cc c cc cccc cccc cccc cccc}
        \toprule
            \multicolumn{1}{c}{\centering \bf Step 1} &
           \multicolumn{3}{c}{\centering \bf Step 2}  & \multicolumn{1}{c}{\centering \bf Step 3} & \multicolumn{4}{c}{\centering \bf Office31} & \multicolumn{4}{c}{\centering \bf Office-Home}  & \multicolumn{4}{c}{\centering \bf VisDA}  & \multicolumn{4}{c}{\centering \bf DomainNet}  \\
        \cmidrule(l{4pt}r{4pt}){1-1} \cmidrule(l{4pt}r{4pt}){2-4} \cmidrule(l{4pt}r{4pt}){5-5} \cmidrule(l{4pt}r{4pt}){6-9} \cmidrule(l{4pt}r{4pt}){10-13} \cmidrule(l{4pt}r{4pt}){14-17} \cmidrule(l{4pt}r{4pt}){18-21}
          {\centering -- } & {\centering Top-1} & {\centering Margin} &  {\centering Ours} & {\centering --} & {\centering OPDA} & {\centering ODA} & {\centering PDA}  & {\centering CDA} & {\centering OPDA} & {\centering ODA} & {\centering PDA}  & {\centering CDA} & {\centering OPDA} & {\centering ODA} & {\centering PDA}  & {\centering CDA} & {\centering OPDA} & {\centering ODA} & {\centering PDA}  & {\centering CDA}  \\ 
        \midrule
        \cmark & \cmark &  & & \cmark & 88.8 & 91.5 & 89.9 & 71.3 & 86.6 & 83.8 & 72.7 & 66.3 & 76.6 & 75.2 & 64.7 & 54.1 & 74.8 & 75.8 & 66.2 & 65.3  \\  
        \cmark &   & \cmark & & \cmark & 89.2 & 92.1 & 91.6 & 71.7 & 87.3 & 84.5 & 72.9 & 66.4 & 77.8 & 77.4 & 67.2 & 57.2 & 75.1 & 76.1 & 66.6 & 65.5 \\  
         &   &  & \cmark & \cmark & 88.6 & 92.0 & 92.3 & 76.8 & 87.2 & 86.4 & 76.9 & 72.2 & 86.3 & 82.2 & 75.6 & 70.8 &\textbf{ 76.1} & \textbf{77.1} & 69.1 & 67.6 \\ 
        \cmark &   &  & \cmark & & \textbf{90.0} & 91.9 & 94.6 & 78.6 & 89.8 & 85.2 & 79.7 & 74.2 & 47.3 & 45.7 & 28.9 & 28.8 & 71.6 & 71.1 & 62.8 & 60.4
 \\  
        \rowcolor{gray!10} \cmark &  &  & \cmark & \cmark &	89.9	&	\textbf{92.6}	&	\textbf{96.1}	&	\textbf{80.5}	
    &	\textbf{91.3}	&	\textbf{89.2}	&	\textbf{82.0}	&	\textbf{79.0}
    &	\textbf{86.3}	&	\textbf{83.9}	&	\textbf{81.8}	&	\textbf{72.6}	&	
75.4	&	76.9	&	\textbf{70.1}	&	\textbf{68.9 }\\      
        \bottomrule
    \end{tabular}}
    \vspace{-3mm}
    \label{tab:filtering}
\end{table*}

\begin{table}[ht]
\centering
\footnotesize
\caption{Ablation study comparing H-scores across different prediction set decision thresholds}
\label{tab:IAT}
\begin{tabular}{c  c c c c c}
\toprule
                         methods               & OPDA & ODA & PDA & CDA  & Avg\\ \midrule
                         w/o $\tau_{gap}$      &  \textbf{85.8} & \textbf{85.6} & 81.0 & 75.0 & 81.9  \\ 
                         w/o $\tau_{mean}$       & 84.1 & 83.7 & 77.5 &  	71.2 & 79.1  \\ \midrule
                         \rowcolor{gray!10} Ours      & 85.7 & \textbf{85.6} & \textbf{82.5}& \textbf{75.2} & \textbf{82.3} \\  \bottomrule
\end{tabular}
\vspace{1px}
\end{table}

\begin{table*}[!t]
   \caption{ Ablation study with different components of our self-training approach. Delay refers to how long the teacher model's updates are delayed. }
    \vspace{-2mm}
    \resizebox{\textwidth}{!}{
    \setlength{\tabcolsep}{4pt}
    \begin{tabular}{cc c cc cccc cccc cccc cccc}
        \toprule
            \multicolumn{1}{c}{\centering \bf Delay} &
           \multicolumn{2}{c}{\centering \bf Selection Strategy}   & \multicolumn{4}{c}{\centering \bf Office31} & \multicolumn{4}{c}{\centering \bf Office-Home}  & \multicolumn{4}{c}{\centering \bf VisDA}  & \multicolumn{4}{c}{\centering \bf DomainNet}  \\
        \cmidrule(l{4pt}r{4pt}){1-1} \cmidrule(l{4pt}r{4pt}){2-3} \cmidrule(l{4pt}r{4pt}){4-7} \cmidrule(l{4pt}r{4pt}){8-11} \cmidrule(l{4pt}r{4pt}){12-15} \cmidrule(l{4pt}r{4pt}){16-19}
          {\centering --} & {\centering Top-k\%} & {\centering Balanced Top-k}  & {\centering OPDA} & {\centering ODA} & {\centering PDA}  & {\centering CDA} & {\centering OPDA} & {\centering ODA} & {\centering PDA}  & {\centering CDA} & {\centering OPDA} & {\centering ODA} & {\centering PDA}  & {\centering CDA} & {\centering OPDA} & {\centering ODA} & {\centering PDA}  & {\centering CDA}  \\ 
        \midrule
        \multicolumn{3}{c}{\centering  Universal Classifier zero-shot } &89.9& 92.6 & 96.1 &	80.5 & 91.3 &89.2 &	82.0 & 79.0 & 86.3 & 83.9 &	81.8 & 72.6 & \textbf{75.4} & \textbf{76.9} & 70.1 & \textbf{68.9} \\ 
        \midrule
        &   & & 91.5 & 94.7 & 97.2 & 82.3 & 93.3 & 90.9 & 84.6 & 81.5 & 87.1 & 85.8 & 85.0 & 75.7 & 73.9 & 74.4 & 72.1 & 67.0 \\ 
        \cmark &  &  & 91.6 & 95.1 & \textbf{97.7} & 82.5 & \textbf{93.7} & 91.3 & 85.2 & 81.2 & 87.2 & 86.0 &\textbf{ 85.4} & \textbf{76.1} & 73.9 & 74.4 & 72.2 & 66.6 \\  
        \cmark & \cmark  &  & 91.5 & 94.7 & 97.2 & 82.3 & 93.3 & 90.9 & 84.6 & 81.5 & 87.1 & 85.8 & 85.0 & 75.7 & 73.9 & 74.4 & 72.1 & 67.0 \\   
        \rowcolor{gray!10} \cmark &   & \cmark   & \textbf{91.9} & \textbf{95.2} & 96.9 & \textbf{82.9} & 93.5 & \textbf{91.4} & \textbf{85.3} & \textbf{81.8 }& \textbf{87.6} & \textbf{86.0} & 83.0 & 75.6 & 74.3 & 74.9 & \textbf{74.4} & 67.9 \\      
        \bottomrule
    \end{tabular}}
    \vspace{-3mm}
    \label{tab:self-training}
\end{table*}

\noindent{\textbf{Ablation Study of Two Criteria for Label Space Alignment.}}
We conducted an ablation study on the two thresholds, $\tau_{\text{avg}}$and $\tau_{\text{gap}}$, used in adaptive filtering. As outlined in Tab.~\ref{tab:IAT}, $\tau_{\text{avg}}$significantly impacts performance more than $\tau_{\text{gap}}$. Furthermore, $\tau_{\text{avg}}$and $\tau_{\text{gap}}$operate as complementary thresholds. Our design ensures robustness against split-setting variations and diverse datasets.

\noindent{\textbf{Ablation Study of Self-Training Components.}}
We conducted experiments to enhance the reliability of our self-training design. The key observation was that incorporating delayed EMA components (Delay in Tab.~\ref{tab:self-training}) slowed the training process of the teacher model and significantly improved the model performance. Moreover, our approach surpassed the top-k\% pseudo-label sampling strategy proposed in a previous study~\citep{zara2023autolabel} (Tab. ~\ref{tab:self-training}).

\subsection{Further Analysis}

\begin{table}[h]
\centering
\footnotesize
\caption{Comparison of H-score on DomainNet to analyze the sensitivity of hyperparameter k in semantic label alignment and $\epsilon$ in frequency-based noisy candidate filtering.}
\label{tab:k}
\begin{tabular}{c  c c c c c}
\toprule
                         k              & OPDA & ODA & PDA & CDA & Avg \\ \midrule
                         4       &  \textbf{75.9} & \textbf{77.3} & 69.5 & 68.4 & 72.7 \\ 
                         \rowcolor{gray!10} 5    &  75.4 & 76.9 & 70.1 & 68.9 & \textbf{72.8}\\ 
                         6       & 74.6 & 76.3 & 70.4 & 69.3 & 72.6 \\
                         7       & 72.1 & 74.8 & \textbf{70.8} &\textbf{ 69.6} & 71.8\\ \midrule
                         $\epsilon$              & OPDA & ODA & PDA & CDA  & Avg \\ \midrule
                         \rowcolor{gray!10} 0.01       &  \textbf{75.4}&	\textbf{76.9}	& 70.1	& 68.9 & \textbf{72.8}\\ 
                         0.02      & 73.4 &	75.7&	70.6&	69.6 & 72.3\\ 
                         0.03       & 72.1 &	74.4	&\textbf{71.0} & 69.7 & 71.8 \\
                         0.04       & 71.1 & 73.2 &	\textbf{71.0} &	\textbf{69.8} & 71.3 \\ \bottomrule
\end{tabular}
\vspace{1px}
\end{table}

This section includes sensitivity analyses of hyperparameters during the alignment process, scalability evaluations relative to backbone size, learning dynamics during self-training, and analyses of trade-offs related to source sample usage.

\noindent{\textbf{Sensitivity Analysis for Hyperparameters in TLSA.}}
We analyze the sensitivity of our hyperparameters, $k$, in the top-$k$ prediction within the label space alignment pipeline and $\epsilon$ in frequency-based noisy candidate filtering. The results are presented in Tab.~\ref{tab:k}. The experiments on the DomainNet dataset determined the optimal value based on average performance.

Specifically, we performed a sensitivity analysis on the $k$ value used for top-$k$ predictions in Step 2 (semantic label alignment) of the filtering pipeline. Our pipeline consistently outperformed other UniDA methods, leading us to select $k=5$ for its robustness across various configurations.

We perform a sensitivity analysis on the value of $\epsilon$ used for frequency-based noisy candidate filtering. A larger $\epsilon$ results in a higher threshold, accepting fewer candidate labels, while a smaller $\epsilon$ leads to a lower threshold, allowing a broader range of target private label candidates. We observe that the performance remains robust across different values of the size of $\epsilon$. Among these, we select 0.01, which shows superior performance across 4 datasets. 

\begin{table}[h]
    \centering
    \setlength{\tabcolsep}{6pt}
    \caption{Comparison of H-score with larger backbones: OH denotes Office-Home, VD denotes VisDA, and DN denotes DomainNet.}
    \begin{tabular}{lcccc}
    \toprule[1.0pt]
    \multirow{2}{*}{Backbone}  & \multicolumn{4}{c}{Split Setting} \\
    \cmidrule{2-5}
     & Office31 & OH & VD & DN \\
    \midrule
    \multicolumn{5}{l}{\textbf{(a) \ours}} \\
    ViT-B/16 & 85.7 & 85.6 & 82.5 & 75.2 \\
    ViT-L/14@336px & \textbf{92.9} & \textbf{90.3} & \textbf{84.6} & \textbf{78.9} \\
    \midrule
    \multicolumn{5}{l}{\textbf{(b) \ours+ST}} \\
    ViT-B/16 & 86.8 & 86.9 & \textbf{84.9} & 77.0 \\
    ViT-L/14@336px &\textbf{94.0 }& \textbf{92.0} & 83.5 & \textbf{79.9}\\
    \bottomrule[1.0pt]
    \end{tabular}
    \label{tab:scale}
\end{table}

\noindent{\textbf{Scalability Analysis for Model Sizes.}}
We conducted experiments using the CLIP-ViT-B/16 backbone. A crucial aspect of validating our method is determining whether performance scales with the backbone size. As illustrated in Tab.~\ref{tab:scale}, increasing the backbone to CLIP-ViT-L/14@336px leads to a significant performance enhancement.

\begin{figure}[t]
    \centering
    \includegraphics[width=\columnwidth]{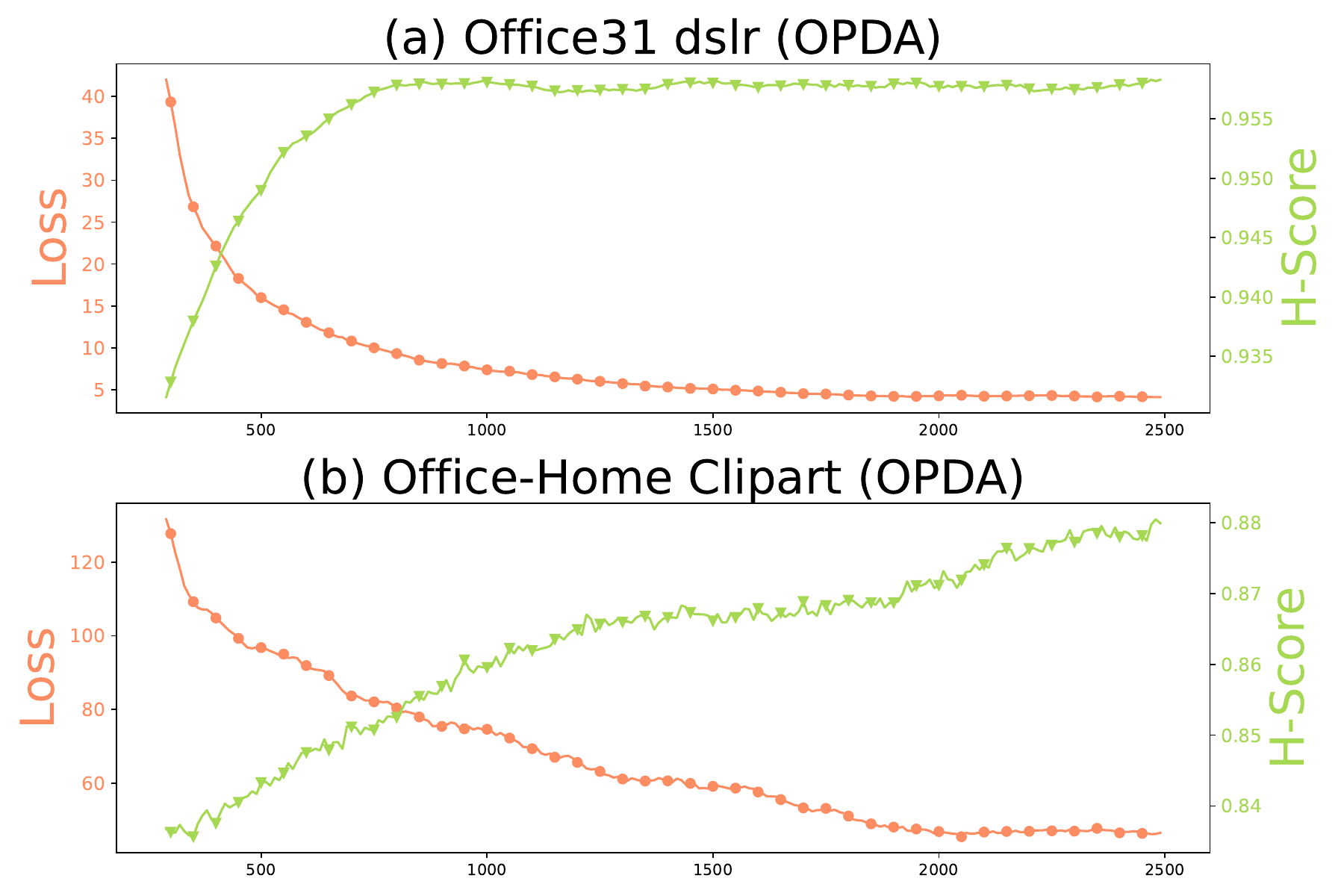}
    \caption{Learning Statistics of Our Self-Training. The x-axis denotes the iteration.}
    \label{fig:learning_statistics}
\end{figure}

\noindent{\textbf{Learning Statistics of Self-Training.}}
We present self-training learning statistics for Office31 DSLR and Office-Home Clipart under the OPDA setting in Fig.~\ref{fig:learning_statistics}. A notable performance improvement occurs after a certain number of iterations.


\begin{table}[h]
\centering
\footnotesize
\caption{Comparison of H-scores between self-training combined with source training and ours (self-training only)}
\label{tab:source}
\begin{tabular}{lcccc}
\toprule
\textbf{Method} & \textbf{OPDA} & \textbf{ODA} & \textbf{PDA} & \textbf{CDA} \\ 
\midrule
\textbf{ w/ $L_{src}$} & 85.2 &	86.7	 &\textbf{85.3}	 & \textbf{80.0} \\
\rowcolor{gray!10} \textbf{Ours (w/o $L_{src}$)} &\textbf{86.8}&	\textbf{86.9}	&84.9&	77.0 \\  
\bottomrule
\end{tabular}
\vspace{1px}
\end{table}

\noindent{\textbf{Trade-off for Training Using the Source Samples.}} We observe that incorporating the source loss term $L_{src}$ introduces a tradeoff. Tab.~\ref{tab:source} demonstrates that integrating source training significantly boosts performance in partial domain adaptation (PDA) and closed-set domain adaptation (CDA) compared to self-training. However, this approach leads to higher computational demands and slightly reduced performance in open-set domain adaptation (OPDA) and open domain adaptation (ODA). Therefore, the method can be selectively employed based on the objective (Tab.~\ref{tab:source}).

\section{Limitations}
A key limitation of our approach is that it does not explicitly address source-private class detection, which may result in suboptimal performance within the universal label space. However, we note that explicit source-private class detection often leads to mis-detection in practice, and focusing on target-private class detection can achieve most of the performance gains.

\section{Conclusion}
In this paper, we introduce Training-free Label Space Alignment (TLSA), a framework that leverages CLIP to address the Universal Domain Adaptation (UniDA) problem through label space alignment. TLSA utilizes a generative vision-language model to discover labels for all target domain samples and filters out those classified as source classes. On a mini-batch level, we eliminate lexical synonyms of source labels through synonym label alignment and remove semantically ambiguous labels via semantic label alignment. At the dataset level, frequency-based noisy candidate filtering removes unreliable labels based on occurrence patterns. This approach is highly efficient in both time and memory, completing the entire process in a single epoch without requiring further tuning. After alignment, we extend the source class set with the newly predicted target-private labels to construct a universal classifier, which significantly outperforms existing methods. Additionally, through self-training with balanced top-k confident sample selection, we surpass state-of-the-art UniDA baselines.


\section*{Data Availability}
All the data utilized in this paper is publicly available at~\citep{saenko2010adapting,venkateswara2017deep,peng2017visda,peng2019moment}.

\section*{Declaration of generative AI in scientific writing}
During the preparation of this work, the author used ChatGPT~\citep{achiam2023gpt} in order to improve readability. After using this tool/service, the author reviewed and edited the content as needed and takes full responsibility for the content of the publication.

\appendix

\section{Pseudo Code}
In this section, we provide a pseudo-code of our semantic label alignment approach in Alg.~\ref{alg:iat}.

\begin{algorithm}[t]
\caption{Semantic Label Alignment}
\label{alg:iat}
\begin{algorithmic}[1]
\Require Target samples $\mathbf{x}^t$, filtered discovered labels $\tilde{R}$, source labels $C_s$
\Ensure Frequency bank $F$

\State Compute target embeddings: $\mathbf{z}_v \gets f_v(\mathbf{x}^t)$
\State Compute label embeddings: $\mathbf{z}_t \gets f_t([C_s, \tilde{R}])$
\State Compute similarity matrix: $S \gets sim(\mathbf{z}_v, \mathbf{z}_t)$
\State Extract top-$k$ scores and indices: $(S_k, I_k) \gets \operatorname{topk}(S)$
\State Find largest gap index: $\mathbf{J} \gets \arg\max_j (s_{j+1} - s_j)$
\State Define thresholds:
\Statex \hspace{0.5cm} $\tau_{\text{gap}} \gets s_{\mathbf{J}}$
\Statex \hspace{0.5cm} $\tau_{\text{avg}} \gets \mathbb{E}[S_k]$
\Statex \hspace{0.5cm} $\tau_{\text{set}} \gets \min\{\tau_{\text{gap}}, \tau_{\text{avg}}\}$

\For{$i = 1$ to $N$}
    \State Prediction set:
    $\mathcal{C}^i \gets \{\mathbf{t}_{i,j} \mid j \in I_k^i,\; s_{i,j} > \tau_{\text{set}}^i\}$
    \If{$\mathcal{C}^i \cap C_s = \emptyset$}
        \State $F[r_i] \gets F[r_i] + 1$
    \Else
        \State $F[p_i] \gets F[p_i] + 1$
    \EndIf
\EndFor
\end{algorithmic}
\end{algorithm}

\begin{figure*}[h!]
    \centering
    \includegraphics[width=0.7\textwidth]{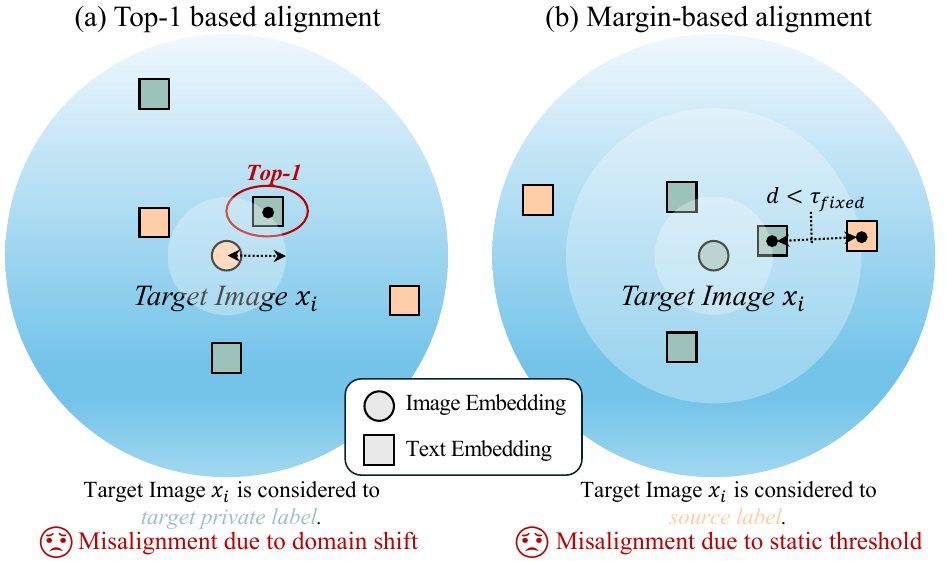}
    \caption{Illustrations of two naive filtering methods: (a) shows the misalignment in margin-based alignment. The example demonstrates a case where $\tau$ is not suitable for this particular dataset. (b) illustrates the misalignment in top-1 prediction-based alignment. This example shows how misalignment occurs when the source label is slightly less similar to the discovered label}
    \label{fig:naive}
\end{figure*}

\section{Examples of Label Space Alignment}
Fig.~\ref{fig:naive} illustrates the detailed process of two naive filtering methods, margin-based alignment and top-1 prediction-based alignment. In margin-based alignment, deriving the margin threshold $\tau$ is a non-trivial task because the class similarity between the source class and the target private class is not consistent across datasets (e.g., the case of ``clipboard" in source labels and ``notebook" in target private labels). In this scenario, ``notebook" is consistently hindered by "clipboard" due to the lack of a significantly high margin of similarity. Therefore, $\tau$ must be adaptively adjusted. Fig.~\ref{fig:naive}a demonstrates this misalignment case.

Additionally, in top-1 prediction-based alignment, in the case of ``backpack" in source labels and ``bag" in target private labels, since ``backpack" and ``bag" are not synonyms according to the WordNet hierarchy, ``bag" is often chosen as the target private candidate label on a per-instance basis. Fig.~\ref{fig:naive}b demonstrates this misalignment case.

\begin{figure}[h!] 
\centering
\caption{Label space alignment with examples. This figure demonstrates the alignment process by showcasing the discovered labels, corresponding candidate prediction sets, and the final filtered results for each item. The output is stored in the frequency bank, with counts incremented by one. Discovered labels are colored in \textcolor{violet}{violet}, and source labels are colored in \textcolor{orange}{orange}.}
\label{fig:examples}

\begin{minipage}{0.48\linewidth}
        \centering
        \subcaption{An Example of Source Class Samples}
        \resizebox{\linewidth}{!}{%
        \begin{tabular}{>{\centering\arraybackslash}m{1.5cm} 
                        >{\centering\arraybackslash}m{2.0cm}
                        >{\centering\arraybackslash}m{2.0cm}
                        >{\centering\arraybackslash}m{2.0cm}}
        \toprule
        \textbf{GT} & \textbf{backpack} & \textbf{projector} & \textbf{monitor} \\
        \midrule
        \textbf{\scriptsize{Image}} & 
        \includegraphics[width=1.0cm, height=1.0cm]{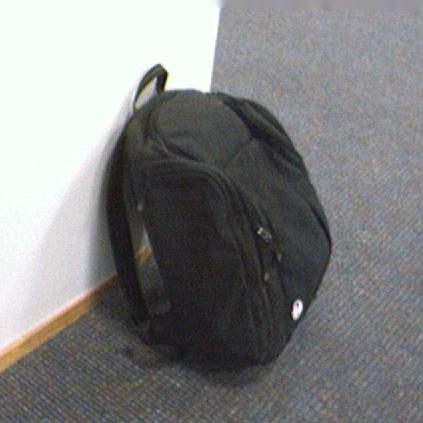} &  
        \includegraphics[width=1.0cm, height=1.0cm]{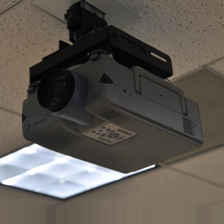} &  
        \includegraphics[width=1.0cm, height=1.0cm]{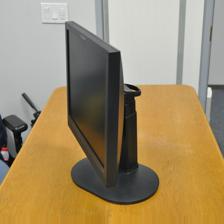} \\
        \midrule
        \textbf{\scriptsize{Discovered label}} & 
        \textcolor{violet}{bag} & 
        \textcolor{violet}{camera} & 
        \textcolor{violet}{tv} \\
        \midrule
        \textbf{\scriptsize{Prediction set}} & 
        1. \textcolor{orange}{backpack} & 1. \textcolor{orange}{projector} & 1. \textcolor{orange}{desktop}  \\
        \textbf{\scriptsize{$\mathcal{C}^i$}} & 2. \textcolor{violet}{bag} & 2. \textcolor{violet}{camera} & 2. \textcolor{orange}{monitor} \\ 
        \midrule
        \textbf{\scriptsize{Output}} & 
        \textcolor{orange}{backpack} & 
        \textcolor{orange}{projector} & 
        \textcolor{orange}{monitor} \\
        \bottomrule
        \end{tabular}}
    \end{minipage}
    \hfill
    \begin{minipage}{0.48\linewidth}
        \centering
        \subcaption{An Example of Target-Private Class Samples}
        \resizebox{\linewidth}{!}{%
        \begin{tabular}{>{\centering\arraybackslash}m{1.5cm} 
                        >{\centering\arraybackslash}m{2.0cm}
                        >{\centering\arraybackslash}m{2.0cm}
                        >{\centering\arraybackslash}m{2.0cm}}
        \toprule
        \textbf{GT} & \textbf{printer} & \textbf{stapler} & \textbf{trash can} \\
        \midrule
        \textbf{\scriptsize{Image}} & 
        \includegraphics[width=1.0cm, height=1.0cm]{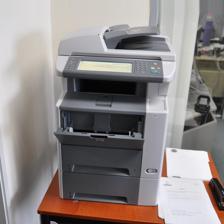} &  
        \includegraphics[width=1.0cm, height=1.0cm]{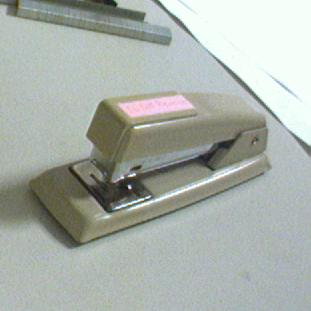} &  
        \includegraphics[width=1.0cm, height=1.0cm]{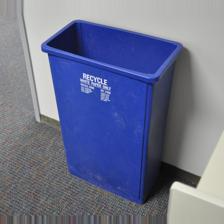} \\
        \midrule
        \textbf{\scriptsize{Discovered label}} & 
        \textcolor{violet}{printer} & 
        \textcolor{violet}{stapler} & 
        \textcolor{violet}{recycling} \\
        \midrule
        \textbf{\scriptsize{Prediction set}} & 
        1. \textcolor{violet}{fax machine} & 1. \textcolor{violet}{stapler} & 1. \textcolor{violet}{recycling} \\
        \textbf{\scriptsize{$\mathcal{C}^i$}} & 2. \textcolor{violet}{modem} &  & 2. \textcolor{violet}{trash can} \\
        & & & 3. \textcolor{violet}{recycle bin} \\
        \midrule
        \textbf{\scriptsize{Output}} & 
        \textcolor{violet}{printer} & 
        \textcolor{violet}{stapler} & 
        \textcolor{violet}{recycling} \\
        \bottomrule
        \end{tabular}}
    \end{minipage}
\end{figure}

\section{Details of Semantic Label Alignment}

Fig.~\ref{fig:examples} presents random examples from the Office31 dataset under the DSLR OPDA setting. It shows the exact discovered labels $r_i$ for each image $i$, the corresponding prediction set $\mathcal{C}^i$, and the specific elements stored in the frequency bank for each case. In the prediction set, the numbers indicate the ranking based on image-text similarity.

\section{Analysis of Labels to be Added to the Frequency Bank} \label{sec:blip}

In Sec.~\ref{sec:instance}, we noted that storing BLIP's discovered label to the frequency bank is more beneficial for performance compared to storing CLIP's top-1 prediction to the frequency bank. As shown in Tab.~\ref{tab:analysis2}, storing CLIP's top-1 predictions often leads to an overestimation of incorrect target private labels (e.g., overly specific or entirely incorrect classes), which in turn causes the correct target private labels (\eg, `train', `truck', `skateboard')  to be underestimated. This imbalance ultimately results in degraded model performance. In Tab.~\ref{tab:analysis2}, the correct target private label is underestimated when using CLIP's top-1 prediction compared to using BLIP's discovered label.

\begin{table}[h!]
    \centering
    \caption{Comparison of predicted target private labels between label selection strategies in label space alignment in VisDA open-partial setting. When BLIP's discovered label is incorporated into the target private candidate labels, the H-score (\%) is estimated at \textbf{86.3}. In contrast, using CLIP's top-1 prediction for target private candidate labels results in an H-score (\%) of \textbf{80.1}.}
    \resizebox{\columnwidth}{!}{
        \begin{tabular}{c c c c}
            \toprule
            \multicolumn{2}{c}{\textbf{Ours (BLIP's prediction)}} & \multicolumn{2}{c}{\textbf{CLIP's Top-1 Prediction}} \\
            \cmidrule(r){1-2} \cmidrule(l){3-4}
            Label & Frequency & Label & Frequency \\
            \midrule
            \multicolumn{4}{c}{\textbf{Correct Target-Private Label}} \\
            \midrule
            \cellcolor{gray!10}{train} & \cellcolor{gray!10}{3328} & train & 3099 \\
            \cellcolor{gray!10}{truck} & \cellcolor{gray!10}{2215} & truck & 1567 \\
            \cellcolor{gray!10}{skateboard} & \cellcolor{gray!10}{2035} & skateboard & 1857 \\
            \midrule
            \multicolumn{4}{c}{\textbf{Incorrect Target-Private Label}} \\
            \midrule
            fire truck & 221 & \cellcolor{gray!10}{fire truck} & \cellcolor{gray!10}{293} \\
            van & 533 & \cellcolor{gray!10}{van} & \cellcolor{gray!10}{574} \\
            \cellcolor{gray!10}{dog} & \cellcolor{gray!10}{152} & dog & 96 \\
            dump truck & 97 & \cellcolor{gray!10}{dump truck} & \cellcolor{gray!10}{205} \\
            skateboarder & 108 & \cellcolor{gray!10}{skateboarder} & \cellcolor{gray!10}{350} \\
            food truck & 136 & \cellcolor{gray!10}{food truck} & \cellcolor{gray!10}{214} \\
            \cellcolor{gray!10}{pizza} & \cellcolor{gray!10}{113} & pizza & 93 \\
            \cellcolor{gray!10}{umbrella} & \cellcolor{gray!10}{121} & umbrella & 87 \\
            taxi & 172 & \cellcolor{gray!10}{taxi} & \cellcolor{gray!10}{388} \\
            parking meter & 136 & \cellcolor{gray!10}{parking meter} & \cellcolor{gray!10}{252} \\
            \cellcolor{gray!10}{jeep} & \cellcolor{gray!10}{113} & jeep & 105 \\
            \hline
            fire hydrant & 81 & fork & 94 \\
            & & carriage & 96 \\
            & & locomotive & 98 \\
            & & suv & 110 \\
            & & police & 83 \\
            & & freightliner & 83 \\
            & & caboose & 107 \\
            & & trolley & 83 \\
            \bottomrule
        \end{tabular}
    }
    \label{tab:analysis2}
\end{table}

\section{Details of Prompts to Generative VLM}
This section provides details of the question prompts used for label discovery. To ensure diversity while maintaining the same meaning, we employed five different styles of prompts. Tab.~\ref{tab:prompts} demonstrates the detailed examples of question prompts for generative VLM.

\begin{table}[h!]
    \centering
    \caption{Examples of prompts used for generative VLM.}
    \resizebox{\columnwidth}{!}{
        \begin{tabular}{p{0.9\columnwidth}}
            \toprule
            ``What is the exact name of the object in the painting? Please tell me." \\ 
            ``What can you identify in the illustration? Please tell me in a word." \\ 
            ``What is the specific name of the object captured in the photograph? Please tell me." \\ 
            ``What appears in the picture with its accurate name in the drawing? Please tell me." \\
            ``What is portrayed in the photo with its exact name? Please tell me." \\
            \bottomrule
        \end{tabular}
    }
    \label{tab:prompts}
\end{table}

\section{Time Complexity of Our Method}
\begin{figure}[t] \centering \includegraphics[width=0.7\textwidth]{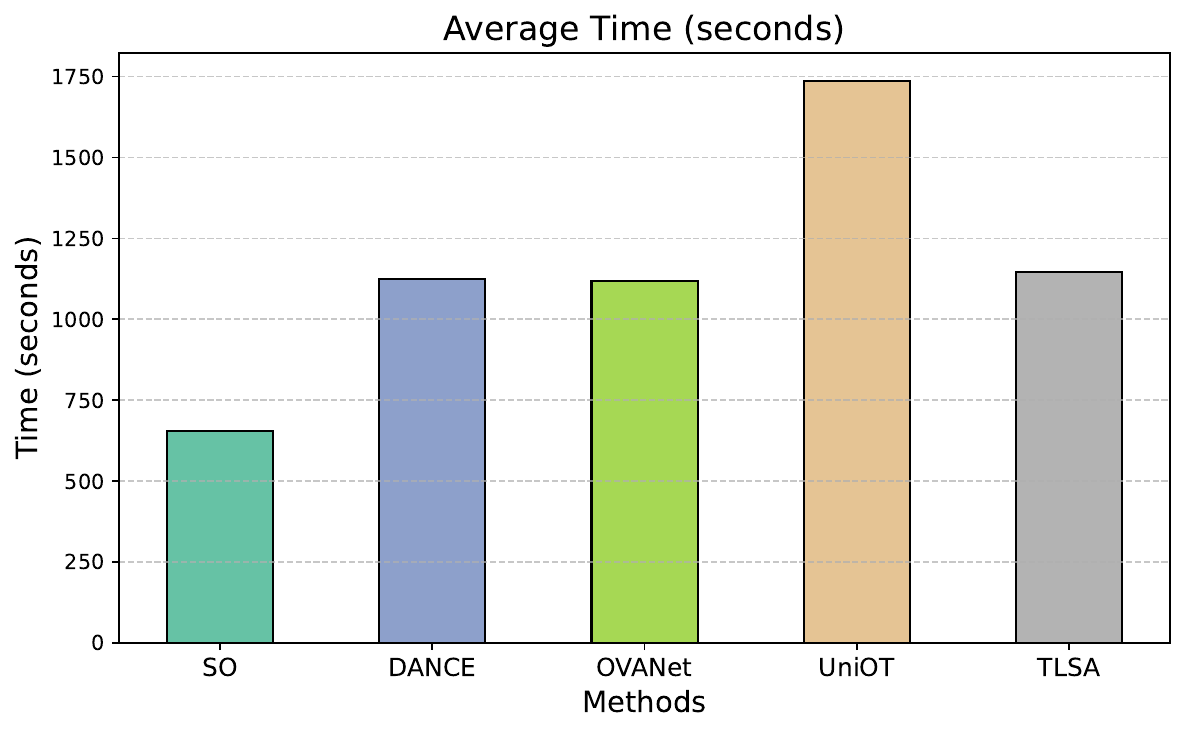} \caption{Average time spent on adaptation.} \label{fig:time} \end{figure}

Fig.~\ref{fig:time} illustrates the time complexity of our approach. 
Although two VLMs are employed, the framework is entirely training-free, introducing minimal overhead compared to prior works. 
Furthermore, it achieves faster adaptation than clustering-based methods~\citep{chang2022unified}.

A potential limitation lies in scalability, since generative VLM inference is required for each target-domain sample, leading to a runtime that grows linearly with dataset size. 
In practice, however, the cost remains manageable: even on DomainNet, the largest dataset in DomainBed, inference completed within 1.5 hours on a single GPU, demonstrating practical feasibility.

\section{Details of Main Results}
This section details the experimental results (H-score, H$^3$-score) for domain-to-domain. Tab.~\ref{tab:office31-CLIP-10-0} through Tab.~\ref{tab:DomainNet-CLIP-345-0} offer a comprehensive comparison between baseline methods and the proposed approach using the CLIP ViT-B/16 backbone.


\begin{table*}[tp]
    \centering
    \caption{Office31 ViT-B/16 OPDA}
    \begin{adjustbox}{width=0.5\textwidth,center}
    \begin{tabular}{c cccccc c} \toprule
Methods	&	A2D	&	A2W	&	D2A	&	D2W	&	W2A	&	W2D	&	Avg	\\ \midrule
\multicolumn{8}{c}{H-score} \\ \midrule
SO	&	84.4 & 79.0 & 88.1 & 93.3 & 80.2 & 89.8 & 85.8	\\
DANCE	&	91.6 & 84.7 & \textbf{93.7} & \textbf{95.8} & 87.8 & \textbf{96.4} & 91.7	\\
OVANet	&	86.8 & 76.1 & 90.1 & 94.9 & 91.0 & 95.2 & 89.0 \\
UniOT	&	78.5 & 73.4 & 82.5 & 88.4 & 83.0 & 88.1 & 82.3	\\
CLIP-Distill&	81.6 & 81.1 & 87.7 & 81.0 & 87.7 & 81.8 & 83.5 \\
MemSPM+DCC	&	88.0	&	 84.6	&	88.7	&	 87.6	&	 87.9	& 94.3 & 88.5	\\ \hline
TLSA	&	92.7	&	86.8	&	89.7	&	86.8	&	89.7	&	92.7	&	89.9	\\ 
TLSA+ST	&	\textbf{95.4} & \textbf{ 88.7} & 91.5 & 88.7 & \textbf{91.5} & 95.4 & \textbf{91.9}	\\ \midrule
\multicolumn{8}{c}{H$^3$-score} \\ \midrule
SO	&	85.6 & 81.7 & 78.5 & 91.4 & 74.1 & 89.2 & 83.4	\\
DANCE	&	90.4 & 85.7 & 81.4 & \textbf{93.0} & 78.3 & \textbf{93.4} & \textbf{87.0}	\\
OVANet	&	87.2 & 79.6 & 79.5 & 92.4 & 80.0 & 92.6 & 85.2
	\\
UniOT	&	80.5 & 76.9 & 75.2 & 88.5 & 75.4 & 87.7 & 80.7
	\\ 
CLIP-Distill	&83.7 & 83.2 & 78.2 & 83.2 & 78.2 & 83.8 & 81.7	 \\ \hline
TLSA	&	\textbf{91.1}	&	\textbf{87.1}	&	79.4	&	87.1	&	79.4	&	91.1	&	85.9	\\ 
TLSA+ST	&	88.7 & 85.2 & \textbf{81.9} & 85.2 & \textbf{81.9} & 88.7 & 85.3 \\
\bottomrule
    \end{tabular}
    \end{adjustbox}
    \label{tab:office31-CLIP-10-10}
\end{table*}

\begin{table*}[tp]
    \centering
    \caption{Office31 ViT-B/16 ODA}
    \begin{adjustbox}{width=0.5\textwidth,center}
    \begin{tabular}{c cccccc c} \toprule
Methods	&	A2D	&	A2W	&	D2A	&	D2W	&	W2A	&	W2D	&	Avg	\\ \midrule
\multicolumn{8}{c}{H-score} \\ \midrule
SO	&	87.2 & 87.9 & 85.1 & 90.7 & 84.1 & 86.0 & 86.8	\\
DANCE	&	92.4 & 89.3 & 93.4 & 96.7 & 91.7 & \textbf{95.6} & 93.2	\\
OVANet	&	86.7 & 80.4 & 91.7 & \textbf{97.4} & 84.4 & 95.1 & 89.3 	\\
UniOT	&	91.4 & 91.0 & 92.4 & 96.8 & 94.2 & 97.7 & 93.9	\\
CLIP-Distill&80.1 & 79.7 & 88.0 & 81.2 & 90.2 & 86.5 & 84.3 \\ \hline
TLSA	&	92.2	&	91.7	&	94.1	&	91.7	&	94.1  &	92.2	&	92.7	\\ 
TLSA+Self training	&	\textbf{94.3} & \textbf{95.4} & \textbf{95.9} & 95.4 & \textbf{95.9} & 94.3 & \textbf{95.2}	\\ \midrule
\multicolumn{8}{c}{H$^3$-score} \\ \midrule
SO	&	87.8 & 88.8 & 80.5 & 90.6 & 79.9 & 87.0 & 85.8	\\
DANCE	&	91.3 & 89.7 & 85.3 & 94.6 & 84.3 & \textbf{93.4} & 89.8	\\
OVANet	&	87.5 & 83.5 & 84.4 & \textbf{95.0} & 80.1 & 93.1 & 87.3	\\
UniOT	&	90.7 & 89.7 & 83.8 & 94.2 & 85.1 & 95.5 & 89.8	\\ 
CLIP-Distill	&	82.9 & 83.0 & 82.2 & 84.1 & 83.4 & 87.4 & 83.8	 \\ \hline
TLSA	&	\textbf{91.2}	&	91.1	&	85.7	&	91.1	&	85.7	&	91.2	&	89.3	\\ 
TLSA+ST	&	90.9 & \textbf{91.6} & \textbf{87.1} & 91.6 & \textbf{87.1} & 90.9 & \textbf{89.9}	\\
\bottomrule 
    \end{tabular}
    \end{adjustbox}
    \label{tab:office31-CLIP-10-0}
\end{table*}

\begin{table*}[tp]
    \centering
    \caption{Office31 ViT-B/16 PDA}
    \begin{adjustbox}{width=0.5\textwidth,center}
    \begin{tabular}{c cccccc c} \toprule
Methods	&	A2D	&	A2W	&	D2A	&	D2W	&	W2A	&	W2D	&	Avg	\\ \midrule
\multicolumn{8}{c}{H-score} \\ \midrule
SO	&	77.0 & 67.6 & 87.2 & 96.9 & 78.5 & 95.4 & 83.7	\\
DANCE	&	51.7 & 49.3 & 86.2 & 89.3 & 67.2 & \textbf{99.0} & 73.8	\\
OVANet	&	74.6 & 64.4 & 87.7 & 97.2 & 82.7 & 97.6 & 84.0 \\
UniOT	&	28.9 & 26.4 & 27.8 & 38.1 & 29.5 & 38.3 & 31.5 \\
CLIP-Distill	&	82.6 & 82.7 & 91.5 & 82.3 & 92.1 & 82.6 & 85.6\\ \hline
TLSA	&	92.7	&	86.8	&	89.7	&	86.8	&	89.7	&	92.7	&	89.9	\\ 
TLSA+ST	&	\textbf{96.2} & \textbf{98.4} & \textbf{96.1} & \textbf{98.4} & \textbf{96.1} & 96.2 & \textbf{96.9}	\\ 
\bottomrule
    \end{tabular}
    \end{adjustbox}
    \label{tab:office31-CLIP-10-21}
\end{table*}

\begin{table*}[tp]
    \centering
    \caption{Office31 ViT-B/16 CDA}
    \begin{adjustbox}{width=0.5\textwidth,center}
    \begin{tabular}{c cccccc c} \toprule
Methods	&	A2D	&	A2W	&	D2A	&	D2W	&	W2A	&	W2D	&	Avg	\\ \midrule
\multicolumn{8}{c}{H-score} \\ \midrule
SO	&	63.4 & 57.5 & 58.1 & 93.3 & 56.1 & 96.2 & 70.8 \\
DANCE	&  58.5 & 50.7 & 53.7 & 93.3 & 52.3 & 97.6 & 67.7 \\
OVANet	&	56.5 & 51.8 & 51.3 & 89.1 & 51.2 & 95.7 & 65.9 \\
UniOT	&	51.9 & 49.8 & 48.4 & 65.3 & 46.2 & 63.7 & 54.2 \\
CLIP-Distill	&	75.5 & 72.0 & 73.6 & 71.4 & 74.5 & 75.4 & 73.7 \\ \midrule
TLSA	&	82.7 & 80.5 & 79.4 & 80.5 & 79.4 & 82.7 & 79.0 \\ 
TLSA+ST	&	\textbf{84.6} & \textbf{84.0} & \textbf{80.1} & \textbf{84.0} & \textbf{80.1} & \textbf{84.6} & \textbf{82.9}
	\\ 
\bottomrule
    \end{tabular}
    \end{adjustbox}
    \label{tab:office31-CLIP-31-0}
\end{table*}

\begin{table*}[tp]
    \centering
    \caption{Office-Home ViT-B/16 OPDA}
    \begin{adjustbox}{width=0.8\textwidth,center}
    \begin{tabular}{c cccccccccccc c} \toprule
Methods	&	A2C	&	A2P	&	A2R	&	C2A	&	C2P	&	C2R	& P2A & P2C & P2R & R2A & R2C & R2P &	Avg	\\ \midrule
\multicolumn{14}{c}{H-score} \\ \midrule
SO	&	69.8 & 80.3 & 83.8 & 80.5 & 81.5 & 85.3 & 77.5 & 71.2 & 84.7 & 80.6 & 76.0 & 77.6 & 79.0\\
DANCE	&	78.0 & 86.4 & 87.1 & 81.1 & 85.2 & 85.6 & 82.4 & 79.9 & 90.5 & 83.6 & 79.6 & 84.1 & 83.6 \\
OVANet	&	72.7 & 88.0 & 86.4 & 80.6 & 84.0 & 85.3 & 75.2 & 63.6 & 89.1 & 81.8 & 74.4 & 85.0 & 80.5 \\
UniOT	&	85.5 & 89.8 & 91.1 & 79.9 & 87.3 & 89.3 & 80.8 & 83.5 & 88.9 & 83.5 & 84.3 & 91.6 & 86.3 \\
CLIP-Distill	&81.4 & 79.6 & 86.1 & 87.9 & 80.3 & 86.7 & 87.0 & 81.3 & 91.0 & 87.0 & 81.9 & 86.8 & 84.7 \\
MemSPM+DCC	&	78.1	&	90.3	&	90.7	
&	81.9	&	90.5	&	88.3 
&	79.2	&	77.4	&	87.8 
&	 78.8	&	76.2	&	 91.6	&	 84.2  \\ \hline
TLSA	&	87.4	&	95.2	&	95.8	
&	86.8	&	95.2	&	95.8 
&	86.8	&	87.4	&	95.8	
&	86.8	&	87.4	&	95.2	&	91.3	\\
TLSA+ST	&	\textbf{90.2} & \textbf{96.3} & \textbf{96.7} & \textbf{90.6} & \textbf{96.3} & \textbf{96.7} & \textbf{90.6} &\textbf{ 90.2 }& \textbf{96.7} & \textbf{90.6} & \textbf{90.2} & \textbf{96.3} & \textbf{93.4 }\\ \midrule
\multicolumn{14}{c}{H$^3$-score} \\ \midrule
SO	&	67.5 & 82.4 & 82.7 & 73.7 & 83.2 & 83.7 & 71.9 & 68.4 & 83.3 & 73.7 & 71.3 & 80.5 & 76.9 \\
DANCE	&72.4 & 86.6 & 84.8 & 74.0 & 85.8 & 83.8 & 74.7 & 73.5 & 86.9 & 75.4 & 73.3 & 85.0 & 79.7 \\
OVANet	&	69.3 & 87.7 & 84.4 & 73.7 & 85.0 & 83.7 & 70.6 & 63.5 & 86.1 & 74.4 & 70.3 & 85.7 & 77.9 \\
UniOT	&78.9 & 89.4 & 88.1 & 75.9 & 87.9 & 87.1 & 76.2 & 76.9 & 86.6 & 77.8 & 77.9 & 90.5 & 82.8 \\ 
CLIP-Distill	&74.4 & 81.9 & 84.2 & 77.6 & 82.4 & 84.6 & 77.1 & 74.3 & 87.3 & 77.1 & 74.6 & 86.9 & 80.2 \\ \hline
TLSA	&	79.8	&	92.2	&	89.5	
&	76.8	&	92.7	&	89.5 
&	76.8	&	79.8	&	89.5	
&	76.8	&	79.8	&	92.2	&	84.6	\\
TLSA+ST	&	\textbf{85.5} & \textbf{93.9} & \textbf{94.1} & \textbf{87.5} & \textbf{93.9} & \textbf{94.1 }& \textbf{87.5} & \textbf{85.5} & \textbf{94.1} & \textbf{87.5} & \textbf{85.5} & \textbf{93.9 }& \textbf{90.2}	\\
\bottomrule
    \end{tabular}
    \end{adjustbox}
    \label{tab:officehome-CLIP-10-5}
\end{table*}

\begin{table*}[tp]
    \centering
    \caption{Office-Home ViT-B/16 ODA}
    \begin{adjustbox}{width=0.8\textwidth,center}
    \begin{tabular}{c cccccccccccc c} \toprule
Methods	&	A2C	&	A2P	&	A2R	&	C2A	&	C2P	&	C2R	& P2A & P2C & P2R & R2A & R2C & R2P &	Avg	\\ \midrule
\multicolumn{14}{c}{H-score} \\ \midrule
SO	&	69.6 & 77.3 & 80.5 & 76.8 & 81.0 & 82.2 & 70.2 & 69.7 & 81.4 & 78.7 & 73.2 & 77.1 & 76.5 \\
DANCE	 & 75.2 & 81.5 & 82.9 & 76.2 & 77.1 & 81.6 & 70.2 & 71.5 & 85.6 & 80.0 & 74.8 & 82.6 & 78.3 \\
OVANet	&	71.6 & 81.7 & 82.7 & 76.5 & 83.3 & 81.6 & 64.0 & 56.9 & 84.1 & 78.6 & 70.5 & 84.4 & 76.3 \\
UniOT	&81.0 & 85.9 & 85.0 & 76.8 & 85.5 & 83.9 & 72.1 & 77.8 & 83.7 & 80.4 & 79.3 & 89.7 & 81.7 \\
CLIP-Distill	&77.3 & 79.1 & 84.6 & 83.4 & 79.7 & 85.1 & 79.6 & 75.9 & 88.6 & 79.7 & 76.9 & 85.9 & 81.3 \\
MemSPM+DCC	&	69.7	&	 83.2	&	 85.2	
&	72.0	&	79.2	&	 81.2
&	72.3	&	66.7	&	 85.2	
&	 72.7	&	 66.0	&	84.5
&	76.5 \\ \hline
TLSA	&	85.5	&	92.3	&	93.9	
&	85.1	&	92.3	&	93.9
&	85.1	&	85.5	&	93.9
&	85.1	&	85.5	&	92.3
&	89.2 \\
TLSA+ST &\textbf{89.2} & \textbf{94.4} & \textbf{94.2} & \textbf{88.0} & \textbf{94.4 }& \textbf{94.2} & \textbf{88.0} & \textbf{89.2} & \textbf{94.2} & \textbf{88.0} & \textbf{89.2} & \textbf{94.4} & \textbf{91.4} \\ \midrule
\multicolumn{14}{c}{H$^3$-score} \\ \midrule
SO	&67.4 & 80.3 & 80.5 & 71.5 & 82.9 & 81.7 & 67.6 & 67.5 & 81.2 & 72.6 & 69.6 & 80.2 & 75.2 \\
DANCE	&70.8 & 83.3 & 82.1 & 71.2 & 80.1 & 81.3 & 67.6 & 68.6 & 83.9 & 73.4 & 70.5 & 84.1 & 76.4 \\
OVANet	&68.7 & 83.4 & 82.0 & 71.4 & 84.5 & 81.3 & 63.7 & 58.9 & 82.9 & 72.6 & 68.0 & 85.2 & 75.2 \\
UniOT	&75.7 & 86.6 & 83.9 & 73.4 & 86.8 & 83.3 & 70.2 & 73.7 & 83.1 & 76.0 & 74.6 & 89.1 & 79.7 \\ 
CLIP-Distill	&72.0 & 81.6 & 83.3 & 75.2 & 82.0 & 83.6 & 73.1 & 71.2 & 85.8 & 73.2 & 71.8 & 86.3 & 78.3 \\ \hline
TLSA	&	78.7	&	90.4	&	88.3
&	75.9	&	90.4	&	88.3
&	75.9	&	78.7	&	88.3	
&	75.9	&	78.7	&	90.4	&	83.3	\\
TLSA+ST	&\textbf{85.0} &\textbf{ 92.8} & \textbf{92.3} & \textbf{85.7} & \textbf{92.8} & \textbf{92.3}& \textbf{85.7} & \textbf{85.0} & \textbf{92.3} & \textbf{85.7} & \textbf{85.0} & \textbf{92.8} & \textbf{89.0} \\
\bottomrule
    \end{tabular}
    \end{adjustbox}
    \label{tab:officehome-CLIP-15-0}
\end{table*}

\begin{table*}[tp]
    \centering
    \caption{Office-Home ViT-B/16 PDA}
    \begin{adjustbox}{width=0.8\textwidth,center}
    \begin{tabular}{c cccccccccccc c} \toprule
Methods	&	A2C	&	A2P	&	A2R	&	C2A	&	C2P	&	C2R	& P2A & P2C & P2R & R2A & R2C & R2P &	Avg	\\ \midrule
\multicolumn{14}{c}{H-score} \\ \midrule
SO	&37.8 & 58.5 & 63.7 & 38.6 & 56.4 & 59.4 & 41.0 & 32.9 & 68.5 & 50.2 & 43.4 & 71.1 & 51.8 \\
DANCE	&34.3 & 52.6 & 66.0 & 30.2 & 41.5 & 59.1 & 30.7 & 19.9 & 68.2 & 47.9 & 39.5 & 72.4 & 46.8 \\
OVANet	&51.5 & 62.7 & 73.1 & 52.3 & 59.4 & 65.7 & 44.8 & 35.4 & 73.3 & 61.7 & 49.1 & 76.8 & 58.8 \\
UniOT	&31.8 & 34.3 & 31.1 & 26.0 & 32.2 & 29.4 & 26.9 & 30.0 & 33.1 & 36.9 & 31.5 & 38.4 & 31.8 \\
CLIP-Distill	&64.9 & 86.7 & 87.6 & 71.5 & 85.6 & 86.5 & 63.3 & 57.2 & 82.9 & 65.7 & 59.0 & 82.5 & 74.4\\
MemSPM+DCC	&	64.7	&	81.1	&	84.5	
&	74.8	&	 74.7 	&	77.5	 
&	58.7	&	60.3	&	84.2	
&	 70.3	&	77.2	&	 85.8	&	74.5	\\ \hline
TLSA	&	74.5	&	\textbf{87.8}	&	90.4	
&	75.3	&	\textbf{87.8}	&	90.4	
&	75.3	&	74.5	&	90.4	
&	75.3	&	74.5	&	\textbf{87.8}	&	\textbf{82.0}	\\
TLSA+ST	&\textbf{79.2} & 86.3 & \textbf{93.5} & \textbf{82.2} & 86.3 & \textbf{93.5} & \textbf{82.2} & \textbf{79.2} & \textbf{93.5} & \textbf{82.2} & \textbf{79.2} & 86.3 & \textbf{85.3}	\\
\bottomrule
    \end{tabular}
    \end{adjustbox}
    \label{tab:officehome-CLIP-25-40}
\end{table*}

\begin{table*}[tp]
    \centering
    \caption{Office-Home ViT-B/16 CDA}
    \begin{adjustbox}{width=0.8\textwidth,center}
    \begin{tabular}{c cccccccccccc c} \toprule
Methods	&	A2C	&	A2P	&	A2R	&	C2A	&	C2P	&	C2R	& P2A & P2C & P2R & R2A & R2C & R2P &	Avg	\\ \midrule
\multicolumn{14}{c}{H-score} \\ \midrule
SO	&31.3 & 57.0 & 59.8 & 36.8 & 60.9 & 56.5 & 35.6 & 28.3 & 64.9 & 50.2 & 36.3 & 77.5 & 49.6 \\
DANCE	&31.1 & 57.1 & 60.6 & 36.7 & 58.3 & 56.7 & 33.1 & 28.5 & 64.6 & 50.5 & 36.2 & 78.8 & 49.4 \\
OVANet	&41.5 & 60.5 & 70.1 & 49.8 & 61.6 & 64.2 & 42.3 & 30.7 & 70.4 & 60.2 & 40.8 & 81.5 & 56.1 \\
UniOT	&41.4 & 46.9 & 50.9 & 41.7 & 50.1 & 50.6 & 36.8 & 38.0 & 50.0 & 46.2 & 42.9 & 61.5 & 46.4 \\
CLIP-Distill&	53.1 & 85.9 & 83.2 & 65.1 & 84.9 & 82.6 & 59.2 & 45.7 & 78.5 & 60.4 & 47.8 & 82.2 & 69.0 \\ \hline
TLSA	&	66.4	&	87.2	&	88.2
&	74.2	&	87.2	&	88.2 &	74.2	&	66.4	&	88.2	
&	74.2	&	66.4	&	87.2	&	79.0	\\
TLSA+ST	&\textbf{71.3} &\textbf{ 88.6 }& \textbf{89.9 }& \textbf{77.3} & \textbf{88.6} &\textbf{ 89.9} & \textbf{77.3} & \textbf{71.3 }&\textbf{ 89.9} & \textbf{77.3} & \textbf{71.3 } & \textbf{88.6} & \textbf{81.8} \\ 
\bottomrule
    \end{tabular}
    \end{adjustbox}
    \label{tab:officehome-CLIP-65-0}
\end{table*}

\begin{table*}[tp]
    \centering
    \caption{DomainNet ViT-B/16 OPDA}  
    \begin{adjustbox}{width=0.5\textwidth,center}
    \begin{tabular}{c cccccc c} \toprule
Methods	&	P2R	&	P2S	&	R2P	&	R2S	&	S2P	&	S2R	&	Avg	\\ \midrule
\multicolumn{8}{c}{H-score} \\ \midrule
SO	&64.2 & 50.2 & 53.1 & 52.1 & 48.3 & 67.1 & 55.8 \\
DANCE	&64.2 & 50.1 & 50.8 & 51.2 & 47.4 & 67.2 & 55.1 	\\
OVANet	&68.2 & 58.6 & 60.5 & 59.1 & 56.9 & 68.6 & 62.0 \\
UniOT	&71.4 & 62.8 & 62.5 & 64.6 & 59.5 & 72.7 & 65.6 \\
CLIP-Distill	&74.1 & 61.6 & 58.5 & 60.2 & 59.7 & 74.8 & 64.8 \\
MemSPM+DCC	&	62.4	&	52.8
&	58.5	&	 53.3	
&	 50.4	&	 62.6	&	56.7	\\ \hline
TLSA	&	\textbf{83.6}	&	\textbf{71.2}	
&	\textbf{71.3}	&	\textbf{71.2}	
&	\textbf{71.3}	&	\textbf{83.6}	&	\textbf{75.4}	\\ 
TLSA+ST	&82.3 & 69.4 & 71.2 & 69.4 & 71.2 & 82.3 & 74.3 \\  \midrule
\multicolumn{8}{c}{H$^3$-score} \\ \midrule
SO	&69.3 & 52.5 & 57.2 & 53.8 & 53.4 & 71.5 & 59.6 	\\ 
DANCE	&69.3 & 52.4 & 55.4 & 53.2 & 52.6 & 71.5 & 59.1 \\
OVANet	&72.3 & 58.3 & 62.7 & 58.6 & 60.1 & 72.6 & 64.1 \\ 
UniOT	&73.9 & 61.6 & 64.0 & 63.3 & 61.6 & 75.1 & 66.6 \\
CLIP-Distill	&76.6 & 60.2 & 61.2 & 59.4 & 62.1 & 77.1 & 66.1 \\ \hline
TLSA	&	83.1 &	\textbf{69.8}
&	66.1	&	\textbf{69.8}	
&	66.1	&	83.1&	73.1	\\ 
TLSA+ST	&\textbf{84.0 }& 69.2 & \textbf{72.9} & 69.2 & \textbf{72.9} &\textbf{ 84.0} & \textbf{75.4} \\  
\bottomrule 
    \end{tabular}
    \end{adjustbox}
    \label{tab:DomainNet-CLIP-150-50}
\end{table*}

\begin{table*}[tp]
    \centering
    \caption{DomainNet ViT-B/16 ODA}
    \begin{adjustbox}{width=0.5\textwidth,center}
    \begin{tabular}{c cccccc c} \toprule
Methods	&	P2R	&	P2S	&	R2P	&	R2S	&	S2P	&	S2R	&	Avg	\\ \midrule
\multicolumn{8}{c}{H-score} \\ \midrule
SO	&68.5 & 55.0 & 57.0 & 56.6 & 51.3 & 70.0 & 59.7 \\
DANCE	&68.6 &	54.7&	55.5&	56.3&	51.1&	70.2&	59.4	\\
OVANet	&70.4	&61.4	&62.5	&62.1	&59.4	&70.6	&64.4\\
UniOT	&76.1	&66.9	&66.8	&68.2	&63.3	&75.3	&69.4\\
CLIP-Distill	&76.2	&63.6	&59.7&	62.4	&61.6	&76.5	&66.7\\ \hline
TLSA	&	\textbf{85.6}	&	\textbf{72.9}	
&	\textbf{72.1}	&	\textbf{72.9}	
&	\textbf{72.1}	&	\textbf{85.6}	&	\textbf{76.9}	\\ 
TLSA+ST	&83.1	&70.4	&71.2&	70.4	&71.2	&83.1	&74.9\\  \midrule
\multicolumn{8}{c}{H$^3$-score} \\ \midrule
SO	&	72.2	&56.6	&59.6	&57.8	&55.3	&73.3	&62.5\\
DANCE	&72.3	&56.5	&58.5	&57.6	&55.2	&73.4	&62.2\\ 
OVANet	&73.7	&61.0	&63.4	&61.5	&61.3	&73.8	&65.8\\
UniOT	&76.7	&65.0	&66.3	&66.3	&63.6	&76.3	&69.0\\ 
CLIP-Distill	&77.7	&62.5	&61.5	&61.7	&62.8	&77.9	&67.4\\ \hline
TLSA	&	83.7	&	68.1
&	69.5	&	68.1	
&	69.5	&	83.7	&	73.8	\\ 
TLSA+ST	&\textbf{83.9}	&\textbf{70.3}	&\textbf{72.3}	&\textbf{70.3}	&\textbf{72.3}	&\textbf{83.9} &\textbf{75.5}\\
\bottomrule 
    \end{tabular}
    \end{adjustbox}
    \label{tab:DomainNet-CLIP-150-0}
\end{table*}

\begin{table*}[tp]
    \centering
    \caption{DomainNet ViT-B/16 PDA}
    \begin{adjustbox}{width=0.5\textwidth,center}
    \begin{tabular}{c cccccc c} \toprule
Methods	&	P2R	&	P2S	&	R2P	&	R2S	&	S2P	&	S2R	&	Avg	\\ \midrule
\multicolumn{8}{c}{H-score} \\ \midrule
SO	& 44.4 & 26.5 & 30.0 & 28.6 & 24.0 & 44.0 & 32.9 \\
DANCE	& 43.2 & 25.3 & 26.3 & 25.4 & 20.2 & 39.8 & 30.0 \\
OVANet	& 55.8 & 36.9 & 41.1 & 39.1 & 38.6 & 58.4 & 45.0 \\
UniOT	& 46.5 & 39.4 & 39.1 & 41.4 & 33.6 & 45.2 & 40.9 \\
CLIP-Distill	& 85.0 & 75.0 & 49.1 & 47.9 & 46.4 & 75.0 & 63.1 \\ \hline
TLSA	& 84.6 & 64.7 & 60.9 & 64.7 & 60.9 & 84.6 & 70.1 \\ 
TLSA+ST	& \textbf{86.8} & \textbf{70.0} & \textbf{66.4} & \textbf{70.0} & \textbf{66.4} & \textbf{86.8} & \textbf{74.4} \\
\bottomrule 
    \end{tabular}
    \end{adjustbox}
    \label{tab:DomainNet-CLIP-150-195}
\end{table*}

\begin{table*}[tp]
    \centering
    \caption{DomainNet ViT-B/16 CDA}
    \begin{adjustbox}{width=0.5\textwidth,center}
    \begin{tabular}{c cccccc c} \toprule
Methods	&	P2R	&	P2S	&	R2P	&	R2S	&	S2P	&	S2R	&	Avg	\\ \midrule
\multicolumn{8}{c}{H-score} \\ \midrule
SO	&46.9	&26.6	&30.4	&29.3	&26.7	& 48.8	&34.8\\
DANCE	&46.7	&26.5	&29.7	&28.3	&26.1	&47.1	&34.1\\
OVANet	&57.6	&36.9	&40.9	&39.7	&41.2	&60.8	&46.2\\
UniOT	&60.0	&47.1	&46.8	&49.7	&44.6	&60.9	&51.5\\
CLIP-Distill	&71.3	&42.9	&39.5	&42.3	&40.2	&71.5	&51.3\\ \hline
TLSA	&	\textbf{83.7}	&	\textbf{62.6}	
&	60.3	&	\textbf{62.6}	
&	60.3	&	\textbf{83.7}	&	\textbf{68.9}	\\ 
TLSA+ST	&81.0	&61.6	&\textbf{61.1}	&61.6	&\textbf{61.1}	&81.0	&67.9\\
\bottomrule 
    \end{tabular}
    \end{adjustbox}
    \label{tab:DomainNet-CLIP-345-0}
\end{table*}

\clearpage




\bibliographystyle{elsarticle-harv}
\bibliography{elsarticle}

\begin{thebibliography}{73}
\expandafter\ifx\csname natexlab\endcsname\relax\def\natexlab#1{#1}\fi
\providecommand{\url}[1]{\texttt{#1}}
\providecommand{\href}[2]{#2}
\providecommand{\path}[1]{#1}
\providecommand{\DOIprefix}{doi:}
\providecommand{\ArXivprefix}{arXiv:}
\providecommand{\URLprefix}{URL: }
\providecommand{\Pubmedprefix}{pmid:}
\providecommand{\doi}[1]{\href{http://dx.doi.org/#1}{\path{#1}}}
\providecommand{\Pubmed}[1]{\href{pmid:#1}{\path{#1}}}
\providecommand{\bibinfo}[2]{#2}
\ifx\xfnm\relax \def\xfnm[#1]{\unskip,\space#1}\fi
\bibitem[{Achiam et~al.(2023)Achiam, Adler, Agarwal, Ahmad, Akkaya, Aleman, Almeida, Altenschmidt, Altman, Anadkat et~al.}]{achiam2023gpt}
\bibinfo{author}{Achiam, J.}, \bibinfo{author}{Adler, S.}, \bibinfo{author}{Agarwal, S.}, \bibinfo{author}{Ahmad, L.}, \bibinfo{author}{Akkaya, I.}, \bibinfo{author}{Aleman, F.L.}, \bibinfo{author}{Almeida, D.}, \bibinfo{author}{Altenschmidt, J.}, \bibinfo{author}{Altman, S.}, \bibinfo{author}{Anadkat, S.}, et~al., \bibinfo{year}{2023}.
\newblock \bibinfo{title}{Gpt-4 technical report}.
\newblock \bibinfo{journal}{arXiv preprint arXiv:2303.08774} .
\bibitem[{Adila et~al.(2023)Adila, Shin, Cai and Sala}]{adila2023zero}
\bibinfo{author}{Adila, D.}, \bibinfo{author}{Shin, C.}, \bibinfo{author}{Cai, L.}, \bibinfo{author}{Sala, F.}, \bibinfo{year}{2023}.
\newblock \bibinfo{title}{Zero-shot robustification of zero-shot models with foundation models}.
\newblock \bibinfo{journal}{arXiv preprint arXiv:2309.04344} .
\bibitem[{Amini et~al.(2024)Amini, Feofanov, Pauletto, Hadjadj, Devijver and Maximov}]{amini2024self}
\bibinfo{author}{Amini, M.R.}, \bibinfo{author}{Feofanov, V.}, \bibinfo{author}{Pauletto, L.}, \bibinfo{author}{Hadjadj, L.}, \bibinfo{author}{Devijver, E.}, \bibinfo{author}{Maximov, Y.}, \bibinfo{year}{2024}.
\newblock \bibinfo{title}{Self-training: A survey}.
\newblock \bibinfo{journal}{Neurocomputing} , \bibinfo{pages}{128904}.
\bibitem[{Bendale and Boult(2016)}]{bendale2016towards}
\bibinfo{author}{Bendale, A.}, \bibinfo{author}{Boult, T.E.}, \bibinfo{year}{2016}.
\newblock \bibinfo{title}{Towards open set deep networks}, in: \bibinfo{booktitle}{Proceedings of the IEEE conference on computer vision and pattern recognition}, pp. \bibinfo{pages}{1563--1572}.
\bibitem[{Cao et~al.(2018)Cao, Ma, Long and Wang}]{cao2018partial}
\bibinfo{author}{Cao, Z.}, \bibinfo{author}{Ma, L.}, \bibinfo{author}{Long, M.}, \bibinfo{author}{Wang, J.}, \bibinfo{year}{2018}.
\newblock \bibinfo{title}{Partial adversarial domain adaptation}, in: \bibinfo{booktitle}{Proceedings of the European conference on computer vision (ECCV)}, pp. \bibinfo{pages}{135--150}.
\bibitem[{Cen et~al.(2023)Cen, Luan, Zhang, Pei, Zhang, Zhao, Shen and Chen}]{cen2023devil}
\bibinfo{author}{Cen, J.}, \bibinfo{author}{Luan, D.}, \bibinfo{author}{Zhang, S.}, \bibinfo{author}{Pei, Y.}, \bibinfo{author}{Zhang, Y.}, \bibinfo{author}{Zhao, D.}, \bibinfo{author}{Shen, S.}, \bibinfo{author}{Chen, Q.}, \bibinfo{year}{2023}.
\newblock \bibinfo{title}{The devil is in the wrongly-classified samples: Towards unified open-set recognition}.
\newblock \bibinfo{journal}{arXiv preprint arXiv:2302.04002} .
\bibitem[{Chang et~al.(2022)Chang, Shi, Tuan and Wang}]{chang2022unified}
\bibinfo{author}{Chang, W.}, \bibinfo{author}{Shi, Y.}, \bibinfo{author}{Tuan, H.}, \bibinfo{author}{Wang, J.}, \bibinfo{year}{2022}.
\newblock \bibinfo{title}{Unified optimal transport framework for universal domain adaptation}.
\newblock \bibinfo{journal}{Advances in Neural Information Processing Systems} \bibinfo{volume}{35}, \bibinfo{pages}{29512--29524}.
\bibitem[{Chen et~al.(2022a)Chen, Lou, He, Bai and Deng}]{chen2022geometric}
\bibinfo{author}{Chen, L.}, \bibinfo{author}{Lou, Y.}, \bibinfo{author}{He, J.}, \bibinfo{author}{Bai, T.}, \bibinfo{author}{Deng, M.}, \bibinfo{year}{2022}a.
\newblock \bibinfo{title}{Geometric anchor correspondence mining with uncertainty modeling for universal domain adaptation}, in: \bibinfo{booktitle}{Proceedings of the IEEE/CVF Conference on Computer Vision and Pattern Recognition}, pp. \bibinfo{pages}{16134--16143}.
\bibitem[{Chen et~al.(2022b)Chen, Ge, Tong, Wang, Song, Wang and Luo}]{chen2022adaptformer}
\bibinfo{author}{Chen, S.}, \bibinfo{author}{Ge, C.}, \bibinfo{author}{Tong, Z.}, \bibinfo{author}{Wang, J.}, \bibinfo{author}{Song, Y.}, \bibinfo{author}{Wang, J.}, \bibinfo{author}{Luo, P.}, \bibinfo{year}{2022}b.
\newblock \bibinfo{title}{Adaptformer: Adapting vision transformers for scalable visual recognition}.
\newblock \bibinfo{journal}{Advances in Neural Information Processing Systems} \bibinfo{volume}{35}, \bibinfo{pages}{16664--16678}.
\bibitem[{Deng and Jia(2023)}]{deng2023universal}
\bibinfo{author}{Deng, B.}, \bibinfo{author}{Jia, K.}, \bibinfo{year}{2023}.
\newblock \bibinfo{title}{Universal domain adaptation from foundation models}.
\newblock \bibinfo{journal}{arXiv preprint arXiv:2305.11092} .
\bibitem[{Devlin et~al.(2018)Devlin, Chang, Lee and Toutanova}]{devlin2018bert}
\bibinfo{author}{Devlin, J.}, \bibinfo{author}{Chang, M.W.}, \bibinfo{author}{Lee, K.}, \bibinfo{author}{Toutanova, K.}, \bibinfo{year}{2018}.
\newblock \bibinfo{title}{Bert: Pre-training of deep bidirectional transformers for language understanding}.
\newblock \bibinfo{journal}{arXiv preprint arXiv:1810.04805} .
\bibitem[{Dosovitskiy et~al.(2020)Dosovitskiy, Beyer, Kolesnikov, Weissenborn, Zhai, Unterthiner, Dehghani, Minderer, Heigold, Gelly et~al.}]{dosovitskiy2020image}
\bibinfo{author}{Dosovitskiy, A.}, \bibinfo{author}{Beyer, L.}, \bibinfo{author}{Kolesnikov, A.}, \bibinfo{author}{Weissenborn, D.}, \bibinfo{author}{Zhai, X.}, \bibinfo{author}{Unterthiner, T.}, \bibinfo{author}{Dehghani, M.}, \bibinfo{author}{Minderer, M.}, \bibinfo{author}{Heigold, G.}, \bibinfo{author}{Gelly, S.}, et~al., \bibinfo{year}{2020}.
\newblock \bibinfo{title}{An image is worth 16x16 words: Transformers for image recognition at scale}.
\newblock \bibinfo{journal}{arXiv preprint arXiv:2010.11929} .
\bibitem[{Esmaeilpour et~al.(2022)Esmaeilpour, Liu, Robertson and Shu}]{esmaeilpour2022zero}
\bibinfo{author}{Esmaeilpour, S.}, \bibinfo{author}{Liu, B.}, \bibinfo{author}{Robertson, E.}, \bibinfo{author}{Shu, L.}, \bibinfo{year}{2022}.
\newblock \bibinfo{title}{Zero-shot out-of-distribution detection based on the pre-trained model clip}, in: \bibinfo{booktitle}{Proceedings of the AAAI conference on artificial intelligence}, pp. \bibinfo{pages}{6568--6576}.
\bibitem[{Fu et~al.(2020)Fu, Cao, Long and Wang}]{fu2020learning}
\bibinfo{author}{Fu, B.}, \bibinfo{author}{Cao, Z.}, \bibinfo{author}{Long, M.}, \bibinfo{author}{Wang, J.}, \bibinfo{year}{2020}.
\newblock \bibinfo{title}{Learning to detect open classes for universal domain adaptation}, in: \bibinfo{booktitle}{Computer Vision--ECCV 2020: 16th European Conference, Glasgow, UK, August 23--28, 2020, Proceedings, Part XV 16}, \bibinfo{organization}{Springer}. pp. \bibinfo{pages}{567--583}.
\bibitem[{Ganin et~al.(2016)Ganin, Ustinova, Ajakan, Germain, Larochelle, Laviolette, March and Lempitsky}]{ganin2016domain}
\bibinfo{author}{Ganin, Y.}, \bibinfo{author}{Ustinova, E.}, \bibinfo{author}{Ajakan, H.}, \bibinfo{author}{Germain, P.}, \bibinfo{author}{Larochelle, H.}, \bibinfo{author}{Laviolette, F.}, \bibinfo{author}{March, M.}, \bibinfo{author}{Lempitsky, V.}, \bibinfo{year}{2016}.
\newblock \bibinfo{title}{Domain-adversarial training of neural networks}.
\newblock \bibinfo{journal}{Journal of machine learning research} \bibinfo{volume}{17}, \bibinfo{pages}{1--35}.
\bibitem[{Ge et~al.(2023)Ge, Ren, Gallagher, Wang, Yang, Adam, Itti, Lakshminarayanan and Zhao}]{ge2023improving}
\bibinfo{author}{Ge, Y.}, \bibinfo{author}{Ren, J.}, \bibinfo{author}{Gallagher, A.}, \bibinfo{author}{Wang, Y.}, \bibinfo{author}{Yang, M.H.}, \bibinfo{author}{Adam, H.}, \bibinfo{author}{Itti, L.}, \bibinfo{author}{Lakshminarayanan, B.}, \bibinfo{author}{Zhao, J.}, \bibinfo{year}{2023}.
\newblock \bibinfo{title}{Improving zero-shot generalization and robustness of multi-modal models}, in: \bibinfo{booktitle}{Proceedings of the IEEE/CVF Conference on Computer Vision and Pattern Recognition}, pp. \bibinfo{pages}{11093--11101}.
\bibitem[{Ge et~al.(2017)Ge, Demyanov, Chen and Garnavi}]{ge2017generative}
\bibinfo{author}{Ge, Z.}, \bibinfo{author}{Demyanov, S.}, \bibinfo{author}{Chen, Z.}, \bibinfo{author}{Garnavi, R.}, \bibinfo{year}{2017}.
\newblock \bibinfo{title}{Generative openmax for multi-class open set classification}.
\newblock \bibinfo{journal}{arXiv preprint arXiv:1707.07418} .
\bibitem[{Goyal et~al.(2017)Goyal, Khot, Summers-Stay, Batra and Parikh}]{balanced_vqa_v2}
\bibinfo{author}{Goyal, Y.}, \bibinfo{author}{Khot, T.}, \bibinfo{author}{Summers-Stay, D.}, \bibinfo{author}{Batra, D.}, \bibinfo{author}{Parikh, D.}, \bibinfo{year}{2017}.
\newblock \bibinfo{title}{Making the v in vqa matter: Elevating the role of image understanding in visual question answering}, in: \bibinfo{booktitle}{Proceedings of the IEEE conference on computer vision and pattern recognition}, pp. \bibinfo{pages}{6904--6913}.
\bibitem[{Hartigan and Wong(1979)}]{hartigan1979algorithm}
\bibinfo{author}{Hartigan, J.A.}, \bibinfo{author}{Wong, M.A.}, \bibinfo{year}{1979}.
\newblock \bibinfo{title}{Algorithm as 136: A k-means clustering algorithm}.
\newblock \bibinfo{journal}{Journal of the royal statistical society. series c (applied statistics)} \bibinfo{volume}{28}, \bibinfo{pages}{100--108}.
\bibitem[{He et~al.(2016)He, Zhang, Ren and Sun}]{he2016deep}
\bibinfo{author}{He, K.}, \bibinfo{author}{Zhang, X.}, \bibinfo{author}{Ren, S.}, \bibinfo{author}{Sun, J.}, \bibinfo{year}{2016}.
\newblock \bibinfo{title}{Deep residual learning for image recognition}, in: \bibinfo{booktitle}{Proceedings of the IEEE conference on computer vision and pattern recognition}, pp. \bibinfo{pages}{770--778}.
\bibitem[{Jia et~al.(2021)Jia, Yang, Xia, Chen, Parekh, Pham, Le, Sung, Li and Duerig}]{jia2021scaling}
\bibinfo{author}{Jia, C.}, \bibinfo{author}{Yang, Y.}, \bibinfo{author}{Xia, Y.}, \bibinfo{author}{Chen, Y.T.}, \bibinfo{author}{Parekh, Z.}, \bibinfo{author}{Pham, H.}, \bibinfo{author}{Le, Q.}, \bibinfo{author}{Sung, Y.H.}, \bibinfo{author}{Li, Z.}, \bibinfo{author}{Duerig, T.}, \bibinfo{year}{2021}.
\newblock \bibinfo{title}{Scaling up visual and vision-language representation learning with noisy text supervision}, in: \bibinfo{booktitle}{International conference on machine learning}, \bibinfo{organization}{PMLR}. pp. \bibinfo{pages}{4904--4916}.
\bibitem[{Karim et~al.(2023)Karim, Mithun, Rajvanshi, Chiu, Samarasekera and Rahnavard}]{karim2023c}
\bibinfo{author}{Karim, N.}, \bibinfo{author}{Mithun, N.C.}, \bibinfo{author}{Rajvanshi, A.}, \bibinfo{author}{Chiu, H.p.}, \bibinfo{author}{Samarasekera, S.}, \bibinfo{author}{Rahnavard, N.}, \bibinfo{year}{2023}.
\newblock \bibinfo{title}{C-sfda: A curriculum learning aided self-training framework for efficient source free domain adaptation}, in: \bibinfo{booktitle}{Proceedings of the IEEE/CVF Conference on Computer Vision and Pattern Recognition}, pp. \bibinfo{pages}{24120--24131}.
\bibitem[{Kim et~al.(2024)Kim, Yoon, In, Moon, Kim and Park}]{kim2024adaptive}
\bibinfo{author}{Kim, K.}, \bibinfo{author}{Yoon, K.}, \bibinfo{author}{In, Y.}, \bibinfo{author}{Moon, J.}, \bibinfo{author}{Kim, D.}, \bibinfo{author}{Park, C.}, \bibinfo{year}{2024}.
\newblock \bibinfo{title}{Adaptive self-training framework for fine-grained scene graph generation}.
\newblock \bibinfo{journal}{arXiv preprint arXiv:2401.09786} .
\bibitem[{Kumar et~al.(2022)Kumar, Raghunathan, Jones, Ma and Liang}]{kumar2022fine}
\bibinfo{author}{Kumar, A.}, \bibinfo{author}{Raghunathan, A.}, \bibinfo{author}{Jones, R.}, \bibinfo{author}{Ma, T.}, \bibinfo{author}{Liang, P.}, \bibinfo{year}{2022}.
\newblock \bibinfo{title}{Fine-tuning can distort pretrained features and underperform out-of-distribution}.
\newblock \bibinfo{journal}{arXiv preprint arXiv:2202.10054} .
\bibitem[{Lai et~al.(2023)Lai, Liu, Zhou and Zhou}]{lai2023memory}
\bibinfo{author}{Lai, Y.}, \bibinfo{author}{Liu, X.}, \bibinfo{author}{Zhou, T.}, \bibinfo{author}{Zhou, Y.}, \bibinfo{year}{2023}.
\newblock \bibinfo{title}{Memory-assisted sub-prototype mining for universal domain adaptation}.
\newblock \bibinfo{journal}{arXiv preprint arXiv:2310.05453} .
\bibitem[{Li et~al.(2021a)Li, Kang, Zhu, Wei and Yang}]{li2021domain}
\bibinfo{author}{Li, G.}, \bibinfo{author}{Kang, G.}, \bibinfo{author}{Zhu, Y.}, \bibinfo{author}{Wei, Y.}, \bibinfo{author}{Yang, Y.}, \bibinfo{year}{2021}a.
\newblock \bibinfo{title}{Domain consensus clustering for universal domain adaptation}, in: \bibinfo{booktitle}{Proceedings of the IEEE/CVF conference on computer vision and pattern recognition}, pp. \bibinfo{pages}{9757--9766}.
\bibitem[{Li et~al.(2023)Li, Li, Savarese and Hoi}]{li2023blip}
\bibinfo{author}{Li, J.}, \bibinfo{author}{Li, D.}, \bibinfo{author}{Savarese, S.}, \bibinfo{author}{Hoi, S.}, \bibinfo{year}{2023}.
\newblock \bibinfo{title}{Blip-2: Bootstrapping language-image pre-training with frozen image encoders and large language models}, in: \bibinfo{booktitle}{International conference on machine learning}, \bibinfo{organization}{PMLR}. pp. \bibinfo{pages}{19730--19742}.
\bibitem[{Li et~al.(2022)Li, Li, Xiong and Hoi}]{li2022blip}
\bibinfo{author}{Li, J.}, \bibinfo{author}{Li, D.}, \bibinfo{author}{Xiong, C.}, \bibinfo{author}{Hoi, S.}, \bibinfo{year}{2022}.
\newblock \bibinfo{title}{Blip: Bootstrapping language-image pre-training for unified vision-language understanding and generation}, in: \bibinfo{booktitle}{International conference on machine learning}, \bibinfo{organization}{PMLR}. pp. \bibinfo{pages}{12888--12900}.
\bibitem[{Li et~al.(2021b)Li, Selvaraju, Gotmare, Joty, Xiong and Hoi}]{li2021align}
\bibinfo{author}{Li, J.}, \bibinfo{author}{Selvaraju, R.}, \bibinfo{author}{Gotmare, A.}, \bibinfo{author}{Joty, S.}, \bibinfo{author}{Xiong, C.}, \bibinfo{author}{Hoi, S.C.H.}, \bibinfo{year}{2021}b.
\newblock \bibinfo{title}{Align before fuse: Vision and language representation learning with momentum distillation}.
\newblock \bibinfo{journal}{Advances in neural information processing systems} \bibinfo{volume}{34}, \bibinfo{pages}{9694--9705}.
\bibitem[{Liang et~al.(2020)Liang, Hu and Feng}]{liang2020we}
\bibinfo{author}{Liang, J.}, \bibinfo{author}{Hu, D.}, \bibinfo{author}{Feng, J.}, \bibinfo{year}{2020}.
\newblock \bibinfo{title}{Do we really need to access the source data? source hypothesis transfer for unsupervised domain adaptation}, in: \bibinfo{booktitle}{International conference on machine learning}, \bibinfo{organization}{PMLR}. pp. \bibinfo{pages}{6028--6039}.
\bibitem[{Liu et~al.(2024)Liu, Li, Wu and Lee}]{liu2024visual}
\bibinfo{author}{Liu, H.}, \bibinfo{author}{Li, C.}, \bibinfo{author}{Wu, Q.}, \bibinfo{author}{Lee, Y.J.}, \bibinfo{year}{2024}.
\newblock \bibinfo{title}{Visual instruction tuning}.
\newblock \bibinfo{journal}{Advances in neural information processing systems} \bibinfo{volume}{36}.
\bibitem[{Long et~al.(2015)Long, Cao, Wang and Jordan}]{long2015learning}
\bibinfo{author}{Long, M.}, \bibinfo{author}{Cao, Y.}, \bibinfo{author}{Wang, J.}, \bibinfo{author}{Jordan, M.}, \bibinfo{year}{2015}.
\newblock \bibinfo{title}{Learning transferable features with deep adaptation networks}, in: \bibinfo{booktitle}{International conference on machine learning}, \bibinfo{organization}{PMLR}. pp. \bibinfo{pages}{97--105}.
\bibitem[{Long et~al.(2018)Long, Cao, Wang and Jordan}]{long2018conditional}
\bibinfo{author}{Long, M.}, \bibinfo{author}{Cao, Z.}, \bibinfo{author}{Wang, J.}, \bibinfo{author}{Jordan, M.I.}, \bibinfo{year}{2018}.
\newblock \bibinfo{title}{Conditional adversarial domain adaptation}.
\newblock \bibinfo{journal}{Advances in neural information processing systems} \bibinfo{volume}{31}.
\bibitem[{Long et~al.(2016)Long, Zhu, Wang and Jordan}]{long2016unsupervised}
\bibinfo{author}{Long, M.}, \bibinfo{author}{Zhu, H.}, \bibinfo{author}{Wang, J.}, \bibinfo{author}{Jordan, M.I.}, \bibinfo{year}{2016}.
\newblock \bibinfo{title}{Unsupervised domain adaptation with residual transfer networks}.
\newblock \bibinfo{journal}{Advances in neural information processing systems} \bibinfo{volume}{29}.
\bibitem[{Long et~al.(2017)Long, Zhu, Wang and Jordan}]{long2017deep}
\bibinfo{author}{Long, M.}, \bibinfo{author}{Zhu, H.}, \bibinfo{author}{Wang, J.}, \bibinfo{author}{Jordan, M.I.}, \bibinfo{year}{2017}.
\newblock \bibinfo{title}{Deep transfer learning with joint adaptation networks}, in: \bibinfo{booktitle}{International conference on machine learning}, \bibinfo{organization}{PMLR}. pp. \bibinfo{pages}{2208--2217}.
\bibitem[{Lu et~al.(2024)Lu, Shen, Ma, Xie and Lai}]{lu2024mlnet}
\bibinfo{author}{Lu, Y.}, \bibinfo{author}{Shen, M.}, \bibinfo{author}{Ma, A.J.}, \bibinfo{author}{Xie, X.}, \bibinfo{author}{Lai, J.H.}, \bibinfo{year}{2024}.
\newblock \bibinfo{title}{Mlnet: Mutual learning network with neighborhood invariance for universal domain adaptation}, in: \bibinfo{booktitle}{Proceedings of the AAAI Conference on Artificial Intelligence}, pp. \bibinfo{pages}{3900--3908}.
\bibitem[{Miller et~al.(1990)Miller, Beckwith, Fellbaum, Gross and Miller}]{miller1990introduction}
\bibinfo{author}{Miller, G.A.}, \bibinfo{author}{Beckwith, R.}, \bibinfo{author}{Fellbaum, C.}, \bibinfo{author}{Gross, D.}, \bibinfo{author}{Miller, K.J.}, \bibinfo{year}{1990}.
\newblock \bibinfo{title}{Introduction to wordnet: An on-line lexical database}.
\newblock \bibinfo{journal}{International journal of lexicography} \bibinfo{volume}{3}, \bibinfo{pages}{235--244}.
\bibitem[{Ming et~al.(2022)Ming, Cai, Gu, Sun, Li and Li}]{ming2022delving}
\bibinfo{author}{Ming, Y.}, \bibinfo{author}{Cai, Z.}, \bibinfo{author}{Gu, J.}, \bibinfo{author}{Sun, Y.}, \bibinfo{author}{Li, W.}, \bibinfo{author}{Li, Y.}, \bibinfo{year}{2022}.
\newblock \bibinfo{title}{Delving into out-of-distribution detection with vision-language representations}.
\newblock \bibinfo{journal}{Advances in neural information processing systems} \bibinfo{volume}{35}, \bibinfo{pages}{35087--35102}.
\bibitem[{Oquab et~al.(2023)Oquab, Darcet, Moutakanni, Vo, Szafraniec, Khalidov, Fernandez, Haziza, Massa, El-Nouby et~al.}]{oquab2023dinov2}
\bibinfo{author}{Oquab, M.}, \bibinfo{author}{Darcet, T.}, \bibinfo{author}{Moutakanni, T.}, \bibinfo{author}{Vo, H.}, \bibinfo{author}{Szafraniec, M.}, \bibinfo{author}{Khalidov, V.}, \bibinfo{author}{Fernandez, P.}, \bibinfo{author}{Haziza, D.}, \bibinfo{author}{Massa, F.}, \bibinfo{author}{El-Nouby, A.}, et~al., \bibinfo{year}{2023}.
\newblock \bibinfo{title}{Dinov2: Learning robust visual features without supervision}.
\newblock \bibinfo{journal}{arXiv preprint arXiv:2304.07193} .
\bibitem[{Oza and Patel(2019)}]{oza2019c2ae}
\bibinfo{author}{Oza, P.}, \bibinfo{author}{Patel, V.M.}, \bibinfo{year}{2019}.
\newblock \bibinfo{title}{C2ae: Class conditioned auto-encoder for open-set recognition}, in: \bibinfo{booktitle}{Proceedings of the IEEE/CVF conference on computer vision and pattern recognition}, pp. \bibinfo{pages}{2307--2316}.
\bibitem[{Panareda~Busto and Gall(2017)}]{panareda2017open}
\bibinfo{author}{Panareda~Busto, P.}, \bibinfo{author}{Gall, J.}, \bibinfo{year}{2017}.
\newblock \bibinfo{title}{Open set domain adaptation}, in: \bibinfo{booktitle}{Proceedings of the IEEE international conference on computer vision}, pp. \bibinfo{pages}{754--763}.
\bibitem[{Peng et~al.(2019)Peng, Bai, Xia, Huang, Saenko and Wang}]{peng2019moment}
\bibinfo{author}{Peng, X.}, \bibinfo{author}{Bai, Q.}, \bibinfo{author}{Xia, X.}, \bibinfo{author}{Huang, Z.}, \bibinfo{author}{Saenko, K.}, \bibinfo{author}{Wang, B.}, \bibinfo{year}{2019}.
\newblock \bibinfo{title}{Moment matching for multi-source domain adaptation}, in: \bibinfo{booktitle}{Proceedings of the IEEE/CVF international conference on computer vision}, pp. \bibinfo{pages}{1406--1415}.
\bibitem[{Peng et~al.(2017)Peng, Usman, Kaushik, Hoffman, Wang and Saenko}]{peng2017visda}
\bibinfo{author}{Peng, X.}, \bibinfo{author}{Usman, B.}, \bibinfo{author}{Kaushik, N.}, \bibinfo{author}{Hoffman, J.}, \bibinfo{author}{Wang, D.}, \bibinfo{author}{Saenko, K.}, \bibinfo{year}{2017}.
\newblock \bibinfo{title}{Visda: The visual domain adaptation challenge}.
\newblock \bibinfo{journal}{arXiv preprint arXiv:1710.06924} .
\bibitem[{Qu et~al.(2024a)Qu, Hui, Cai and Liu}]{qu2024lmc}
\bibinfo{author}{Qu, H.}, \bibinfo{author}{Hui, X.}, \bibinfo{author}{Cai, Y.}, \bibinfo{author}{Liu, J.}, \bibinfo{year}{2024}a.
\newblock \bibinfo{title}{Lmc: Large model collaboration with cross-assessment for training-free open-set object recognition}.
\newblock \bibinfo{journal}{Advances in Neural Information Processing Systems} \bibinfo{volume}{36}.
\bibitem[{Qu et~al.(2024b)Qu, Zou, He, R{\"o}hrbein, Knoll, Chen and Jiang}]{qu2024lead}
\bibinfo{author}{Qu, S.}, \bibinfo{author}{Zou, T.}, \bibinfo{author}{He, L.}, \bibinfo{author}{R{\"o}hrbein, F.}, \bibinfo{author}{Knoll, A.}, \bibinfo{author}{Chen, G.}, \bibinfo{author}{Jiang, C.}, \bibinfo{year}{2024}b.
\newblock \bibinfo{title}{Lead: Learning decomposition for source-free universal domain adaptation}, in: \bibinfo{booktitle}{Proceedings of the IEEE/CVF Conference on Computer Vision and Pattern Recognition}, pp. \bibinfo{pages}{23334--23343}.
\bibitem[{Qu et~al.(2023a)Qu, Zou, R{\"o}hrbein, Lu, Chen, Tao and Jiang}]{qu2023upcycling}
\bibinfo{author}{Qu, S.}, \bibinfo{author}{Zou, T.}, \bibinfo{author}{R{\"o}hrbein, F.}, \bibinfo{author}{Lu, C.}, \bibinfo{author}{Chen, G.}, \bibinfo{author}{Tao, D.}, \bibinfo{author}{Jiang, C.}, \bibinfo{year}{2023}a.
\newblock \bibinfo{title}{Upcycling models under domain and category shift}, in: \bibinfo{booktitle}{Proceedings of the IEEE/CVF Conference on Computer Vision and Pattern Recognition}, pp. \bibinfo{pages}{20019--20028}.
\bibitem[{Qu et~al.(2023b)Qu, Zou, R{\"o}hrbein, Lu, Chen, Tao and Jiang}]{sanqing2023GLC}
\bibinfo{author}{Qu, S.}, \bibinfo{author}{Zou, T.}, \bibinfo{author}{R{\"o}hrbein, F.}, \bibinfo{author}{Lu, C.}, \bibinfo{author}{Chen, G.}, \bibinfo{author}{Tao, D.}, \bibinfo{author}{Jiang, C.}, \bibinfo{year}{2023}b.
\newblock \bibinfo{title}{Upcycling models under domain and category shift}, in: \bibinfo{booktitle}{Proceedings of the IEEE/CVF conference on computer vision and pattern recognition}, pp. \bibinfo{pages}{20019--20028}.
\bibitem[{Qu et~al.(2024c)Qu, Zou, R{\"o}hrbein, Lu, Chen, Tao and Jiang}]{qu2024glc++}
\bibinfo{author}{Qu, S.}, \bibinfo{author}{Zou, T.}, \bibinfo{author}{R{\"o}hrbein, F.}, \bibinfo{author}{Lu, C.}, \bibinfo{author}{Chen, G.}, \bibinfo{author}{Tao, D.}, \bibinfo{author}{Jiang, C.}, \bibinfo{year}{2024}c.
\newblock \bibinfo{title}{Glc++: Source-free universal domain adaptation through global-local clustering and contrastive affinity learning}.
\newblock \bibinfo{journal}{arXiv preprint arXiv:2403.14410} .
\bibitem[{Radford et~al.(2021)Radford, Kim, Hallacy, Ramesh, Goh, Agarwal, Sastry, Askell, Mishkin, Clark et~al.}]{radford2021learning}
\bibinfo{author}{Radford, A.}, \bibinfo{author}{Kim, J.W.}, \bibinfo{author}{Hallacy, C.}, \bibinfo{author}{Ramesh, A.}, \bibinfo{author}{Goh, G.}, \bibinfo{author}{Agarwal, S.}, \bibinfo{author}{Sastry, G.}, \bibinfo{author}{Askell, A.}, \bibinfo{author}{Mishkin, P.}, \bibinfo{author}{Clark, J.}, et~al., \bibinfo{year}{2021}.
\newblock \bibinfo{title}{Learning transferable visual models from natural language supervision}, in: \bibinfo{booktitle}{International conference on machine learning}, \bibinfo{organization}{PMLR}. pp. \bibinfo{pages}{8748--8763}.
\bibitem[{Russakovsky et~al.(2015)Russakovsky, Deng, Su, Krause, Satheesh, Ma, Huang, Karpathy, Khosla, Bernstein et~al.}]{russakovsky2015imagenet}
\bibinfo{author}{Russakovsky, O.}, \bibinfo{author}{Deng, J.}, \bibinfo{author}{Su, H.}, \bibinfo{author}{Krause, J.}, \bibinfo{author}{Satheesh, S.}, \bibinfo{author}{Ma, S.}, \bibinfo{author}{Huang, Z.}, \bibinfo{author}{Karpathy, A.}, \bibinfo{author}{Khosla, A.}, \bibinfo{author}{Bernstein, M.}, et~al., \bibinfo{year}{2015}.
\newblock \bibinfo{title}{Imagenet large scale visual recognition challenge}.
\newblock \bibinfo{journal}{International journal of computer vision} \bibinfo{volume}{115}, \bibinfo{pages}{211--252}.
\bibitem[{Saenko et~al.(2010)Saenko, Kulis, Fritz and Darrell}]{saenko2010adapting}
\bibinfo{author}{Saenko, K.}, \bibinfo{author}{Kulis, B.}, \bibinfo{author}{Fritz, M.}, \bibinfo{author}{Darrell, T.}, \bibinfo{year}{2010}.
\newblock \bibinfo{title}{Adapting visual category models to new domains}, in: \bibinfo{booktitle}{Computer Vision--ECCV 2010: 11th European Conference on Computer Vision, Heraklion, Crete, Greece, September 5-11, 2010, Proceedings, Part IV 11}, \bibinfo{organization}{Springer}. pp. \bibinfo{pages}{213--226}.
\bibitem[{Saito et~al.(2020)Saito, Kim, Sclaroff and Saenko}]{saito2020universal}
\bibinfo{author}{Saito, K.}, \bibinfo{author}{Kim, D.}, \bibinfo{author}{Sclaroff, S.}, \bibinfo{author}{Saenko, K.}, \bibinfo{year}{2020}.
\newblock \bibinfo{title}{Universal domain adaptation through self supervision}.
\newblock \bibinfo{journal}{Advances in neural information processing systems} \bibinfo{volume}{33}, \bibinfo{pages}{16282--16292}.
\bibitem[{Saito et~al.(2023)Saito, Kim, Teterwak, Feris and Saenko}]{saito2023mind}
\bibinfo{author}{Saito, K.}, \bibinfo{author}{Kim, D.}, \bibinfo{author}{Teterwak, P.}, \bibinfo{author}{Feris, R.}, \bibinfo{author}{Saenko, K.}, \bibinfo{year}{2023}.
\newblock \bibinfo{title}{Mind the backbone: Minimizing backbone distortion for robust object detection}.
\newblock \bibinfo{journal}{arXiv preprint arXiv:2303.14744} .
\bibitem[{Saito and Saenko(2021)}]{saito2021ovanet}
\bibinfo{author}{Saito, K.}, \bibinfo{author}{Saenko, K.}, \bibinfo{year}{2021}.
\newblock \bibinfo{title}{Ovanet: One-vs-all network for universal domain adaptation}, in: \bibinfo{booktitle}{Proceedings of the ieee/cvf international conference on computer vision}, pp. \bibinfo{pages}{9000--9009}.
\bibitem[{Saito et~al.(2018)Saito, Watanabe, Ushiku and Harada}]{saito2018maximum}
\bibinfo{author}{Saito, K.}, \bibinfo{author}{Watanabe, K.}, \bibinfo{author}{Ushiku, Y.}, \bibinfo{author}{Harada, T.}, \bibinfo{year}{2018}.
\newblock \bibinfo{title}{Maximum classifier discrepancy for unsupervised domain adaptation}, in: \bibinfo{booktitle}{Proceedings of the IEEE conference on computer vision and pattern recognition}, pp. \bibinfo{pages}{3723--3732}.
\bibitem[{Scheirer et~al.(2012)Scheirer, de~Rezende~Rocha, Sapkota and Boult}]{scheirer2012toward}
\bibinfo{author}{Scheirer, W.J.}, \bibinfo{author}{de~Rezende~Rocha, A.}, \bibinfo{author}{Sapkota, A.}, \bibinfo{author}{Boult, T.E.}, \bibinfo{year}{2012}.
\newblock \bibinfo{title}{Toward open set recognition}.
\newblock \bibinfo{journal}{IEEE transactions on pattern analysis and machine intelligence} \bibinfo{volume}{35}, \bibinfo{pages}{1757--1772}.
\bibitem[{Shu et~al.(2022)Shu, Nie, Huang, Yu, Goldstein, Anandkumar and Xiao}]{shu2022test}
\bibinfo{author}{Shu, M.}, \bibinfo{author}{Nie, W.}, \bibinfo{author}{Huang, D.A.}, \bibinfo{author}{Yu, Z.}, \bibinfo{author}{Goldstein, T.}, \bibinfo{author}{Anandkumar, A.}, \bibinfo{author}{Xiao, C.}, \bibinfo{year}{2022}.
\newblock \bibinfo{title}{Test-time prompt tuning for zero-shot generalization in vision-language models}.
\newblock \bibinfo{journal}{Advances in Neural Information Processing Systems} \bibinfo{volume}{35}, \bibinfo{pages}{14274--14289}.
\bibitem[{Tang et~al.(2024)Tang, Su, Ye and Zhu}]{tang2024source}
\bibinfo{author}{Tang, S.}, \bibinfo{author}{Su, W.}, \bibinfo{author}{Ye, M.}, \bibinfo{author}{Zhu, X.}, \bibinfo{year}{2024}.
\newblock \bibinfo{title}{Source-free domain adaptation with frozen multimodal foundation model}, in: \bibinfo{booktitle}{Proceedings of the IEEE/CVF Conference on Computer Vision and Pattern Recognition}, pp. \bibinfo{pages}{23711--23720}.
\bibitem[{Trivedi et~al.(2023)Trivedi, Koutra and Thiagarajan}]{trivedi2023closer}
\bibinfo{author}{Trivedi, P.}, \bibinfo{author}{Koutra, D.}, \bibinfo{author}{Thiagarajan, J.J.}, \bibinfo{year}{2023}.
\newblock \bibinfo{title}{A closer look at model adaptation using feature distortion and simplicity bias}.
\newblock \bibinfo{journal}{arXiv preprint arXiv:2303.13500} .
\bibitem[{Tzeng et~al.(2017)Tzeng, Hoffman, Saenko and Darrell}]{tzeng2017adversarial}
\bibinfo{author}{Tzeng, E.}, \bibinfo{author}{Hoffman, J.}, \bibinfo{author}{Saenko, K.}, \bibinfo{author}{Darrell, T.}, \bibinfo{year}{2017}.
\newblock \bibinfo{title}{Adversarial discriminative domain adaptation}, in: \bibinfo{booktitle}{Proceedings of the IEEE conference on computer vision and pattern recognition}, pp. \bibinfo{pages}{7167--7176}.
\bibitem[{Vaze et~al.(2022)Vaze, Han, Vedaldi and Zisserman}]{vaze2022openset}
\bibinfo{author}{Vaze, S.}, \bibinfo{author}{Han, K.}, \bibinfo{author}{Vedaldi, A.}, \bibinfo{author}{Zisserman, A.}, \bibinfo{year}{2022}.
\newblock \bibinfo{title}{Open-set recognition: A good closed-set classifier is all you need}, in: \bibinfo{booktitle}{International Conference on Learning Representations}.
\newblock \URLprefix \url{https://openreview.net/forum?id=5hLP5JY9S2d}.
\bibitem[{Venkateswara et~al.(2017)Venkateswara, Eusebio, Chakraborty and Panchanathan}]{venkateswara2017deep}
\bibinfo{author}{Venkateswara, H.}, \bibinfo{author}{Eusebio, J.}, \bibinfo{author}{Chakraborty, S.}, \bibinfo{author}{Panchanathan, S.}, \bibinfo{year}{2017}.
\newblock \bibinfo{title}{Deep hashing network for unsupervised domain adaptation}, in: \bibinfo{booktitle}{Proceedings of the IEEE conference on computer vision and pattern recognition}, pp. \bibinfo{pages}{5018--5027}.
\bibitem[{Wang et~al.(2023)Wang, Li, Yao and Li}]{wang2023clipn}
\bibinfo{author}{Wang, H.}, \bibinfo{author}{Li, Y.}, \bibinfo{author}{Yao, H.}, \bibinfo{author}{Li, X.}, \bibinfo{year}{2023}.
\newblock \bibinfo{title}{Clipn for zero-shot ood detection: Teaching clip to say no}, in: \bibinfo{booktitle}{Proceedings of the IEEE/CVF International Conference on Computer Vision}, pp. \bibinfo{pages}{1802--1812}.
\bibitem[{Wang et~al.(2024a)Wang, Zhang, Song, Li, Rosin and Zhang}]{wang2024exploiting}
\bibinfo{author}{Wang, Y.}, \bibinfo{author}{Zhang, L.}, \bibinfo{author}{Song, R.}, \bibinfo{author}{Li, H.}, \bibinfo{author}{Rosin, P.L.}, \bibinfo{author}{Zhang, W.}, \bibinfo{year}{2024}a.
\newblock \bibinfo{title}{Exploiting inter-sample affinity for knowability-aware universal domain adaptation}.
\newblock \bibinfo{journal}{International Journal of Computer Vision} \bibinfo{volume}{132}, \bibinfo{pages}{1800--1816}.
\bibitem[{Wang et~al.(2024b)Wang, Liang, Sheng, He, Wang and Tan}]{wang2024hard}
\bibinfo{author}{Wang, Z.}, \bibinfo{author}{Liang, J.}, \bibinfo{author}{Sheng, L.}, \bibinfo{author}{He, R.}, \bibinfo{author}{Wang, Z.}, \bibinfo{author}{Tan, T.}, \bibinfo{year}{2024}b.
\newblock \bibinfo{title}{A hard-to-beat baseline for training-free clip-based adaptation}.
\newblock \bibinfo{journal}{arXiv preprint arXiv:2402.04087} .
\bibitem[{Wortsman et~al.(2022)Wortsman, Ilharco, Kim, Li, Kornblith, Roelofs, Lopes, Hajishirzi, Farhadi, Namkoong et~al.}]{wortsman2022robust}
\bibinfo{author}{Wortsman, M.}, \bibinfo{author}{Ilharco, G.}, \bibinfo{author}{Kim, J.W.}, \bibinfo{author}{Li, M.}, \bibinfo{author}{Kornblith, S.}, \bibinfo{author}{Roelofs, R.}, \bibinfo{author}{Lopes, R.G.}, \bibinfo{author}{Hajishirzi, H.}, \bibinfo{author}{Farhadi, A.}, \bibinfo{author}{Namkoong, H.}, et~al., \bibinfo{year}{2022}.
\newblock \bibinfo{title}{Robust fine-tuning of zero-shot models}, in: \bibinfo{booktitle}{Proceedings of the IEEE/CVF conference on computer vision and pattern recognition}, pp. \bibinfo{pages}{7959--7971}.
\bibitem[{You et~al.(2019a)You, Long, Cao, Wang and Jordan}]{You_2019_CVPR}
\bibinfo{author}{You, K.}, \bibinfo{author}{Long, M.}, \bibinfo{author}{Cao, Z.}, \bibinfo{author}{Wang, J.}, \bibinfo{author}{Jordan, M.I.}, \bibinfo{year}{2019}a.
\newblock \bibinfo{title}{Universal domain adaptation}, in: \bibinfo{booktitle}{Proceedings of the IEEE/CVF conference on computer vision and pattern recognition}, pp. \bibinfo{pages}{2720--2729}.
\bibitem[{You et~al.(2019b)You, Long, Cao, Wang and Jordan}]{you2019universal}
\bibinfo{author}{You, K.}, \bibinfo{author}{Long, M.}, \bibinfo{author}{Cao, Z.}, \bibinfo{author}{Wang, J.}, \bibinfo{author}{Jordan, M.I.}, \bibinfo{year}{2019}b.
\newblock \bibinfo{title}{Universal domain adaptation}, in: \bibinfo{booktitle}{Proceedings of the IEEE/CVF conference on computer vision and pattern recognition}, pp. \bibinfo{pages}{2720--2729}.
\bibitem[{Zanella and Ben~Ayed(2024)}]{zanella2024test}
\bibinfo{author}{Zanella, M.}, \bibinfo{author}{Ben~Ayed, I.}, \bibinfo{year}{2024}.
\newblock \bibinfo{title}{On the test-time zero-shot generalization of vision-language models: Do we really need prompt learning?}, in: \bibinfo{booktitle}{Proceedings of the IEEE/CVF Conference on Computer Vision and Pattern Recognition}, pp. \bibinfo{pages}{23783--23793}.
\bibitem[{Zara et~al.(2023)Zara, Roy, Rota and Ricci}]{zara2023autolabel}
\bibinfo{author}{Zara, G.}, \bibinfo{author}{Roy, S.}, \bibinfo{author}{Rota, P.}, \bibinfo{author}{Ricci, E.}, \bibinfo{year}{2023}.
\newblock \bibinfo{title}{Autolabel: Clip-based framework for open-set video domain adaptation}, in: \bibinfo{booktitle}{Proceedings of the IEEE/CVF Conference on Computer Vision and Pattern Recognition}, pp. \bibinfo{pages}{11504--11513}.
\bibitem[{Zhang et~al.(2023)Zhang, Shen and Foo}]{zhang2023rethinking}
\bibinfo{author}{Zhang, W.}, \bibinfo{author}{Shen, L.}, \bibinfo{author}{Foo, C.S.}, \bibinfo{year}{2023}.
\newblock \bibinfo{title}{Rethinking the role of pre-trained networks in source-free domain adaptation}, in: \bibinfo{booktitle}{Proceedings of the IEEE/CVF International Conference on Computer Vision}, pp. \bibinfo{pages}{18841--18851}.
\bibitem[{Zhao et~al.(2024)Zhao, Dai, Li, Hu, Zhang and Liu}]{zhao2024ltgc}
\bibinfo{author}{Zhao, Q.}, \bibinfo{author}{Dai, Y.}, \bibinfo{author}{Li, H.}, \bibinfo{author}{Hu, W.}, \bibinfo{author}{Zhang, F.}, \bibinfo{author}{Liu, J.}, \bibinfo{year}{2024}.
\newblock \bibinfo{title}{Ltgc: Long-tail recognition via leveraging llms-driven generated content}, in: \bibinfo{booktitle}{Proceedings of the IEEE/CVF Conference on Computer Vision and Pattern Recognition}, pp. \bibinfo{pages}{19510--19520}.
\bibitem[{Zhou et~al.(2022)Zhou, Yang, Loy and Liu}]{zhou2022learning}
\bibinfo{author}{Zhou, K.}, \bibinfo{author}{Yang, J.}, \bibinfo{author}{Loy, C.C.}, \bibinfo{author}{Liu, Z.}, \bibinfo{year}{2022}.
\newblock \bibinfo{title}{Learning to prompt for vision-language models}.
\newblock \bibinfo{journal}{International Journal of Computer Vision} \bibinfo{volume}{130}, \bibinfo{pages}{2337--2348}.

\end{thebibliography}

\end{document}